\documentclass[journal]{IEEEtran}

\pdfoutput=1
\usepackage{hyperref}       % hyperlinks
\usepackage{url}            % simple URL typesetting
\usepackage{booktabs}       % professional-quality tables
\usepackage{amsfonts}       % blackboard math symbols
\usepackage{nicefrac}       % compact symbols for 1/2, etc.
\usepackage{microtype}      % microtypography
\usepackage{lipsum}
\usepackage[ruled,vlined,linesnumbered]{algorithm2e}
\usepackage{setspace}
\usepackage{amsmath}
\usepackage{amsthm}

\usepackage{graphicx}
\usepackage{subcaption}
\usepackage{xcolor}
\usepackage{cite}
\usepackage{mathtools}
\usepackage{soul}

\newcommand{\fedave}{\textsf{FedAvg}}
\newcommand{\fedprox}{\textsf{FedProx}}
\newcommand{\fednu}{\textsf{FedNu}}
\newcommand{\lb}{$\mathsf{LB}$}
\newcommand{\method}{$\mathsf{FOLB}$}
\newcommand{\norm}[1]{\left\lVert#1\right\rVert}
\allowdisplaybreaks

\newtheorem{theorem}{Theorem}
\newtheorem{definition}{Definition}
\newtheorem{lemma}{Lemma}
\newtheorem{proposition}{Proposition}

\DeclareMathOperator{\sign}{sign}

\begin{document}

\title{Fast-Convergent Federated Learning}
%
%
% author names and IEEE memberships
% note positions of commas and nonbreaking spaces ( ~ ) LaTeX will not break
% a structure at a ~ so this keeps an author's name from being broken across
% two lines.
% use \thanks{} to gain access to the first footnote area
% a separate \thanks must be used for each paragraph as LaTeX2e's \thanks
% was not built to handle multiple paragraphs
%
\author{Hung T. Nguyen, Vikash Sehwag, Seyyedali   Hosseinalipour,~\IEEEmembership{Member,~IEEE}, Christopher G. Brinton,~\IEEEmembership{Senior Member,~IEEE,} Mung Chiang,~\IEEEmembership{Fellow,~IEEE,} H. Vincent Poor,~\IEEEmembership{Fellow,~IEEE}
	
\thanks{The work of Hung T. Nguyen and Mung Chiang was supported in part by the Defense Advanced Research Projects Agency (DARPA) under contract no. AWD1005371 and AWD1005468. The work of H. V. Poor was supported in part by the  U.S. National Science Foundation under Grant CCF-1908308.}
	
\thanks{H. Nguyen, V. Sehwag and H. V. Poor are with the Department of Electrical Engineering, Princeton University, Princeton, NJ 08544. email: \{hn4,vvikash,poor\}@princeton.edu.}
\thanks{S. Hosseinalipour, C. Brinton and M. Chiang are with the School of Electrical and Computer Engineering, Purdue University, West Lafayette, IN 47907. email: \{hosseina,cgb,chiang\}@purdue.edu.}}

% note the % following the last \IEEEmembership and also \thanks - 
% these prevent an unwanted space from occurring between the last author name
% and the end of the author line. i.e., if you had this:
% 
% \author{....lastname \thanks{...} \thanks{...} }
%                     ^------------^------------^----Do not want these spaces!
%
% a space would be appended to the last name and could cause every name on that
% line to be shifted left slightly. This is one of those "LaTeX things". For
% instance, "\textbf{A} \textbf{B}" will typeset as "A B" not "AB". To get
% "AB" then you have to do: "\textbf{A}\textbf{B}"
% \thanks is no different in this regard, so shield the last } of each \thanks
% that ends a line with a % and do not let a space in before the next \thanks.
% Spaces after \IEEEmembership other than the last one are OK (and needed) as
% you are supposed to have spaces between the names. For what it is worth,
% this is a minor point as most people would not even notice if the said evil
% space somehow managed to creep in.

% make the title area
\maketitle

% As a general rule, do not put math, special symbols or citations
% in the abstract or keywords.
\begin{abstract}
Federated learning has emerged recently as a promising solution for distributing machine learning tasks through modern networks of mobile devices. Recent studies have obtained lower bounds on the expected decrease in model loss that is achieved through each round of federated learning. However, convergence generally requires a large number of communication rounds, which induces delay in model training and is costly in terms of network resources. In this paper, we propose a fast-convergent federated learning algorithm, called \method{}, which performs intelligent sampling of devices in each round of model training to optimize the expected convergence speed. We first theoretically characterize a lower bound on improvement that can be obtained in each round if devices are selected according to the expected improvement their local models will provide to the current global model. Then, we show that \method{} obtains this bound through uniform sampling by weighting device updates according to their gradient information. \method{} is able to handle both communication and computation heterogeneity of devices by adapting the aggregations according to estimates of device's capabilities of contributing to the updates. We evaluate \method{} in comparison with existing federated learning algorithms and experimentally show its improvement in trained model accuracy, convergence speed, and/or model stability across various machine learning tasks and datasets.
\end{abstract}

% Note that keywords are not normally used for peerreview papers.
\begin{IEEEkeywords}
Federated learning, distributed optimization, fast convergence rate
\end{IEEEkeywords}

% For peer review papers, you can put extra information on the cover
% page as needed:
% \ifCLASSOPTIONpeerreview
% \begin{center} \bfseries EDICS Category: 3-BBND \end{center}
% \fi
%
% For peerreview papers, this IEEEtran command inserts a page break and
% creates the second title. It will be ignored for other modes.
\IEEEpeerreviewmaketitle

% !TEX root = fedlearn_nonuniform.tex

\section{Introduction}
Over the past decade, the intelligence of devices at the network edge has increased substantially. Today, smartphones, wearables, sensors, and other Internet-connected devices possess significant computation and communication capabilities, especially when considered collectively. This has created interest in migrating computing methodologies from cloud to edge-centric to provide near-real-time results \cite{chiang2016fog}.

Most applications of interest today involve machine learning (ML). Federated learning (FL) has emerged recently as a technique for distributing ML model training across edge devices. It allows solving machine learning tasks in a distributing setting comprising a central server and multiple participating ``worker'' nodes, where the nodes themselves collect the data and never transfer it over the network, which minimizes privacy concerns. At the same time, the  federated learning setting introduces challenges of statistical and system heterogeneity that traditional distributed optimization methods \cite{boyd11,dean12,dekel12,li14,reddi16,richtarik16,shamir14,smith16,zhang15,zhang12} are not designed for and thus may fail to provide convergence guarantees. 

One such challenge is the number of devices that must participate in each round of computation. To provide convergence guarantees, recent studies \cite{yu18,jiang18,yu19,wang19} in distributed learning have to assume full participation of all devices in every round of optimization, which results in excessively high communication costs in edge network settings. On the other hand, \cite{lin18, reddi16, shamir14, stich18,wang18,woodworth18, zhang15} violate the statistical heterogeneity property. In contrast, FL techniques provide flexibility in selecting only a fraction of clients in each round of computations~\cite{mcmahan17}. However, such a selection of devices, which is often done uniformly, naturally causes the convergence rates to be slower.

\begin{figure}[t]
	\vspace{-0.2in}
	\centering
	\includegraphics[width=0.85\linewidth]{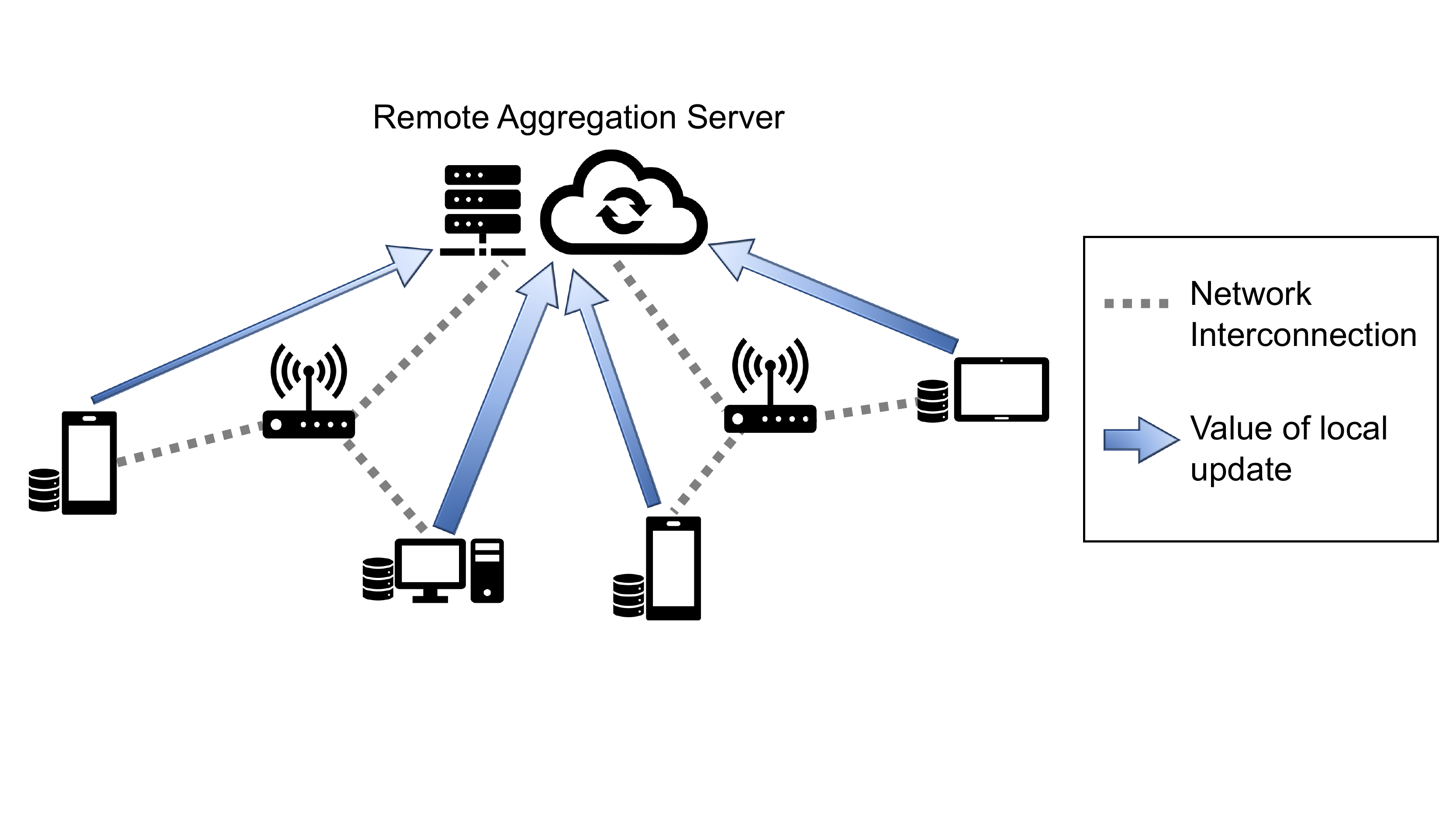}
	\vspace{-0.35in}
	\caption{Different from standard federated learning algorithms which are based on uniform sampling, our proposed methodology improves convergence rates through intelligent sampling that factors in the values of local updates that devices provide.}
	\label{fig:importance_sampling}
	\vspace{-0.15in}
\end{figure}

In this paper, we take into consideration that in each computation round, some clients provide more valuable updates in terms of reducing the overall model loss than others, as illustrated in Figure~\ref{fig:importance_sampling}. By taking this into account, we show that the convergence in federated learning can be vastly improved with an appropriate non-uniform device selection method. We first theoretically characterize the overall loss decrease of the \emph{non-uniform} version of the recent state-of-the-art \fedprox{} algorithm \cite{tian2019}, where clients in each round are selected based on a target probability distribution. Under such a non-uniform device selection scheme, we obtain a lower bound on the expected decrease in global loss function at every computation round at the central server. We further improve this bound by incorporating gradient information from each device into the aggregation of local parameter updates and characterize a device selection distribution, named \emph{LB-near-optimal}, which can achieve a near-optimal lower bound over all non-uniform distributions at each round.

Straightforwardly computing such distribution in every round involves a heavy communication step across \emph{all devices} which defeats the purpose of federated learning where the assumption is that only a subset of devices participates in each round. We address this communication challenge with a novel federated learning algorithm, named \method{}, which is based on a simple yet effective re-weighting mechanism of updated parameters received from participating devices in every round. With twice the number of devices selected in baseline federated learning settings, i.e., as in the popular \fedave{} and \fedprox{} algorithms, \method{} achieves the near-optimal decrease in global loss as that of the LB-near-optimal device selection distribution, whereas with the same number of devices, \method{} provides a guarantee of global loss decrease close to that of the LB-near-optimal and even better in some cases.

% with minimal additional computation and communication overhead compared with baseline federated learning algorithms. We demonstrate that an LB-near-optimal decrease in global loss can be obtained by using only twice the number of devices selected in baseline federated learning setting. However, it still incurs twice the communication cost compared to the baseline. 

Another challenge in federated learning is device heterogeneity, which affects the computation and communication capabilities across devices. We demonstrate that \method{} can easily adapt to such device heterogeneity by adjusting its re-weighting mechanism of the updated parameters returned from participating devices. Computing the re-weighting coefficients involves presumed constants which are related to the loss function characteristics and solvers used in distributed devices, and more importantly, may not be available beforehand. Even estimating those constants may be difficult and incur considerable computation and communication overhead. Thus, we show a greater flexibility of \method{} that its re-weighting mechanism can group all presumed constants into a \emph{single} hyper-parameter which can be optimized with line search.

%\subsection{Summary of Contributions}
\subsection{Outline and Summary of Contributions}
Compared to related work (discussed next), in this paper we make the following contributions:
\begin{itemize}
	\item We provide a theoretical characterization of fast federated learning based on a non-uniform selection of participating devices. In particular, we establish lower bounds on decrease in global loss given a non-uniform device selection from any target distribution, and compare these bounds directly with \fedprox{}. We demonstrate how local gradient information from each devices can be aggregated to improve the lower bound and also compute a near-optimal distribution for device selection (Section~\ref{sec:nufedprox}). 
	\item We propose \method{}, a federated learning algorithm which employs an accurate and communication-efficient approximation of a near-optimal distribution of device selection to accelerate convergence (Section~\ref{sec:folb}). 
	\item We show a successful generalization on \method{} in federated learning with computation and communication heterogeneity among participating devices (Section~\ref{sec:hetero_folb}). 
	\item We perform extensive experiments on synthetic, vision, and language datasets to demonstrate the success of \method{} over \fedave{} and \fedprox{} algorithms in terms of model accuracy, training stability, and/or convergence speed (Section~\ref{sec:exps}). 
\end{itemize}

\subsection{Related work}
Distributed optimization has been vastly studied in the literature \cite{boyd11,dean12,dekel12,li14,reddi16,richtarik16,shamir14,smith16,zhang15,zhang12} which focuses on a datacenter environment model where (i) the distribution of data to different machine is under control, e.g., uniformly at random, and (ii) all the machines are relatively close to one another, e.g., minimal cost of communication. However, those approaches no longer work on the emerging environment of distributed mobile devices due to its peculiar characteristics, including non-i.i.d. and unbalanced data distributions, limited communication, and heterogeneity of computation between devices. Thus, many recent efforts \cite{mcmahan17,tian2019,yu18,jiang18,yu19,wang19,lin18, reddi16, shamir14, stich18,wang18,woodworth18, zhang15,dinh2019federated,reddi2020adaptive,karimireddy2019scaffold} have been devoted to coping with these new challenges.

Most of the existing works \cite{yu18,jiang18,yu19,wang19,lin18, reddi16, shamir14, stich18,wang18,woodworth18, zhang15} either assume the full participation of all devices or violate statistical heterogeneity property inherent in our environment. McMahan \textit{et al.} \cite{mcmahan17} was the first to define federated learning setting in which a learning task is solved by a loose federation of participating devices which are coordinated by a central server and proposed the heuristic \fedave{} algorithm. \fedave{} runs through multiple rounds of optimization, in each round, it randomly selects a small set of $K$ devices to perform local stochastic gradient descent with respect to their local data. Then, the locally updated model parameters are sent back to the central server where an averaging is taken and regarded as new parameters. It was shown in \cite{mcmahan17} to perform well in terms of both performance and communication cost. More recently, \cite{li19} shows convergence rate of \fedave{} when the cost function is strongly convex and smooth. Federated multi-task learning was proposed in \cite{smith17} that allows slightly different models in different devices and framed the problem in multi-task learning framework. More recent work in \cite{reddi2020adaptive,karimireddy2019scaffold} propose federated optimizers and algorithms that improve over \fedave{} in terms of convergence rate subject to a number of assumptions about the loss functions and non-i.i.d. distributions of data. However, heterogeneity in computation and communication across devices have not been a focus of these models.

Very recently, \cite{tian2019} proposed \fedprox{} with the main difference from \fedave{} of adding a proximal term in every local loss function to keep the updated parameters across devices more similar. \fedprox{} follows the same steps as \fedave{}, however, it provides convergence rate for both convex and non-convex losses and deals with statistical heterogeneity. \fedprox{} also allows any local optimizer at the local devices. Our work utilizes the idea of adding a proximal term to local loss function, however, our proposed algorithm \method{} takes a unique approach that aims at a near-optimal device selection distribution to maximize the loss decrease at every round of optimization. On the other hand, \fedprox{} and \fedave{} select devices uniformly at random in each round.

%[Heterogeneity]
%
%[Topology]
%
%[Latency]

Other aspects of federated learning have also been studied, such as privacy of user data \cite{bonawitz17,bhowmick18,agarwal18,ghazi19,liu19}, fairness in federated learning \cite{li19fair}, federated learning over communication systems \cite{jeon2020gradient,amiriconvergence,chen2019joint,chen2020convergence,shlezinger2020federated}, and federated learning for edge networks \cite{wang2020convergence,khan2019federated}. We refer the interested reader to comprehensive surveys in \cite{kairouz19, li19survey} and references therein for more details.

% !TEX root = fedlearn_nonuniform.tex

\section{Preliminaries and Modeling Assumptions}
We first formalize federated learning, including the standard system model (Section \ref{ssec:system}), learning algorithms (Section \ref{ssec:fedalg}), and common theoretical assumptions (Section \ref{ssec:assumptions}).

\subsection{System and Learning Model}
\label{ssec:system}
Consider a network of $N$ devices, indexed $k \in \{1,...,N\}$, where each device possesses its own local (private) dataset $D_k$. Each data point $d \in D_k$ is assumed to contain a feature vector $\mathbf{x}_d$ and a target variable $y_d$. The objective of federated learning is to train a machine learning (ML) model of interest over this network, i.e., to learn a mapping $g_\mathbf{w}: \mathbf{x}_d \rightarrow \hat{y}_d$ from a given input sample $\mathbf{x}_d$ to a predicted output $\hat{y}_d$ parameterized by a vector $\mathbf{w}$, with each device processing its own data to minimize communication overhead.

For our purposes, an ML model is specified according to its parameter vector $\mathbf{w}$ and loss function $f(\mathbf{w}) = (1 / |D|) \sum_{d \in D} l(\mathbf{w}, \mathbf{x}_d, y_d)$ to be minimized. Here, $D$ is the training dataset available, and $l(\mathbf{w}, \mathbf{x}_d, y_d)$ represents the error between $\hat{y}_d$ and $y_d$ (e.g., the squared distance). Thus, we seek to find $\mathbf{w}$ that minimizes $f(\mathbf{w})$ over the data $D = \cup_k D_k$  in the network. In federated learning, this minimization is not performed directly, as each device $k$ only has access to $D_k$. Defining $F_k(\mathbf{w}) = (1 / |D_k|) \sum_{d \in D_k} l_d$ as the local loss function at $k$ over $D_k$, if we assume that $|D_i| = |D_j| \; \forall i,j$, i.e., each device processes the same amount of data, we can express the optimization as an average over the $F_k(\mathbf{w})$:
\vspace{-0.03in}
\begin{equation}
	\label{eq:decomp}
	\min_\mathbf{w} f(\mathbf{w}), \text{ where } f(\mathbf{w}) \coloneqq \frac{1}{N}\sum_{k = 1}^N F_k(\mathbf{w}).
\end{equation}
More generally, nodes may process different amounts of data, e.g., due to heterogeneous compute capabilities. In such cases, we can replace the factor $1 / N$ with $p_k = |D_k| / |D|$ for a weighted average of the $F_k(\mathbf{w})$ \cite{tian2019,tu20}. This is the approach we take throughout this paper.

Federated learning algorithms differ in how \eqref{eq:decomp} is solved. In our case, we will assume that a central server is available to orchestrate the learning across the devices. Such a scenario is increasingly common in fog or edge computing systems, where an edge server may be connected to several edge devices, e.g., in a smart factory \cite{tu20}. We will next introduce the standard algorithms for federated learning in these environments.
\vspace{-0.05in}

\subsection{Standard Federated Learning Algorithms}
\label{ssec:fedalg}
Federated learning algorithms generally solve \eqref{eq:decomp} in three steps: local learning, aggregation, and synchronization, which are repeated over several rounds \cite{mcmahan17}. In each round $t$, the server selects a set $\mathcal{K}_t$ of $K$ devices among the $N$ total to update the current estimate $\mathbf{w}^t$ for the optimal set of parameters $\mathbf{w}^{\star}$. Each device $k \in \mathcal{K}_t$ selected then updates $\mathbf{w}^t$ based on its local loss $F_k(\mathbf{w})$, producing $\mathbf{w}_k^{t+1}$, and sends this back to the server. The server then aggregates these locally updated parameters according to
\begin{equation}
\label{eq:agg}
\mathbf{w}^{t+1} = \frac{1}{K} \sum_{k \in \mathcal{K}_t} \mathbf{w}^{t+1}_k,
\end{equation}
and synchronizes the devices with this update before beginning the next round.

\fedave{} \cite{mcmahan17} is the standard federated learning algorithm that uses this framework. In \fedave{}, the loss $F_k(\mathbf{w})$ is directly minimized during the local update step, using gradient descent techniques. Formally, each device calculates $\mathbf{w}_k^{t+1} = \mathbf{w}^t - \eta \nabla F_k(\mathbf{w}^t)$, where $\nabla F_k(\mathbf{w}^t) = (1 / |D_k|) \sum_{d \in D_k} \nabla l(\mathbf{w}^t, \mathbf{x}_d, y_d)$ is the average of the loss gradient over device $k$'s data. It is also possible to use multiple iterations of local updates between global aggregations \cite{wang19}.

More recently, \fedprox{} was introduced \cite{tian2019}, which differs from \fedave{} in the local update step: instead of minimizing $F_k(\mathbf{w})$ at device $k$, it minimizes
\begin{equation}
	\label{eq:prox}
	h_k(\mathbf{w}, \mathbf{w}^t) = F_k(\mathbf{w}) + \frac{\mu}{2} \norm{\mathbf{w} - \mathbf{w}^t}^2.
\end{equation}
The proximal term $\frac{\mu}{2} \norm{\mathbf{w} - \mathbf{w}^t}^2$ added to each local loss function brings two modeling benefits: (i) it restricts the divergence of parameters between devices that will arise due to heterogeneity in their data distributions, and (ii) for appropriate choice of $\mu$, it will turn a non-convex loss function $F_k(\mathbf{w})$ into a convex $h_k(\mathbf{w}, \mathbf{w}^t)$ which is easier to optimize. The approach we develop beginning in Section \ref{sec:nufedprox} will build on \fedprox{}. Note that by setting $\mu = 0$, $h_k(\mathbf{w},\mathbf{w}^t) = F_k(\mathbf{w})$ and we get back the setting in \fedave{}. Thus, our algorithm \method{} naturally applies on \fedave{} and our theoretical results still hold if all $F_k(\mathbf{w}), k = 1,\dots,N$ are strongly convex.

%Both of these algorithms share a common strategy for optimizing the learning model while securely preserving the privacy. They perform a number of rounds, each of which randomly selects $K$ devices among the total $N$ available and request those to update the parameters according to their loss function $F_k(w)$. After receiving the updated parameters from $K$ devices at the central server, a simple averaging is performed to aggregate those locally optimized parameters as follows:
%where $\mathbf{w}^{t}$ is the global set of parameters at round $t$ and $\mathbf{w}^{t+1}_k$ is the set of updated parameters at device $k$.

\subsection{ML Model Assumptions}
\label{ssec:assumptions}
For theoretical analysis of federated learning algorithms, a few standard assumptions are typically made on the ML models (see e.g., \cite{wang19,tian2019,tu20}). We will employ the following in our analysis: \vspace{0.05in}

\noindent \textbf{Assumption 1 ($L$-Lipschitz gradient).} $F_k(\mathbf{w})$ is $L$-Lipschitz gradient for each device $k \in \{1,...,N\}$, i.e., $\norm{\nabla F_k(\mathbf{w}) - \nabla F_k(\mathbf{w}')} \leq L \norm{\mathbf{w} - \mathbf{w}'}$ for any two parameter vectors $\mathbf{w}, \mathbf{w}'$. This also implies (via the triangle inequality) that that the global $f(\mathbf{w})$ is $L$-Lipschitz gradient. \vspace{0.05in} 

\noindent \textbf{Assumption 2 ($B$-dissimilar gradients).} The gradient of $F_k(\mathbf{w})$ is at most $B$-dissimilar from $f(\mathbf{w})$ for each $k$, i.e., $\norm{\nabla F_k(\mathbf{w})} \leq B \norm{\nabla f(\mathbf{w})}$ for each $\mathbf{w}$. \vspace{0.05in}

\noindent \textbf{Assumption 3 ($\sigma$-bounded Hessians).} The smallest eigenvalue of the Hessian matrix $\nabla^2 F_k$ is $-\sigma$ for each $k$, i.e., $\nabla^2 F_k \succeq -\sigma \mathbf{I}$ for the identity matrix $\mathbf{I}$. This implies that $h_k(\mathbf{w}, \mathbf{w}^t)$ in \eqref{eq:prox} is $\mu'$-strongly convex, where $\mu' = \mu - \sigma$. \vspace{0.05in}

\noindent \textbf{Assumption 4 ($\gamma$-inexact local solvers).} Local updates will yield a $\gamma$-inexact solution $\mathbf{w}_k^{t+1}$ of $\min_\mathbf{w} h_k(\mathbf{w}, \mathbf{w}^t)$ for every $k$ and $t$, i.e., $\norm{\nabla h_k(\mathbf{w}_k^{t+1}, \mathbf{w}^t)} \leq \gamma \norm{\nabla h_k(\mathbf{w}^t, \mathbf{w}^t)}$. We assume that $\gamma$ is in the range $[0,1]$ since $\gamma = 0$ corresponds to solving to optimality, and $\gamma = 1$ happens with the initial parameters $\mathbf{w}_k^{t+1} = \mathbf{w}_k^{t}$ and since the function $h(\mathbf{w}, \mathbf{w}^t)$ is convex, the local optimization algorithm at device should reduce the gradient norm, e.g., gradient descent algorithm.\vspace{0.05in}

In \cite{wang19,tu20}, Assumptions 3\&4 are replaced with a stronger assumption that the $F_k(\mathbf{w})$ are convex. This corresponds to the case where $\sigma \leq 0$ in Assumption 3, meaning $\nabla^2 F_k$ is positive semidefinite, and \fedave{} can be used to minimize the $F_k(\mathbf{w})$ directly without a proximal term. Similar to \cite{tian2019}, the results we derive in this work will more generally hold for non-convex $F_k(\mathbf{w})$, which is true of many ML models today (e.g., neural networks). We also note that \fedprox{} makes a similar assumption to Assumption 4 in deriving its convergence bound \cite{tian2019}, i.e., on the precision of the local solvers. In Section \ref{sec:hetero_folb}, we will present a technique where each device $k$ estimates its own $\gamma_k$ based on its local gradient update.

\textbf{Technical approach:} In the following sections, we first investigate the general non-uniform device selection in federated learning and show that in each round, a device's contribution in reducing the global loss function is bounded by the inner product between its local gradient and the global one. Hence, a near-optimal device selection distribution is introduced, that samples devices according to the inner products between their local and global gradients. Unfortunately, trivial solutions to compute or estimate this distribution are excessively expensive in communication demand. We next introduce \method{} to address this challenge with the core idea of using 2 independent sets of devices, one for estimating the global gradient and another for carrying out local optimization. The locally updated parameters from the second set are then re-weighted by the inner products between their gradients and the estimated global gradient and aggregated to form a new global model. We also analyze the version using a single set of devices and how to handle communication and computation heterogeneity with \method.

% !TEX root = fedlearn_nonuniform.tex

\section{\fednu{}: Non-Uniform Federated Learning}
\label{sec:nufedprox}
In this section, we develop our methodology for improving the convergence speed of federated learning. This includes non-uniform device selection in the local update (Section \ref{ssec:pkt}), and inclusion of gradient information in the aggregation (Section \ref{ssec:gradinfo}). Our theoretical analysis on the expected decrease in loss in each round of learning leads to a selection distribution update that achieves an efficient lower bound (Section \ref{subsec:implication}).
\textcolor{blue}{
\begin{algorithm}[t]
	\caption{Federated learning with non-uniform device selection.}
	\label{alg:fednu_prob}
	\DontPrintSemicolon
	\setstretch{1.2}
	\SetKwInOut{Input}{Input}\SetKwInOut{Output}{output}
	\Input{$K, T, \mu, \gamma, \mathbf{w}^0, N, P^t_k \; k = 1,...,N$}
	%\Output{Array of roots \tcc*[f] {The indexing structure}}
	\For{$t = 0, \dots, T-1$}{
		Server samples (with replacement) a multiset $S_t$ of $K$ devices according to $P^{t}_k, k = 1,...,N$ \\
		Server sends $\mathbf{w}^t$ to all devices $k \in S_t$ \\
		Each device $k \in S_t$ finds a $\mathbf{w}_k^{t+1}$ that is a $\gamma^t_k$-inexact minimizer of $\arg\min_\mathbf{w} h_k(\mathbf{w}, \mathbf{w}^t)$, as defined in \eqref{eq:prox} \\
		Each device $k \in S_t$ sends $\mathbf{w}_k^{t+1}$ back to the server\\
		Server aggregates the $\mathbf{w}_k^{t+1}$ according to $\mathbf{w}^{t+1} = \frac{1}{K}\sum_{k \in S_t} \mathbf{w}^{t+1}_k$
	}
\end{algorithm}
}

\subsection{Non-Uniform Device Selection}
\label{ssec:pkt}
As discussed in Section \ref{ssec:fedalg}, standard federated learning approaches select a set of $K$ devices uniformly at random for local updates in each round. In reality, certain devices will provide better improvements to the global model than others in a round, depending on their local data distributions. If we can estimate the expected decrease in loss each device will provide to the system in a particular round, then the device selections can be made according to those that are expected to provide the most benefit. This will in turn minimize the model convergence time.

%Instead of selecting $K$ devices uniformly at random to calculate gradients and update model parameters, we can deliberately select devices following a different distribution (non-uniform) and opens up the opportunity of deriving a better selection distribution with better loss decreases at every round of optimization.

Formally, we let $P^t_k$ be the probability assigned to device $k$ for selection in round $t$, where $0 \leq P^t_k \leq 1$ and $\sum_{k=1}^{N} P^t_k = 1 \; \forall t$. In our federated learning scheme, during round $t$, the server chooses a multiset $S_t$ of size $K$ by sampling $K$ times from the distribution $P^t_1, ..., P^t_N$. Note that this sampling occurs with replacement, i.e., a device may appear in $S_t$ multiple times and $K$ is the cardinality of this multiset. Each unique $k \in S_t$ then performs a local update on the global model estimate $\mathbf{w}^t$ to find a $\gamma$-inexact minimizer $\mathbf{w}_k^{t+1}$ of $h_k(\mathbf{w}, \mathbf{w}^t)$ in \eqref{eq:prox}, which the server aggregates to form $\mathbf{w}^{t+1}$. Algorithm \ref{alg:fednu_prob} summarizes this procedure, assuming averaging for aggregation; if $k$ appears in $S_t$ more than once, this aggregation effectively places a larger weight on $\mathbf{w}^{t+1}_k$.

Given the introduction of $P^t_k$, we call our methodology \fednu{}, i.e., non-uniform federated learning. A key aspect will be developing an algorithm for $P^t_k$ estimation in each round. The following theorem gives a lower bound on the expected decrease in loss achieved from round $t$ of Algorithm \ref{alg:fednu_prob}, which will assist in this development:

%We modify the \fedprox{} algorithm in \cite{tian2019} to take into accounts a different probability distribution of selecting devices in each update round instead of a trivial uniform distribution. The modified algorithm is presented in Alg.~\ref{alg:fexprox_prob}. 
%Let $P^t_k$ be the probability of selecting device $k$ at update round $t$.

%Based on the above assumptions, we prove the following theorem on the loss decrease at round $t$ of the generalized \fedprox{} algorithm with non-uniform distribution of selecting devices.

\begin{theorem}
	\label{theo:loss}
	With loss functions $F_k$ satisfying Assumptions 1-4, supposing that $\mathbf{w}^t$ is not a stationary solution, in Algorithm \ref{alg:fednu_prob}, the expected decrease in the global loss function satisfies
	\vspace{-0.03in}
	\begin{align}
		\label{eq:loss_thm1}
		&\mathbb{E}[f(\mathbf{w}^{t+1})] \leq f(\mathbf{w}^{t}) - \frac{1}{K\mu} \mathbb{E} \Big[ \sum_{k \in S_t} \langle \nabla f(\mathbf{w}^t),\nabla F_k(\mathbf{w}^t)\rangle \Big] \nonumber \\
		&+ B \Big (\frac{L(\gamma+1)}{\mu \mu'}+\frac{\gamma}{\mu} + \frac{BL(1+\gamma)^2}{2 \mu'^2} \Big) \norm{\nabla f(\mathbf{w}^t)}^2,
	\end{align}
	where $\mu' = \mu - \sigma > 0$, and the expectation $\mathbb{E}$ is with respect to the choice of $K$ devices following probabilities $P^t_k$. As a corollary, after $T$ rounds,
	\vspace{-0.03in}
	\begin{align}
%		\label{eq:loss_convergence}
		&\mathbb{E}[f(\mathbf{w}^{T})] \leq f(\mathbf{w}^{0}) - \frac{1}{K\mu} \mathbb{E} \Big[ \sum_{t = 0}^{T-1} \sum_{k \in S_t} \langle \nabla f(\mathbf{w}^t),\nabla F_k(\mathbf{w}^t)\rangle \Big] \nonumber \\
		&+ B \Big (\frac{L(\gamma+1)}{\mu \mu'}+\frac{\gamma}{\mu} + \frac{BL(1+\gamma)^2}{2 \mu'^2} \Big) \sum_{t = 0}^{T-1} \norm{\nabla f(\mathbf{w}^t)}^2, \nonumber
	\end{align}
	where the expectation is with respect to the random selections of $S_0, S_1, \dots, S_{T-1}$.
\end{theorem}

The full proof of Theorem~\ref{theo:loss} as well as proofs of later theorems/propositions are presented in appendix.

Theorem \ref{theo:loss} provides a bound on how rapidly the global loss can be expected to improve in each iteration based on the selection of devices in Algorithm \ref{alg:fednu_prob}. It shows a dependency on parameters $L$, $B$, $\gamma$, and $\mu$ of the ML model. In particular, we see that $\mathbb{E}[f(\mathbf{w}^{t+1})] \propto B^2$, meaning that as the dissimilarity between local and global model gradients grows larger, the bound weakens. Intuitively, $B$ depends on the variance between local data distributions: as the datasets $D_k$ approach being independent and identically distributed (i.i.d.) across $k$, the gradients will become more similar, and $B$ will approach $1$. As they become less i.i.d., however, the gradients will diverge, and $B$ will increase. Hence, Theorem \ref{theo:loss} gives quantitative insight into the effect of data heterogeneity on federated learning convergence.

Compared to the bound of \fedprox{} \cite{tian2019}, which was shown to work on the particular uniform distribution, our result in Theorem~\ref{theo:loss} is more general and applicable for any given probability distribution. Moreover, our  result offers a new approach to optimize convergence rate through maximizing the inner product term $\mathbb{E} \big[\sum_{k \in S_t}\langle \nabla f(\mathbf{w}^t),\nabla F_k(\mathbf{w}^t)\rangle \big]$. The proof of Theorem~\ref{theo:loss} also takes a different path compared to that of \fedprox{} in \cite{tian2019}, which relies on the uniform distribution to first establish intermediate relations of $f(\mathbf{w}^{t+1})$ and $f(\mathbf{w}^t)$ with $f(\mathbf{\bar w}^{t+1})$, where $\mathbf{\bar w}^{t+1} = \frac{1}{N}\sum_{k =1}^N \mathbf{w}_k^{t+1}$, and then connects $f(\mathbf{w}^{t+1})$ with $f(\mathbf{w}^t)$. Our result applies for any distribution and thus required direct proof of the relation between $f(\mathbf{w}^{t+1})$ and $f(\mathbf{w}^t)$ via bounding each of the terms given by the $L$-Lipschitz continuity of $f$.
\vspace{-0.05in}

\subsection{Aggregation with Gradient Information}
\label{ssec:gradinfo}
An immediate suggestion from the expectation term in Theorem~\ref{theo:loss} is that any devices which have a negative inner product $\langle \nabla f(\mathbf{w}^t),\nabla F_k(\mathbf{w}^t)\rangle < 0$ between their gradients $\nabla F_k(\mathbf{w}^t)$ and the global gradient $\nabla f(\mathbf{w}^t)$ would actually hurt model performance. This is due to the averaging technique used for model aggregation in Algorithm \ref{alg:fednu_prob}, which is common in federated learning algorithms due to its simplicity \cite{wang19,tian2019,tu20}. It is consistent with the characteristics of distributed gradient descent \cite{li14,richtarik16}, where the global gradient (i.e., across the entire dataset) can reduce the overall loss while individual local gradients (i.e., at individual devices) that are not well aligned with the global objective -- in this case, those with negative inner product -- will not help improve the overall loss.

If we assume the server can estimate when a device's inner product is negative, then we can immediately improve \fednu{} with an aggregation rule of
\vspace{-0.03in}
\begin{align}
	\mathbf{w}^{t+1} = \mathbf{w}^t + \frac{1}{K}\sum_{k \in S_t} \sign(\langle \nabla f(\mathbf{w}^t),\nabla F_k(\mathbf{w}^t)\rangle) (\mathbf{w}^{t+1}_k - \mathbf{w}^t)
	\label{eq:new_update}
\end{align}
in Algorithm \ref{alg:fednu_prob} based on the signum function. This negates local updates from devices in $S_t$ that have $\langle \nabla f(\mathbf{w}^t),\nabla F_k(\mathbf{w}^t)\rangle < 0$, and provides a stronger lower-bound than given in Theorem \ref{theo:loss}:

%Given all the gradients, $\nabla F_k(\mathbf{w}^t), \forall k= 1..N$, we can immediately improve \fedprox{} with a stronger lower-bound than that of Theorem~\ref{theo:loss} by negating the effect of updated parameters from devices with $\langle \nabla f(\mathbf{w}^t),\nabla F_k(\mathbf{w}^t)\rangle < 0$. In other words, we can replace the simple averaging with the following aggregation rule:
%\begin{align}
%	\mathbf{w}^{t+1} = \mathbf{w}^t + \frac{1}{K}\sum_{k \in S^{t}} \sign(\langle \nabla f(\mathbf{w}^t),\nabla F_k(\mathbf{w}^t)\rangle) (\mathbf{w}^{t+1}_k - \mathbf{w}^t).
%	\label{eq:new_update}
%\end{align}
%Then, we obtain the following better lower-bound in term of loss decrease in a round $t$:
\begin{proposition}
	\label{theo:loss_improved}
	With the same assumptions on $F_k$ and $\mathbf{w}^t$ as in Theorem~\ref{theo:loss}, with \eqref{eq:new_update} used as the aggregation rule in Algorithm~\ref{alg:fednu_prob} (Line~6), the expected decrease in the global loss satisfies
	\vspace{-0.03in}
	\begin{align}
		\mathbb{E} & [f(\mathbf{w}^{t+1})] \leq f(\mathbf{w}^{t}) - \frac{1}{K\mu} \mathbb{E} \Big[ \sum_{k \in S_t} |\langle \nabla f(\mathbf{w}^t),\nabla F_k(\mathbf{w}^t)\rangle| \Big] \nonumber \\
		& + B \Bigg (\frac{L(\gamma+1)}{\mu \mu'}+\frac{\gamma}{\mu} + \frac{BL(1+\gamma)^2}{2 \mu'^2} \Bigg) \norm{\nabla f(\mathbf{w}^t)}^2.
	\end{align}
\end{proposition}

Proposition~\ref{theo:loss_improved} is clearly stronger than Theorem~\ref{theo:loss}: by incorporating gradient information, the inner products are replaced with their absolute values, making the expected decrease in loss faster. We will next propose a method for setting the selection probabilities $P^t_k$ to optimize this bound, and then develop algorithms to estimate the inner products. 
\vspace{-0.05in}

\subsection{\lb-Near-Optimal Device Selection}
\label{subsec:implication}
The set $S_t$ of selected devices affects Theorem \ref{theo:loss_improved} through the expectation $\mathbb{E} [\sum_{k \in S_t} | \langle \nabla f(\mathbf{w}^t),\nabla F_k(\mathbf{w}^t)\rangle |]$. To maximize the convergence speed, we seek to minimize the upper bound on the loss update in each round $t$, which corresponds to the following optimization problem for choosing $S_t$:
\begin{equation*}
\begin{aligned}
& \underset{P^t_k}{\text{maximize}}
& & \mathbb{E} \Big[ \sum_{k \in S_t} | \langle \nabla f(\mathbf{w}^t),\nabla F_k(\mathbf{w}^t)\rangle | \Big] \\[0.2em]
& \text{subject to}
& & \sum_k P^t_k = 1, \quad P^t_k \geq 0 \;\; \forall k.
\end{aligned}
\end{equation*}
\vspace{-0.1in}

\nonumber This problem is difficult to solve analytically given the sampling relationship between $S_t$ and $P^t_k$.\footnote{Formally, the probability mass function of $S_t$ is formed from $K$ repeated trials of the $N$-dimensional categorical distribution \cite{bishop2006pattern} over $P^t_1, ..., P^t_N$.} It is clear, however, that the solution which maximizes this expectation will assign higher probability of being selected to devices with higher inner product $|\langle \nabla f(\mathbf{w}^t),\nabla F_k(\mathbf{w}^t)\rangle|$. A natural candidate which satisfies this criterion is $P^{t}_k \propto |\langle \nabla f(\mathbf{w}^t),\nabla F_k(\mathbf{w}^t)\rangle|$. We call this distribution \lb-near-optimal, i.e., near-optimal lower-bound, formally defined as follows:

%Theorem~\ref{theo:loss_improved} provides a lower-bound on the loss decrease after every round of optimization. More importantly, this bound depends only on the distribution of selecting devices, i.e. $-\frac{1}{K \mu}E[\sum_{k \in S_t} \langle \nabla f(\mathbf{w}^t),\nabla F_k(\mathbf{w}^t)\rangle]$, plus a fixed amount that depends on a particular round. Thus, we can effectively maximize $E[\sum_{k \in S_t} |\langle \nabla f(\mathbf{w}^t),\nabla F_k(\mathbf{w}^t)\rangle|]$ in order to maximize the loss decrease in every round, leading to a fast-convergence algorithm.
\begin{definition}[\lb-near-optimal selection distribution]
	\label{def:lb_near_optimal}
	The selection distribution ${P_{\mathsf{lb}}}^{t}_k$ achieving a near-optimal lower-bound on loss decrease in Theorem~\ref{theo:loss_improved} is called the \lb-near-optimal selection distribution, and has the form
	\begin{align}
	\label{eq:plb}
		{P_{\mathsf{lb}}}^{t}_k = \frac{|\langle \nabla f(\mathbf{w}^t),\nabla F_k(\mathbf{w}^t)\rangle|}{\sum_{k' = 1}^{N} |\langle \nabla f(\mathbf{w}^t),\nabla F_{k'}(\mathbf{w}^t)\rangle|},
	\end{align}
	with the corresponding lower bound of expected loss being
	\vspace{-0.05in}
	\begin{align}
	\label{eq:lb-plb}
		&\mathbb{E}[f(\mathbf{w}^{t+1})] \leq f(\mathbf{w}^{t}) - \frac{1}{\mu}\sum_{k = 1}^{N} |\langle \nabla f(\mathbf{w}^t),\nabla F_k(\mathbf{w}^t)\rangle| {P_{\mathsf{lb}}}^t_k \nonumber \\
		& + B \Bigg (\frac{L(\gamma+1)}{\mu \mu'}+\frac{\gamma}{\mu} + \frac{BL(1+\gamma)^2}{2 \mu'^2} \Bigg) \norm{\nabla f(\mathbf{w}^t)}^2.
	\end{align}
\end{definition}

\textbf{Comparison to FedProx \cite{tian2019}:} Our lower bound in (\ref{eq:lb-plb}) of Definition~\ref{def:lb_near_optimal}, corresponding to the near-optimal device selection distribution and achieved by our proposed algorithm \method{} in Section~\ref{sec:folb}, is more general than the bound of \fedprox{} in \cite{tian2019}, which is restricted to the uniform distribution. Specifically, our bound in (\ref{eq:lb-plb}) is stronger if 
\vspace{-0.05in}
\begin{align*}
	\frac{1}{\mu}\sum_{k = 1}^{N} |\langle \nabla f(\mathbf{w}^t),\nabla F_k(\mathbf{w}^t)\rangle| {P_{\mathsf{lb}}}^t_k \geq \Big(\frac{1}{\mu} - \frac{B(1+\gamma)\sqrt{2}}{\mu' \sqrt{K}} \nonumber \\
	- \frac{LB^2(1+\gamma)^2}{\mu'^2K} (2\sqrt{2K}+2) \Big) \norm{\nabla f(\mathbf{w}^t)}^2,
\end{align*}
which holds since
\vspace{-0.05in}
\begin{align*}
	& \frac{1}{\mu}\sum_{k = 1}^{N} |\langle \nabla f(\mathbf{w}^t),\nabla F_k(\mathbf{w}^t)\rangle| {P_{\mathsf{lb}}}^t_k \nonumber \\
	& = \frac{1}{\mu} \frac{\sum_{k = 1}^{N} |\langle \nabla f(\mathbf{w}^t),\nabla F_k(\mathbf{w}^t)\rangle|^2}{\sum_{k' = 1}^{N} |\langle \nabla f(\mathbf{w}^t),\nabla F_{k'}(\mathbf{w}^t)\rangle|} \\
	& \geq \frac{1}{\mu}\frac{1}{N} \sum_{k = 1}^{N} |\langle \nabla f(\mathbf{w}^t),\nabla F_k(\mathbf{w}^t)\rangle| \qquad (\text{Cauchy-Schwarz}) \\
	& \geq \frac{1}{\mu} | \frac{1}{N}\sum_{k = 1}^{N} \langle \nabla f(\mathbf{w}^t),\nabla F_k(\mathbf{w}^t)\rangle| \qquad (\text{triangle inequality}) \\
	& \geq \frac{1}{\mu} |  \langle \nabla f(\mathbf{w}^t),\nabla f(\mathbf{w}^t)\rangle| = \frac{1}{\mu} \norm{\nabla f(\mathbf{w}^t)}^2.
\end{align*}
The last inequality holds due to $f(\mathbf{w}^t) = \frac{1}{N}\sum_{k = 1}^{N} F_k(\mathbf{w}^t)$.

\textbf{Convergence property.} Starting from the lower-bound in (\ref{eq:lb-plb}), we can show the convergence rate in the form of  the gradient converging to zero when the parameter settings satisfy certain constraints, similarly to \cite{tian2019}. Furthermore, since the bound in (\ref{eq:lb-plb}) is stronger than that of \fedprox{} in \cite{tian2019}, the corresponding convergence rate is also faster. Specifically, applying (\ref{eq:lb-plb}) for all $t = 0, \dots, T$ gives us a series of inequalities, and taking the sum of these yields the desired form of gradient convergence (see \cite{tian2019} for more details).

In Definition 1, the expectation term in the bound on $\mathbb{E}[f(\mathbf{w}^{t+1})]$ has been computed in terms of the selection distribution ${P_{\mathsf{lb}}}^{t}_k$. Unfortunately, the values of $\langle \nabla f(\mathbf{w}^t), \nabla F_k(\mathbf{w}^t)\rangle$ needed to compute the ${P_{\mathsf{lb}}}^{t}_k$ cannot be evaluated at the server at the beginning of round $t$, since the local and global gradients are not available at the time of device selection. In the rest of this section, and in Section \ref{sec:folb}, our goal will be to develop a federated learning algorithm that (i) achieves the performance of the distribution in Definition 1, i.e., provides the same loss decrease at every round, and (ii) results in an efficient implementation in a client-server network architecture. We refer to such an algorithm as an \lb-near-optimal-efficient federated learning algorithm:

\begin{definition}[\lb-near-optimal-efficient federated learning algorithm]
	\label{def:lp_optimality}
	An iterative federated learning algorithm is called \lb-near-optimal-efficient if it achieves the near-optimal lower-bound of loss decrease in Definition~\ref{def:lb_near_optimal}, which corresponds to the near-optimal selection distribution at every round, and does not require communication between devices that is significantly more expensive than standard federated learning.
\end{definition}

\subsection{Naive Algorithms for Fast Convergence}
We first present two algorithms that are straightforward modifications of the methods described in this section towards the goal of satisfying Definition 2. We will see that each of these fails to satisfy one criterion in Definition 2,  however, motivating our main algorithms in Section~\ref{sec:folb}.

\subsubsection{Direct computation of \lb-near-optimal distribution}
The most straightforward approach to achieving \lb-near-optimality is enabling computation of the \lb-near-optimal distribution ${P_{\mathsf{lb}}}^t_k$ at the beginning of round $t$ and using this to sample devices. This approach requires the server to send $\mathbf{w}^t$ to all $N$ devices, have them compute $\nabla F_k(\mathbf{w}^t)$, and then send it back to the central server. With these values, the server can exactly calculate the \lb-near-optimal distribution through \eqref{eq:plb}.

Clearly, this algorithm will obtain the \lb-near-optimal distribution, leading to a fast convergence rate (assuming that this initial round of communication does not significantly increase the time of each round $t$). However, this algorithm requires one iteration of expensive communication between the server and all $N$ devices. The gradient $\nabla F_k(\mathbf{w}^t)$ is the same dimension as $\mathbf{w}^t$, and the purpose of algorithms like \fedave{} and \fedprox{} selecting $K$ of $N$ devices is to avoid this excessive communication between a server and edge devices in contemporary network architectures \cite{tu20}. %which is normally unacceptable in real-world scenarios since there are possibly millions or even billions of devices and requiring all devices' gradients may take prohibitively long time. 

As an aside, if we were able to afford this extra communication of gradients in each round, then why not just carry out the exact (centralized) gradient descent at the server? Federated learning would still be beneficial in this scenario for two reasons. First, during their local updates, each device usually carries out multiple iterations of gradient descent, saving potentially many more rounds of gradient communication to/from the server \cite{wang19}. Second, while batch gradient descent converges slowly, federated learning has a flavor of stochastic gradient descent which tends to converge faster \cite{lin18}.

%\begin{algorithm}[H]
%	\caption{\fedprox{} with non-uniform distribution of selecting devices}
%	\label{alg:fexprox_prob_lb1}
%	
%	\footnotesize
%	\DontPrintSemicolon
%	
%	\SetKwInOut{Input}{Input}\SetKwInOut{Output}{output}
%	\Input{$K, T, \mu, \gamma, \mathbf{w}^0, N, P^t_k, k = 1..N$}
%	%\Output{Array of roots \tcc*[f] {The indexing structure}}
%	
%	\For{$t = 0, \dots, T-1$}{
%		Server selects a subset $S_t$ of $K$ devices following $P^t_k, k = 1..N$\\
%		Server sends $\mathbf{w}^t$ to all chosen devices \\
%		Each chosen device $k \in S_t$ finds a $\mathbf{w}_k^{t+1}$ which is a $\gamma^t_k$-inexact minimizer of: $\mathbf{w}_k^{t+1} \approx \arg\min_\mathbf{w} h_k(\mathbf{w};\mathbf{w}^t) = F_k(\mathbf{w}) + \frac{\mu}{2}\norm{\mathbf{w}-\mathbf{w}^t}^2$\\
%		Each device $k \in S_t$ sends $\mathbf{w}_k^{t+1}$ back to the server\\
%		Server aggregates all the $\mathbf{w}_k^{t+1}, k = 1..K$ as $\mathbf{w}^{t+1} = \frac{1}{K}\sum_{k \in S_t} \mathbf{w}^{t+1}_k$
%	}
%\end{algorithm}

\subsubsection{Sub-optimal estimation of \lb-near-optimal distribution}
A possible workaround for the issue of expensive communication in the first approach is to further upper bound $|\langle \nabla f(\mathbf{w}^t),\nabla F_k(\mathbf{w}^t)\rangle| \leq \norm{\nabla f(\mathbf{w}^t)} \norm{\nabla F_k(\mathbf{w}^t)}$ using the Cauchy-Schwartz inequality. Since $\norm{\nabla f(\mathbf{w}^t)}$ is the same for all the devices, we can take $P^t_k \propto \norm{\nabla F_k(\mathbf{w}^t)}$. Hence, while this approach still requires the server to send out $\mathbf{w}^t$ to all devices for them to compute gradients, each device $k$ only needs to send back a single number, $\norm{\nabla F_k(\mathbf{w}^t)}$. This is much less expensive given the fact that edge devices tend to have larger download than upload capacities, typically by an order of magnitude \cite{tu20}. 
%This algorithm is described in Algorithm~\ref{alg:fexprox_prob_lb2}.

%Then, it's more efficient for devices to download $\mathbf{w}^t$ from server and update $\norm{\nabla F_k(\mathbf{w}^t)}$ to server at the beginning of each round. 

%\begin{algorithm}[t]
%	\caption{\fednu{} with sub-optimal estimation of the \lb-near-optimal selection distribution.}
%	\label{alg:fexprox_prob_lb2}
%
%	\setstretch{1.2}
%		
%	\DontPrintSemicolon
%	
%	\SetKwInOut{Input}{Input}\SetKwInOut{Output}{output}
%	\Input{$K, T, \mu, \gamma, \mathbf{w}^0, N$}
%	%\Output{Array of roots \tcc*[f] {The indexing structure}}
%	
%	\For{$t = 0, \dots, T-1$}{
%		Each device $k = 1,...,N$ downloads $\mathbf{w}^t$ from server, calculates $\nabla F_k(\mathbf{w}^t)$, and uploads $\norm{\nabla F_k(\mathbf{w}^t)}$ \\
%		Server computes selection distribution $P^t_k = \norm{\nabla F_k(\mathbf{w}^t)} / \sum_{k' = 1}^N \norm{\nabla F_{k'}(\mathbf{w}^t)}, k = 1,...,N$ \\
%		Server samples a multiset $S_t$ of $K$ devices according to the $P^t_k, k = 1,...,N$ \\
%		Each chosen device $k \in S_t$ finds a $\mathbf{w}_k^{t+1}$ that is a $\gamma^t_k$-inexact minimizer of $\arg\min_\mathbf{w} h_k(\mathbf{w}, \mathbf{w}^t)$, as defined in \eqref{eq:prox} \\
%		Each device $k \in S_t$ sends $\mathbf{w}_k^{t+1}$ back to the server\\
%		Server aggregates the $\mathbf{w}_k^{t+1}$ according to $\mathbf{w}^{t+1} = \frac{1}{K}\sum_{k \in S_t} \mathbf{w}^{t+1}_k$
%	}
%\end{algorithm}

%While Algorithm \ref{alg:fexprox_prob_lb2} 
While this algorithm is closer to the communication efficiency of standard federated learning algorithms, there is no guarantee on how accurately $\norm{\nabla f(\mathbf{w}^t)} \norm{\nabla F_k(\mathbf{w}^t)}$ approximates $|\langle \nabla f(\mathbf{w}^t),\nabla F_k(\mathbf{w}^t)\rangle|$, which could result in an inaccurate estimate of ${P_{\mathsf{lb}}}^{t}_k$. Thus, 
%Algorithm \ref{alg:fexprox_prob_lb2} 
it may not satisfy the \lb-near-optimal criteria of Definition~\ref{def:lp_optimality}.

%The advantage of this approach is saving the practically expensive upload of local gradients in possibly very high dimension. The biggest issue with this approach is that the upper-bound $|\langle \nabla f(\mathbf{w}^t),\nabla F_k(\mathbf{w}^t)\rangle| \leq \norm{\nabla f(\mathbf{w}^t)} \norm{\nabla F_k(\mathbf{w}^t)}$ may be too loose and result in inaccurate estimate of $P^t_k$.
\begin{figure}[t]
	%\vspace{-0.05in}
	\centering
	\begin{subfigure}[b]{0.50\linewidth}
		\centering
		\includegraphics[width=\linewidth]{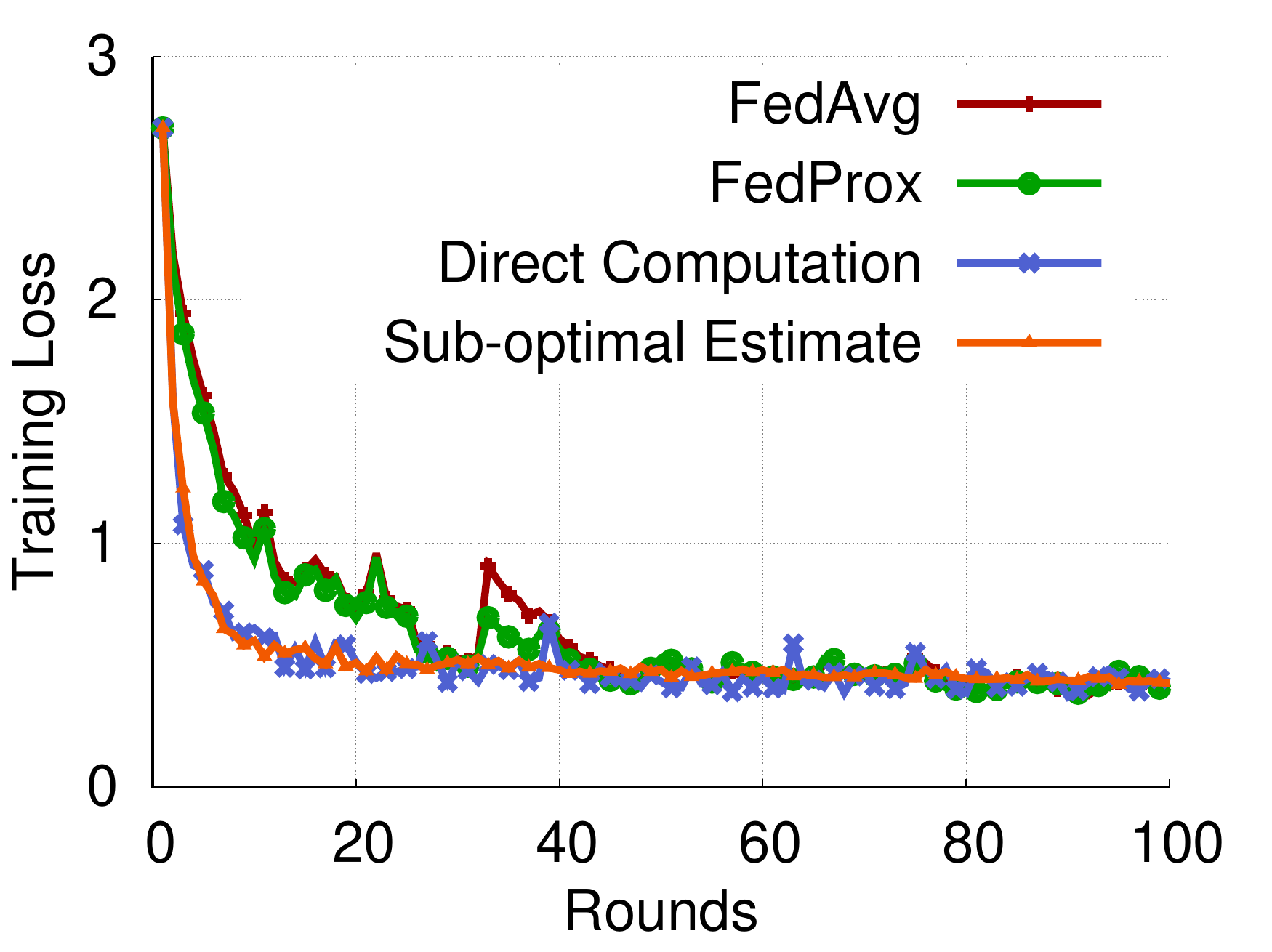}
		\caption{Training loss}
	\end{subfigure}
	\hspace{-0.2in}
	\begin{subfigure}[b]{0.50\linewidth}
		\centering
		\includegraphics[width=\linewidth]{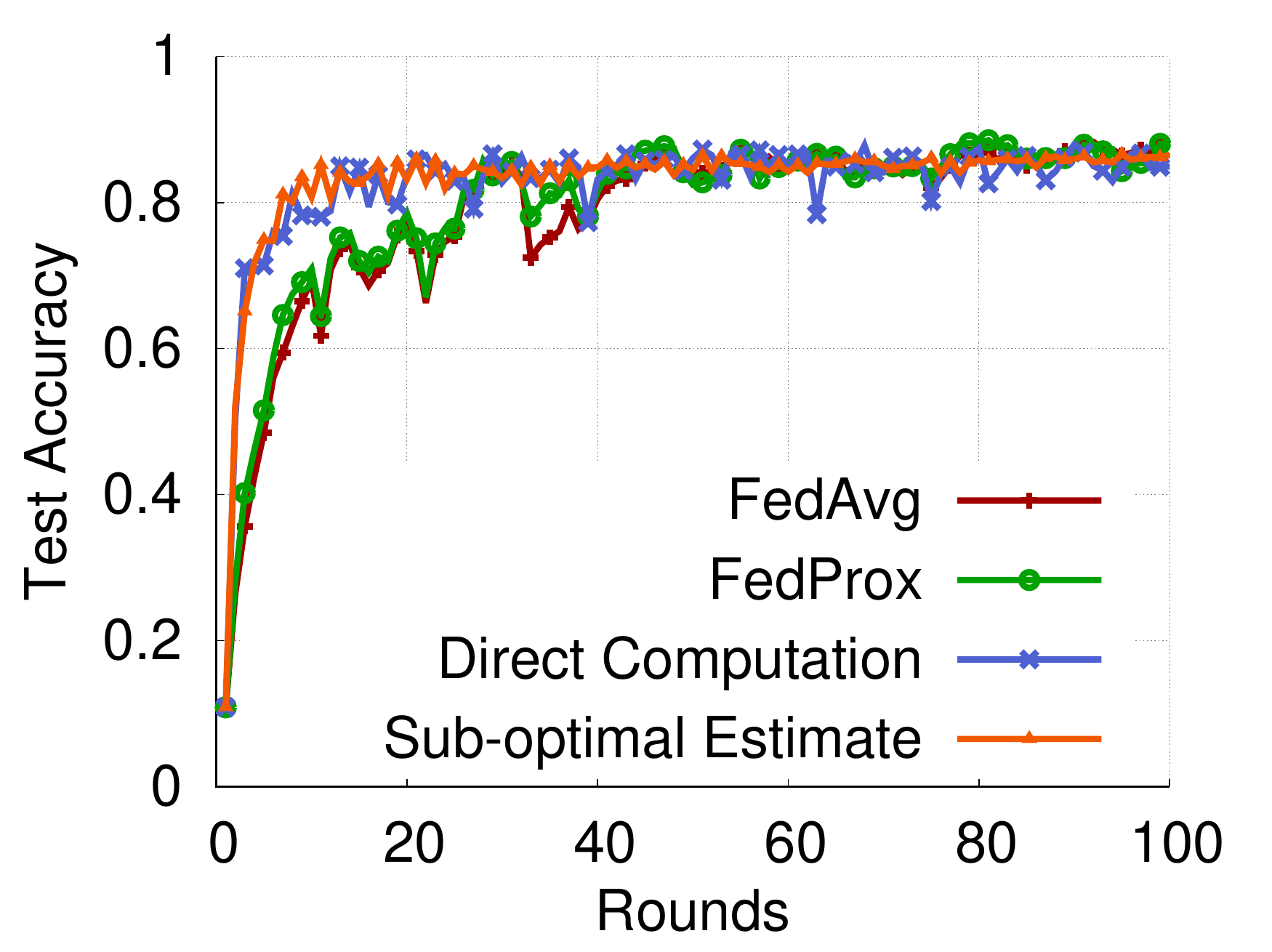}
		\caption{Test accuracy}
	\end{subfigure}
	\caption{Training loss and test accuracy of our motivating idea and state-of-the-art approaches on MNIST dataset ($\mu = 1$, see Sec.~\ref{sec:exps} for details on experimental settings).}
	\vspace{-0.1in}
	\label{fig:loss_acc_fednu}
\end{figure}

We demonstrate the better performance when using directly or estimating the \lb-near-optimal selection distribution than existing state-of-the-art federated learning algorithms in Fig.~\ref{fig:loss_acc_fednu}. Here we run the above two naive algorithms targeting the \lb-near-optimal distribution along with \fedave{} and \fedprox{}, and observe significant improvements over both \fedave{} and \fedprox{} in terms of convergence speed. Our methods quickly converge after only a few rounds of communication. This motivates our proposed algorithm, \method{}, which also targets the \lb-near-optimal distribution, however, removes the communication burden in the naive algorithms.

% !TEX root = fedlearn_nonuniform.tex

\section{\method: An \lb-near-optimal-efficient Federated Learning Algorithm}
\label{sec:folb}
As discussed in Section \ref{subsec:implication}, the \lb-near-optimal selection distribution given in Definition 1 for maximizing the loss decrease in round $t$ cannot be computed by the server at the beginning of round $t$, since it involves all local gradients of the current global estimate $\mathbf{w}^t$. The straightforward approximation using Cauchy-Schwartz still requires one iteration of additional communication where the server sends $\mathbf{w}^t$ to all devices, and does not guarantee \lb-near-optimality. With the goals of fast convergence and low communication overhead in mind, the challenges we face in developing an \lb-near-optimal-efficient federated learning algorithm for \fednu{} described in Definition 2 are two-fold:
\begin{itemize}
	\item[(1)] How can we accurately estimate (preferably with performance guarantees) the \lb-near-optimal probability distribution without involving all local gradients?
	\item[(2)] How can we obtain this estimate efficiently, i.e., with minimal communication overhead on top of standard federated learning algorithms? 
\end{itemize}

\begin{algorithm}[t]
	\caption{\method{} algorithm for \lb-near-optimal-efficient federated learning.}
	\label{alg:fexprox_prob_lb3}
	
	\DontPrintSemicolon
	
	\setstretch{1.2}
	\SetKwInOut{Input}{Input}\SetKwInOut{Output}{output}
	\Input{$K, T, \mu, \gamma, \mathbf{w}^0, N$}
	%\Output{Array of roots \tcc*[f] {The indexing structure}}
	
	\For{$t = 0, \dots, T-1$}{
		Server selects two multisets $S^{t}_1$ and $S^{t}_2$ each of $K$ devices uniformly at random \\
		Server sends $\mathbf{w}^t$ to all $k \in S^{t}_1$ and $k' \in S^{t}_2$ \\
		\For{each device $k \in S^{t}_1$}{
			Device $k$ computes its gradient $\nabla F_k(\mathbf{w}^t)$ \\
			Device $k$ sends $\nabla F_k(\mathbf{w}^t)$ back to the server \\
			Device $k$ finds a $\gamma^t_k$-inexact minimizer of $\arg\min_\mathbf{w} h_k(\mathbf{w}, \mathbf{w}^t)$, as defined in \eqref{eq:prox} \\
			Device $k$ sends $\mathbf{w}_k^{t+1}$ back to the server}
		\For{each device $k' \in S^{t}_2$}{
			Device $k'$ computes its gradient $\nabla F_{k'}(\mathbf{w}^t)$ \\
			Device $k'$ sends $\nabla F_{k'}(\mathbf{w}^t)$ back to the server}
		Server computes $\nabla_1 f(\mathbf{w}^t), \nabla_2 f(\mathbf{w}^t)$ according to \eqref{eq:gradest} and aggregates the $\mathbf{w}_k^{t+1}$ \eqref{eq:aggregate}
	}
\end{algorithm}

In this section, we develop a federated learning algorithm called \method{} (Section \ref{ssec:folb}) that addresses these challenges. The key idea of \method{} is a novel calibration procedure for aggregating local model updates from devices selected uniformly at random. This calibration weighs the updates received by their estimated importance to the model, which we show matches the performance of Theorem \ref{theo:loss_improved} (Section \ref{ssec:folb_proof}). We also demonstrate a technique to further optimize the communication demand of \method{} (Section \ref{ssec:folb_comm}).

\subsection{Proposed \method{} Algorithm}
\label{ssec:folb}
The \method{} algorithm is summarized in Algorithm \ref{alg:fexprox_prob_lb3}. At the start of round $t$, the server selects two multisets $S^t_1$ and $S^t_2$ of devices of size $K$ uniformly at random, and sends $\mathbf{w}^t$ to each $k \in S^t_1$ and $k' \in S^t_2$. Each $k \in S^t_1$ computes its $\gamma^t_k$-inexact local update $\mathbf{w}^{t+1}_k$, sending both $\mathbf{w}^{t+1}_k$ and $\nabla F_k(\mathbf{w}^t)$ back to the server. Each $k' \in S^t_2$, by contrast, only computes $\nabla F_{k'}(\mathbf{w}^t)$ and sends this back, for the purpose of calibrating the updates. Then, instead of simple averaging, the server aggregates the received update parameters according to the following rule:
\begin{align}
	\label{eq:aggregate}
	\mathbf{w}^{t+1} = \mathbf{w}^{t} + \sum_{k \in S^{t}_1} \frac{\langle \nabla F_k(\mathbf{w}^t), \nabla_1 f(\mathbf{w}^t) \rangle }{\sum_{k' \in S^{t}_2} \langle \nabla F_{k'}(\mathbf{w}^t), \nabla_2 f(\mathbf{w}^t) \rangle } \Delta \mathbf{w}^{t+1}_k,
\end{align}
\vspace{-0.25in}

\noindent where
\begin{equation}
	\label{eq:gradest}
	\nabla_i f(\mathbf{w}^t) = \frac{1}{K}\sum_{k \in S^t_i} \nabla F_k(\mathbf{w}^t),
\end{equation}
is the gradient of the global loss $f(\mathbf{w}^t)$ estimated from the local losses across devices in $S^t_i, i \in \{1, 2\}$, and $\Delta \mathbf{w}^{t+1}_k = \mathbf{w}^{t+1}_k - \mathbf{w}^{t}$ is the change that device $k \in S^t_1$ made to $\mathbf{w}^{t}$ at round $t$ during its local update.

The intuition behind \eqref{eq:aggregate} is that the local update of each device $k \in S^t_1$ is weighted by a measure of how correlated its gradient $\nabla F_k(\mathbf{w}^t)$ is with the global gradient $\nabla f(\mathbf{w}^t)$. This correlation is assessed relative to $\nabla_1 f(\mathbf{w}^t)$, which is an unbiased estimate of $\nabla f(\mathbf{w}^t)$ using gradient information obtained from $S^t_1$. The weights are normalized relative to a second unbiased estimate of total correlation among $K$ devices, obtained over $S^t_2$.

%The core of \method{} is a weighted averaging scheme of updated parameters from local devices in each round. It takes into account the importance of each local update with respect to the overall objective of minimizing the global cost function $f(\mathbf{w}) = \frac{1}{N}\sum_{k = 1}^N F_k(\mathbf{w})$ distributed across all devices.
%\textbf{Proposed aggregation of locally updated model parameters}: Instead of using a simple averaging scheme as in all previous works on federated learning, we aggregate the received updated parameters from devices using the following rule:

\subsection{Proof of \lb-Near-Optimality}
\label{ssec:folb_proof}
We now prove that \method{} obtains the same lower-bound of loss decrease at every round as the \lb-near-optimal selection distribution. In particular, we have the following theorem:
\begin{theorem}
	\label{theo:loss_lb}
	In Algorithm \ref{alg:fexprox_prob_lb3}, with the same assumptions on $F_k$ and $\mathbf{w}^t$ as in Theorem \ref{theo:loss}, the lower-bound achieved on the expected decrease of the global loss in round $t$ matches \eqref{eq:lb-plb}, i.e., the \lb-near-optimal selection probability distribution.
%	\begin{align}
%		E[f(\mathbf{w}^{t+1})] \leq f(\mathbf{w}^{t}) - \frac{1}{\mu}\sum_{k = 1}^{N} |\langle \nabla f(\mathbf{w}^t),\nabla F_k(\mathbf{w}^t)\rangle| {P_{\mathsf{lb}}}^t_k \nonumber \\
%		\quad + B \Bigg (\frac{L(\gamma+1)}{\mu \mu'}+\frac{\gamma}{\mu} + \frac{BL(1+\gamma)^2}{2 K\mu'^2} \Bigg) \norm{\nabla f(\mathbf{w}^t)}^2.
%	\end{align}
\end{theorem}

The following lemma provide a key insight into how $S_1^t$ and $S_2^t$ can be used to estimate the global gradient when computing the inner products with local gradients, and will help in proving Theorem \ref{theo:loss_lb} in Appendix~\ref{proof:theo_loss_lb}.
\begin{lemma}
	\label{lem:square}
	Let $\nabla_if(\mathbf{w}^t)$ be defined as in \eqref{eq:gradest}. Then,
	\begin{align}
		\label{lem:s1}
		\mathbb{E} \Big[ \sum_{k \in S^{t}_1} &\langle \nabla F_k(\mathbf{w}^t), \nabla_1 f(\mathbf{w}^t) \rangle^2 \Big] \nonumber \\ &= \frac{K}{N} \sum_{k = 1}^{N} \langle \nabla f(\mathbf{w}^t),\nabla F_k(\mathbf{w}^t)\rangle^2,
	\end{align}
	\vspace{-0.15in}
	
	\noindent and
	\begin{align}
		\label{lem:s2}
		\mathbb{E} \Big[ \sum_{k' \in S^{t}_2} &\langle \nabla F_{k'}(\mathbf{w}^t), \nabla_2 f(\mathbf{w}^t) \rangle \Big] \nonumber \\ & \leq \frac{K}{N} \sum_{k' = 1}^{N} |\langle \nabla f(\mathbf{w}^t),\nabla F_{k'}(\mathbf{w}^t)\rangle|.
	\end{align}
%	\begin{align}
%		&\mathbb{E} \Bigg[ \sum_{k \in S^{t}_1} \langle \nabla F_k(\mathbf{w}^t), \nabla_1 f(\mathbf{w}^t) \rangle^2 \Bigg] \nonumber \\
%		&\quad \quad = \frac{K}{N} \sum_{k = 1}^{N} \langle \nabla f(\mathbf{w}^t),\nabla F_k(\mathbf{w}^t)\rangle^2 .
%	\end{align}
\end{lemma}

\subsection{Optimizing \method{} Communication Efficiency}
\label{ssec:folb_comm}
Theorem \ref{theo:loss_lb} establishes the \lb-near-optimal property of \method{}. Algorithm \ref{alg:fexprox_prob_lb3} does, however, call for local updates from $2K$ devices across the two sets $S^t_1$ and $S^t_2$ in each round (and for $S^t_1$, communication of both the updates and the gradients), whereas standard federated learning algorithms only sample $K$ devices.

To reduce the communication demand further, we can make two practical adjustments to Algorithm \ref{alg:fexprox_prob_lb3}. First, we can set $S^t_1 = S^t_2$ in each round, i.e., only selecting one set of $K$ random devices and using the received gradients both for parameter updates and for normalizing the weights on these updates, dropping the total to $K$. Second, similar to the technique in Section \ref{ssec:gradinfo}, rather than discarding updates from devices with $\langle \nabla F_k (\mathbf{w}^t), \nabla_1 f(\mathbf{w}^t) \rangle < 0$, we can aggregate the negatives of their $\Delta \mathbf{w}^{t+1}_k$, thereby leveraging all $K$. Our modified aggregation rule becomes
\begin{align}
	\label{eq:aggregate_2}
	\mathbf{w}^{t+1} = \mathbf{w}^{t} + \sum_{k \in S^{t}_1} \frac{\langle \nabla F_k(\mathbf{w}^t), \nabla_1 f(\mathbf{w}^t) \rangle} {\sum_{k' \in S^{t}_1} | \langle \nabla F_{k'}(\mathbf{w}^t), \nabla_1 f(\mathbf{w}^t) \rangle| } \Delta \mathbf{w}^{t+1}_k.
\end{align}
\vspace{-0.25in}

\noindent A key step in the proof of Theorem \ref{theo:loss_lb}, for \eqref{eq:lb_loss_3}, relied on the independence between sampling $S^t_1$ and $S^t_2$. With $S^t_1 = S^t_2$, this clearly no longer holds. Instead, we have the following:
\begin{proposition}
	\label{theo:loss_lb_2}
	In \method{}, with the same assumptions on $F_k$ and $\mathbf{w}^t$ as in Theorem~\ref{theo:loss}, and \eqref{eq:aggregate_2} used as the aggregation rule in Algorithm \ref{alg:fexprox_prob_lb3}, the lower-bound on expected decrease in the global objective loss function satisfies
	\vspace{-0.05in}
	\begin{align}
		\mathbb{E}[f(\mathbf{w}^{t+1})] \leq f(\mathbf{w}^{t}) - \frac{K}{\mu N} \sum_{k = 1}^N | \langle \nabla f(\mathbf{w}^t), \nabla F_k (\mathbf{w}^t) \rangle | \nonumber \\
		+ B \Bigg (\frac{L(\gamma+1)}{\mu \mu'}+\frac{\gamma}{\mu} + \frac{BL(1+\gamma)^2}{2 \mu'^2} \Bigg) \norm{\nabla f(\mathbf{w}^t)}^2.
	\end{align}
\end{proposition}

\begin{proof}
The proof is similar to that of Proposition~\ref{theo:loss_lb}, with the key difference being that Lemma~\ref{lem:square} now holds with equality.
\end{proof}

%In \method, the aggregation step only makes use of devices with $\langle \nabla F_k (\mathbf{w}^t), \nabla f(\mathbf{w}^t) \rangle > 0$ while completely wasting those with negative values and also it needs $2K$ devices instead of $K$ as in the case of \fedprox{} algorithm. We improve over \method{} by taking into consideration and only need $K$ devices with the following insight: \emph{those devices with $\langle \nabla F_k (\mathbf{w}^t), \nabla f(\mathbf{w}^t) \rangle < 0$ are still useful by aggregating the negatives of $\Delta \mathbf{w}^{t+1}_k$}. In particular, we modify the aggregation rule to the following simpler one:
\textbf{Comparison}: In comparing our result in Proposition~\ref{theo:loss_lb_2} with that of the \lb-near-optimal selection distribution in Definition~\ref{def:lb_near_optimal}, the new bound is better when $\frac{K}{\mu N} \sum_{k = 1}^N | \langle \nabla f(\mathbf{w}^t), \nabla F_k (\mathbf{w}^t) \rangle | > \frac{1}{\mu}\sum_{k = 1}^{N} |\langle \nabla f(\mathbf{w}^t),\nabla F_k(\mathbf{w}^t)\rangle| {P_{\mathsf{lb}}}^t_k$. This is the case when the data distribution across different devices becomes more uniform. To see this, let us consider two extreme cases: (i) under a uniform distribution of data, ${P_{\mathsf{lb}}}^t_k \approx 1/N$ and the new bound is $K$ times better than the \lb-near-optimal bound; (ii) when only one device has data, then the new bound is $K/N$ times worse than the \lb-near-optimal bound. In practice, the scenarios closer to case 1 will be much more prevalent than those similar to case 2, and thus most of the time, the new bound tends to be better than the earlier one.

%{\color{red}Analytical comparison between Theorem 3\&4 ...}

% !TEX root = fedlearn_nonuniform.tex

\section{Handling Computation and Communication Heterogeneity}
\label{sec:hetero_folb}
A practical consideration of distributed optimization on edge devices is the heterogeneity of computing power and communication between those devices and the central server. In this section, we show how \method{} can be easily adapted to handle heterogeneity by tweaking the aggregation rule slightly.

\subsection{Modeling heterogeneous communication and computation}
Each device participating in the federated learning process has a different communication delay when communicating with the central server and computation resources reserved for optimization. We model these two aspects as follows:

\textbf{Communication delay}: For each device $k$, we assume that the time it takes for one round of communication between device $k$ and central server is bounded above by $T^c_k$. This value $T^c_k$ can be obtained with high confidence by taking the 99th percentile of the distribution used to model the communication delay, e.g. exponential distribution.

\textbf{Computation resources}: Each device $k$ can only reserve a certain amount of resources to carry out optimization of the local function $h_k(\mathbf{w}; \mathbf{w}^t)$. Thus, we relax our assumption of having an uniform $\gamma$-inexact local solver in all devices to allow each device to have particular $\gamma_k$-inexact local solver where $\gamma_k$ can differ at every round of optimization and computed as $\gamma_k = \frac{\norm{\nabla h(\mathbf{w}^{t+1}_k, \mathbf{w}^t_k)}}{\norm{\nabla h(\mathbf{w}^t_k, \mathbf{w}^t_k)}}$. Note that we assume $\gamma_k \in [0,1]$ as in the case of local solvers being gradient descent algorithm. Hence, let $\tau$ is the amount of time for an optimization round dictated by the central server, we allow each selected device $k$ to perform any optimization within $\tau - T^c_k$ time and return the updated parameter $\mathbf{w}^{t+1}_k$ and $\gamma_k$ back to the central server. This scheme allows great flexibility and practicality since a device can use any amount of resources available and any local optimization algorithm that it has access to at every round.
\vspace{-0.05in}

\subsection{\method{} with communication and computation heterogeneity}
We show that \method{} can easily adapt to the inherent heterogeneity nature of communication and computation by adjusting it aggregation scheme to find a near-optimal convergence rate.

\textbf{New loss bound with heterogeneity presence.}
We first prove the following theorem showing the decrease of loss function in non-uniform \fedprox{} with heterogeneity presence:

\begin{theorem}
	\label{theo:loss_hetero}
	With the same assumptions as in Theorem~\ref{theo:loss} and the presence of communication and computation heterogeneity, suppose that $\mathbf{w}^t$ is not a stationary solution, in non-uniform \fedprox{}, we have the following expected decrease in the global objective function:
	\vspace{-0.05in}
	\begin{align}
		&E[f(\mathbf{w}^{t+1})] \leq f(\mathbf{w}^{t}) - \frac{1}{K\mu} E \Bigg[ \sum_{k \in S_t} \Bigg ( \langle \nabla f(\mathbf{w}^t),\nabla F_k(\mathbf{w}^t)\rangle \nonumber \\ & \qquad - B \left(\frac{L}{\mu \mu'} + \frac{1}{\mu} + \frac{3 L B}{2 K \mu'^2}\right)\gamma_k \norm{\nabla f(\mathbf{w}^t)}^2 \Bigg) \Bigg] \nonumber \\
		& \qquad + \left( \frac{L B^2}{2 \mu'^2} + \frac{LB}{\mu\mu'}\right) \norm{\nabla f(\mathbf{w}^t)}^2,
	\end{align}
	where the expectation is with respect to the choice of $K$ devices following probabilities $P^t_k, k = 1,\dots, N$.
\end{theorem}

\textbf{Implications of Theorem~\ref{theo:loss_hetero}}. Theorem~\ref{theo:loss_hetero} states that in the presence of communication and computation heterogeneity, the bound of loss decrease at a round depends not only on the inner products between local and global gradients but also on the optimality of the solutions returned by the individual devices. In other words, a device is more beneficial to the global model if the following two conditions hold:
\begin{itemize}
	\item[(1)] The local gradient $\nabla F_k(\mathbf{w}^{t})$ is well aligned with the global gradient $\nabla f(\mathbf{w}^t)$.
	\item[(2)] It has enough resources to perform optimization to find a decent solution, i.e., small $\gamma_k$.
\end{itemize}
Both of these conditions are intuitive and reflecting the importance of each device during the learning process. Unfortunately, we cannot evaluate any of the two criteria before selecting devices without expensive prior communication and computation. However, we show that \method{} can handle these challenges easily by tweaking the aggregation rule.

\textbf{Near-optimal selection distribution.}
From Theorem~\ref{theo:loss_hetero}, we can obtain a similar optimal selection probability distribution to that of Theorem~\ref{theo:loss_improved} which focuses on devices with high values of $I^t_k = \langle \nabla f(\mathbf{w}^t),\nabla F_k(\mathbf{w}^t)\rangle - B \left(\frac{L}{\mu \mu'} + \frac{1}{\mu} + \frac{3 L B}{2 K \mu'^2}\right)\gamma_k \norm{\nabla f(\mathbf{w}^t)}^2$. In other word, a near-optimal distribution will select device $k$ with probability:
\vspace{-0.03in}
\begin{align}
	{P_{\mathsf{lbh}}}^{t}_k = \frac{|I^t_k| }{\sum_{k' = 1}^N |I^t_{k'}| }
\end{align}
with the loss decrease satisfies:
\vspace{-0.03in}
\begin{align}
	 & E[f(\mathbf{w}^{t+1})] \leq f(\mathbf{w}^{t}) - \frac{1}{\mu} \sum_{k = 1}^N \Bigg ( \langle \nabla f(\mathbf{w}^t),\nabla F_k(\mathbf{w}^t)\rangle \nonumber \\ 
	 & \quad - B \left(\frac{L}{\mu \mu'} + \frac{1}{\mu} + \frac{3 L B}{2 K \mu'^2}\right)\gamma_k \norm{\nabla f(\mathbf{w}^t)}^2 \Bigg) {P_{\mathsf{lbh}}}^{t}_k \nonumber \\
	 & \quad + \left( \frac{L B^2}{2 \mu'^2} + \frac{LB}{\mu\mu'}\right) \norm{\nabla f(\mathbf{w}^t)}^2.
\end{align}

\textbf{\method{} aggregation for communication and computation heterogeneity.}
\method{} with heterogeneity of communication and computation adopts the following aggregation rule:
\vspace{-0.03in}
\begin{align}
	\label{eq:aggregate_hetero}
	\mathbf{w}^{t+1} = \mathbf{w}^{t} + \sum_{k \in S^{t}_1} \frac{ I^t_{1k} }{\sum_{k' \in S^{t}_1} |I^t_{1k'}| } \Delta \mathbf{w}^{t+1}_k,
\end{align}
where $I^t_{1k} = \langle \nabla_1 f(\mathbf{w}^t),\nabla F_k(\mathbf{w}^t)\rangle - B \left(\frac{L}{\mu \mu'} + \frac{1}{\mu} + \frac{3 L B}{2 K \mu'^2}\right)\gamma_k \norm{\nabla_1 f(\mathbf{w}^t)}^2$, and $\nabla_1 f(\mathbf{w}^t)$ is defined in (\ref{eq:gradest}).

%With that aggregation, we can prove a similar result to Theorem~\ref{theo:loss_lb_2} that \method{} achieves the same near-optimal loss decrease at every optimization round. The proof is similar to that of Theorem~\ref{theo:loss_lb_2}.

\textbf{Avoiding constant estimations}: In the new \method{} that deals with heterogeneity, updating the global parameter according to Equation~\ref{eq:aggregate_hetero} becomes more complicated compared to Equation~\ref{eq:aggregate_2} due to involving the set of constants $B, L, \mu'$ which need to be estimated before hand or on-the-air. Instead of requiring all these constants to be estimated, we propose to use a hyper-parameter $\psi = B \left(\frac{L}{\mu \mu'} + \frac{1}{\mu} + \frac{3 L B}{2 K \mu'^2}\right)$ that will be learned though hyper-parameter tuning similarly to $\mu$ in \fedprox{}. For tuning $\psi$, we can use a simple line search with a exponential step size, e.g. $\psi \in \{ 10^{-1}, 1, 10, 10^2 \}$ which is used in our experiments and found to be effective.
\vspace{-0.03in}

\section{Experiments}
\label{sec:exps}
In this section, we experimentally compare our proposed algorithm with existing state-of-the-art approaches and demonstrate faster convergence across different learning tasks in both synthetic and real datasets. We also confirm the advantage of taking into consideration the individual device optimization capability in the presence of communication and computation heterogeneity, showing our approach more suitable for practical federated learning implementations. 
\begin{figure*}[h!]
	\centering
	\begin{subfigure}[b]{\linewidth}
		\centering
		\includegraphics[width=\linewidth]{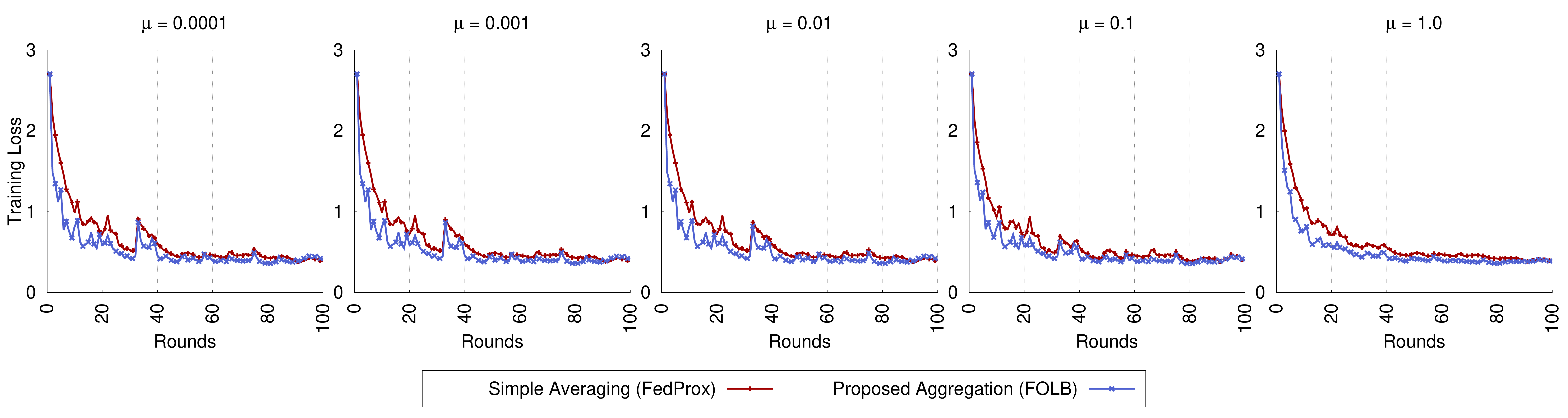}
		\caption{Training loss (Lower is better.)}
	\end{subfigure}
	\begin{subfigure}[b]{\linewidth}
		\centering
		\includegraphics[width=\linewidth]{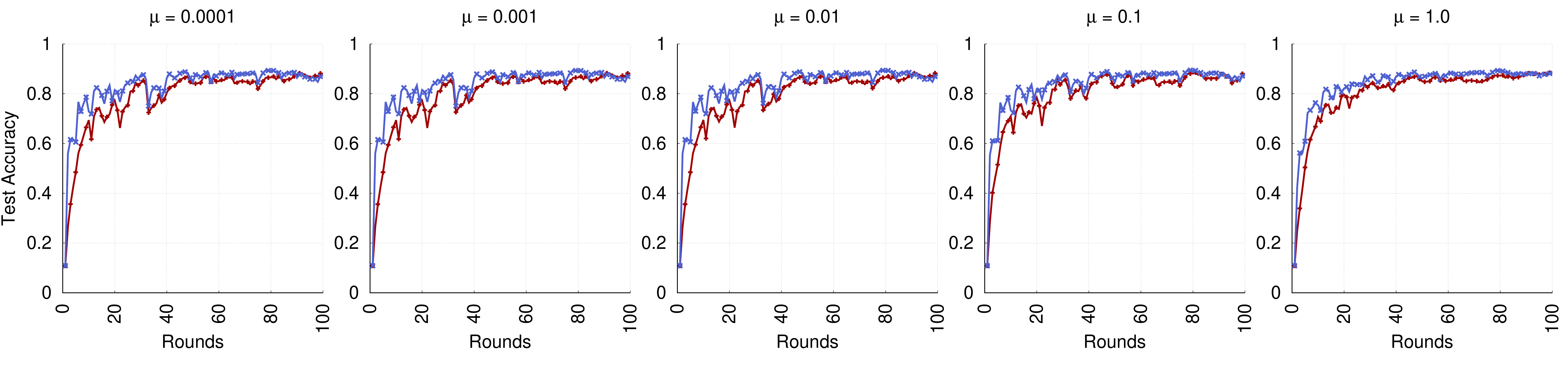}
		\caption{Test accuracy (Higher is better.)}
	\end{subfigure}
	\caption{Effectiveness of our proposed aggregation rule in \method{} compared to simple averaging in \fedprox{} (similarly in \fedave{}) across a wide range of proximal parameter $\mu$.}
	\label{fig:aggregation_effective}
	\vspace{-0.2in}
\end{figure*}
\begin{figure*}[h!]
	\centering
	\begin{minipage}{0.49\linewidth}
		\centering
		\begin{subfigure}[b]{0.51\linewidth}
			\centering
			\includegraphics[width=\linewidth]{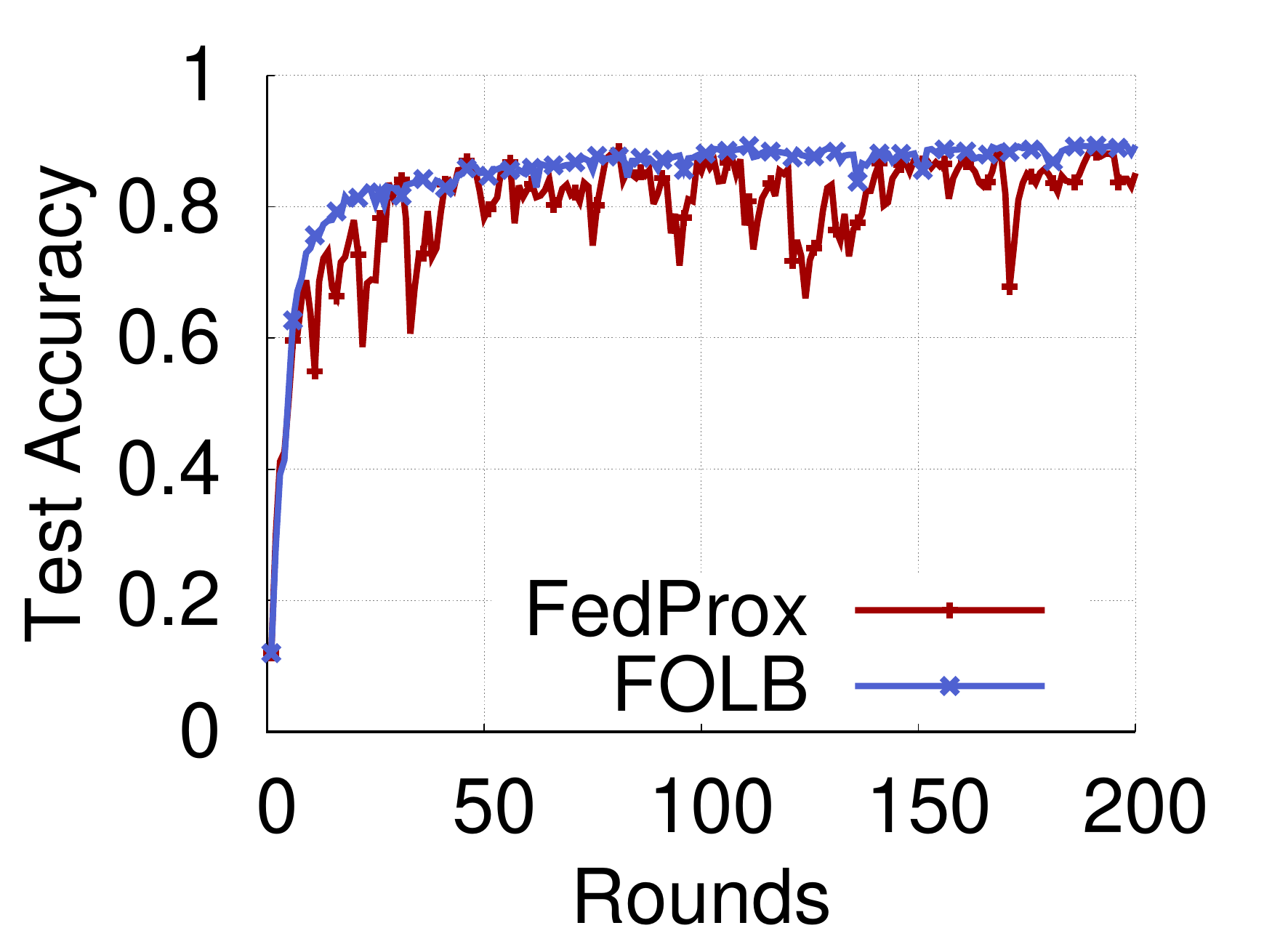}
			\caption{3-layer MLP}
		\end{subfigure}
		\hspace{-0.2in}
		\begin{subfigure}[b]{0.51\linewidth}
			\centering
			\includegraphics[width=\linewidth]{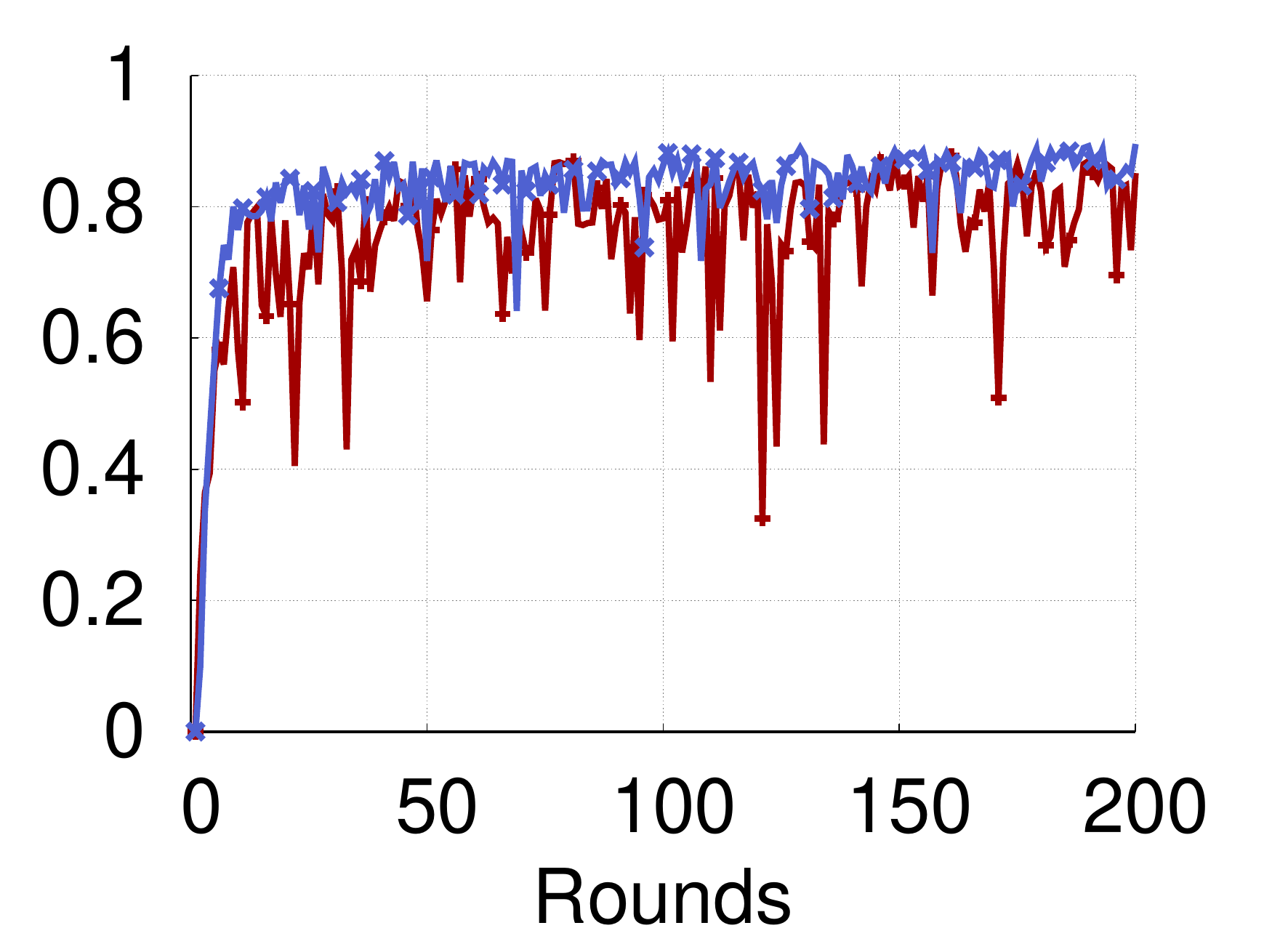}
			\caption{3-layer CNN}
		\end{subfigure}
		%		\vspace{-0.2in}
		\caption{Performance comparison between \method{} and \fedprox{} considering different neural network models, i.e., CNN and MLP with 3 layers, over the MNIST dataset and $\mu=0.01$.  \method{} results in a more stable model accuracy and outperforms \fedprox{}.}
		\label{fig:acc_vs_arch}
	\end{minipage}
	\hspace{0.02in}
	\begin{minipage}{0.49\linewidth}
		\begin{subfigure}[b]{0.51\linewidth}
			\centering
			\includegraphics[width=\linewidth]{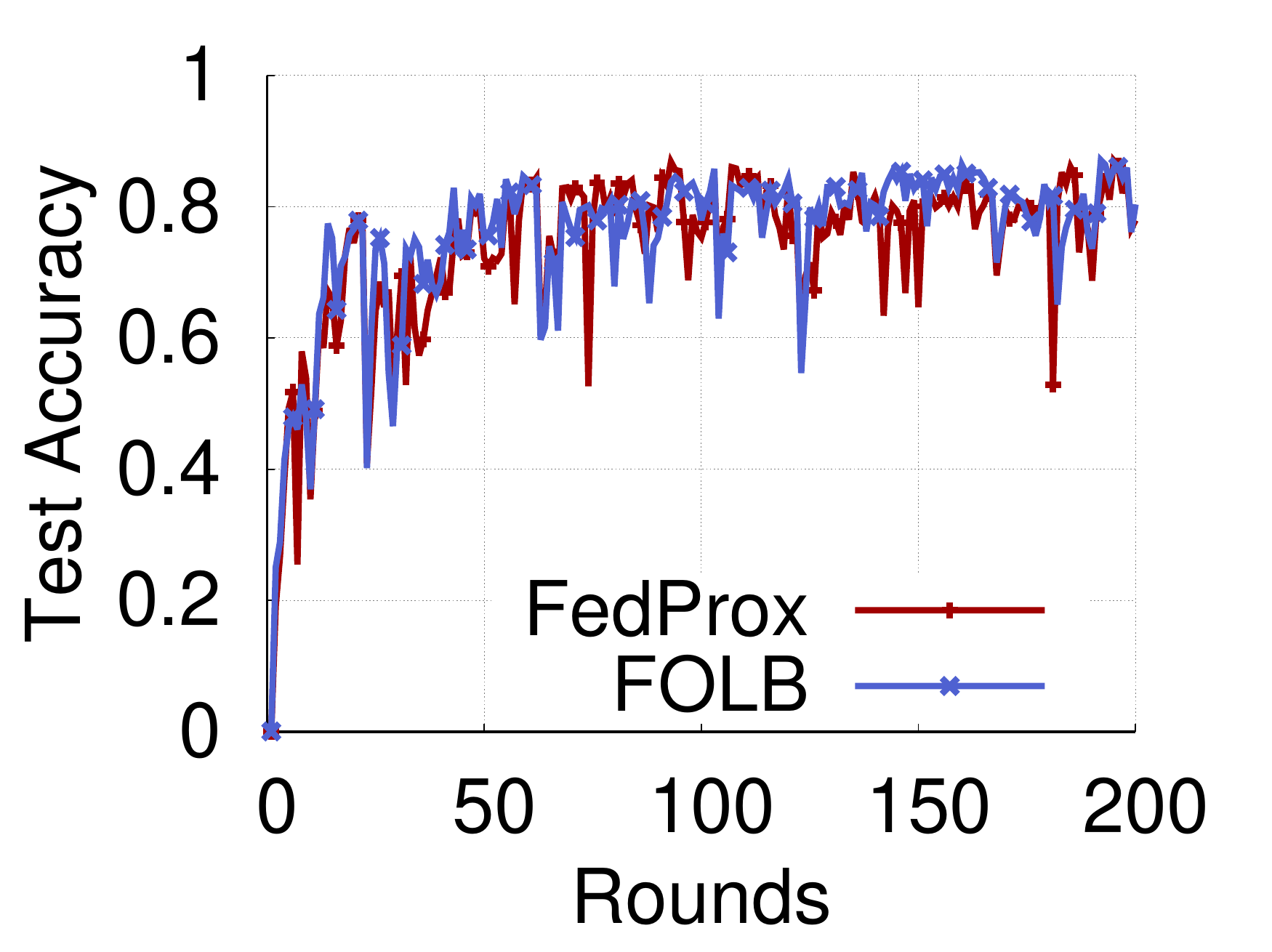}
			\caption{2 devices}
		\end{subfigure}
		\hspace{-0.2in}
		\begin{subfigure}[b]{0.51\linewidth}
			\centering
			\includegraphics[width=\linewidth]{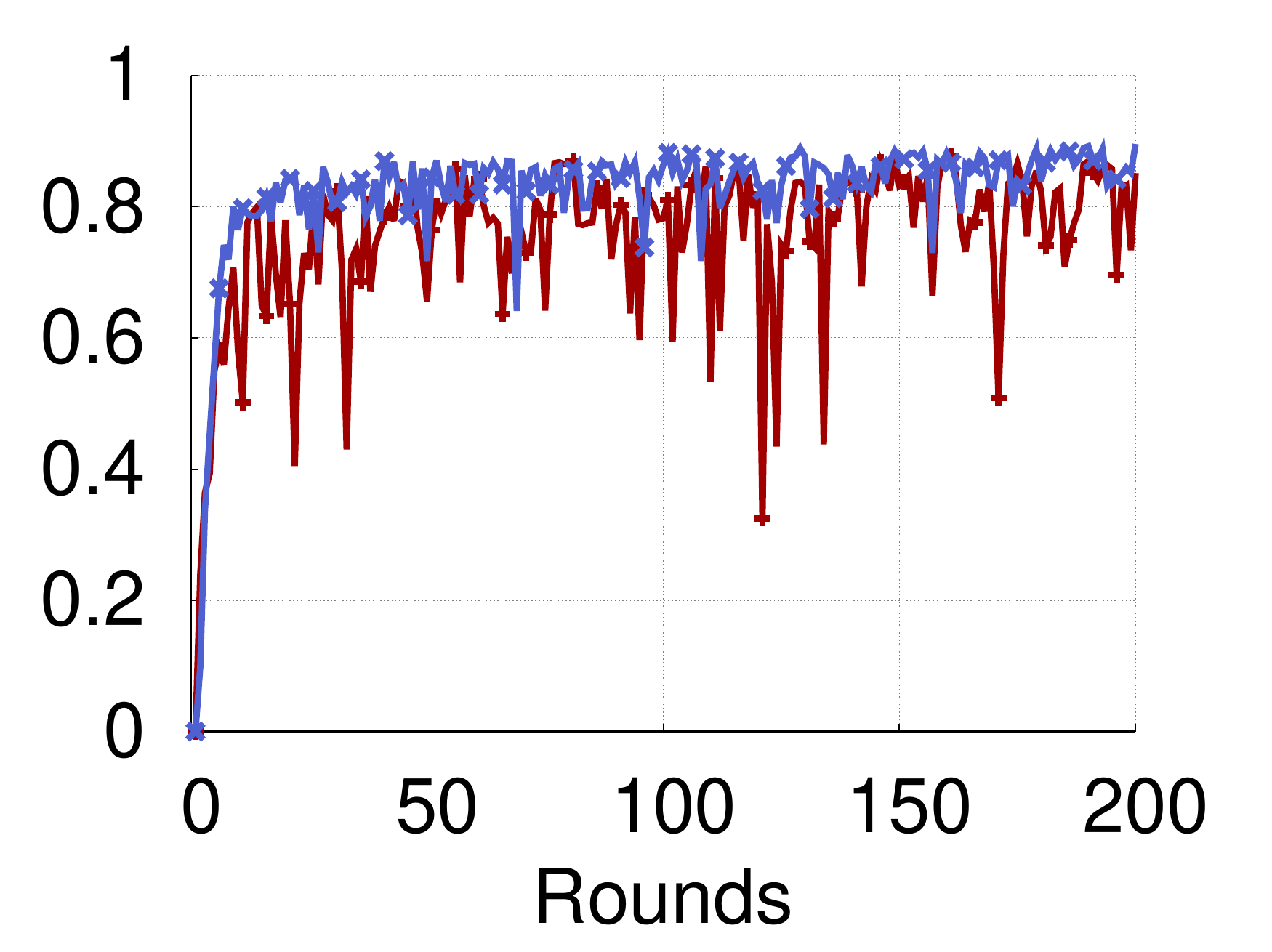}
			\caption{20 devices}
		\end{subfigure}
		%		\vspace{-0.2in}
		\caption{Performance comparison between \method{} and \fedprox{} considering different number of devices. With increasing the number of devices in each round, \method{} converges faster and stabilizes quicker than \fedprox{}. We use MNIST dataset with a 3-layer CNN and $\mu = 0.01$.}
		\label{fig:acc_vs_k}
	\end{minipage}
	\vspace{-0.2in}
\end{figure*}
\begin{figure*}[h!]
	\centering
	\begin{subfigure}[b]{0.261\linewidth}
		\includegraphics[width=\linewidth]{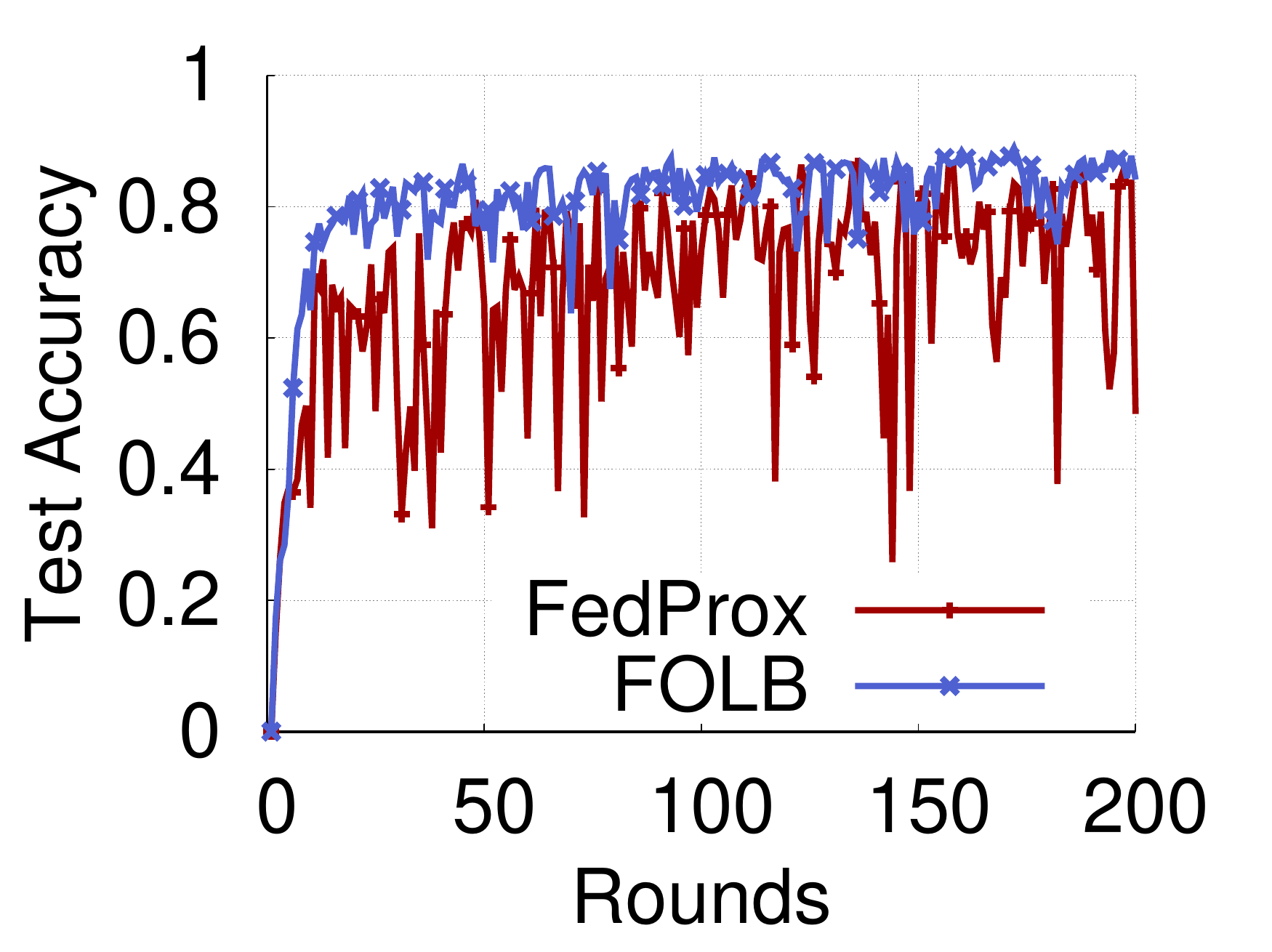}
		\caption{1 class per device (most extreme)}
	\end{subfigure}
	\hspace{-0.2in}
	\begin{subfigure}[b]{0.261\linewidth}
		\includegraphics[width=\linewidth]{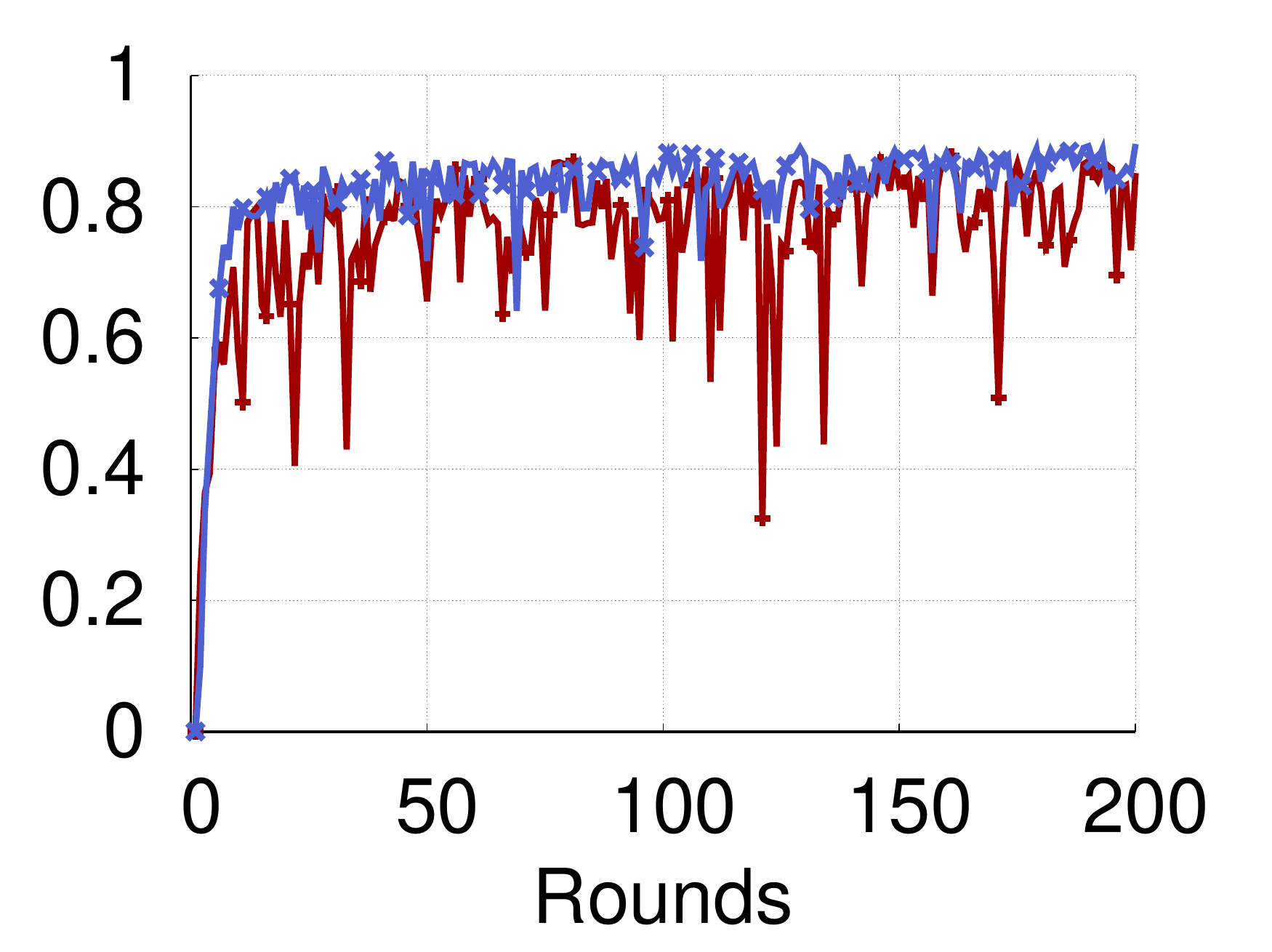}
		\caption{2 classes per device}
	\end{subfigure}
	\hspace{-0.2in}
	\begin{subfigure}[b]{0.261\linewidth}
		\includegraphics[width=\linewidth]{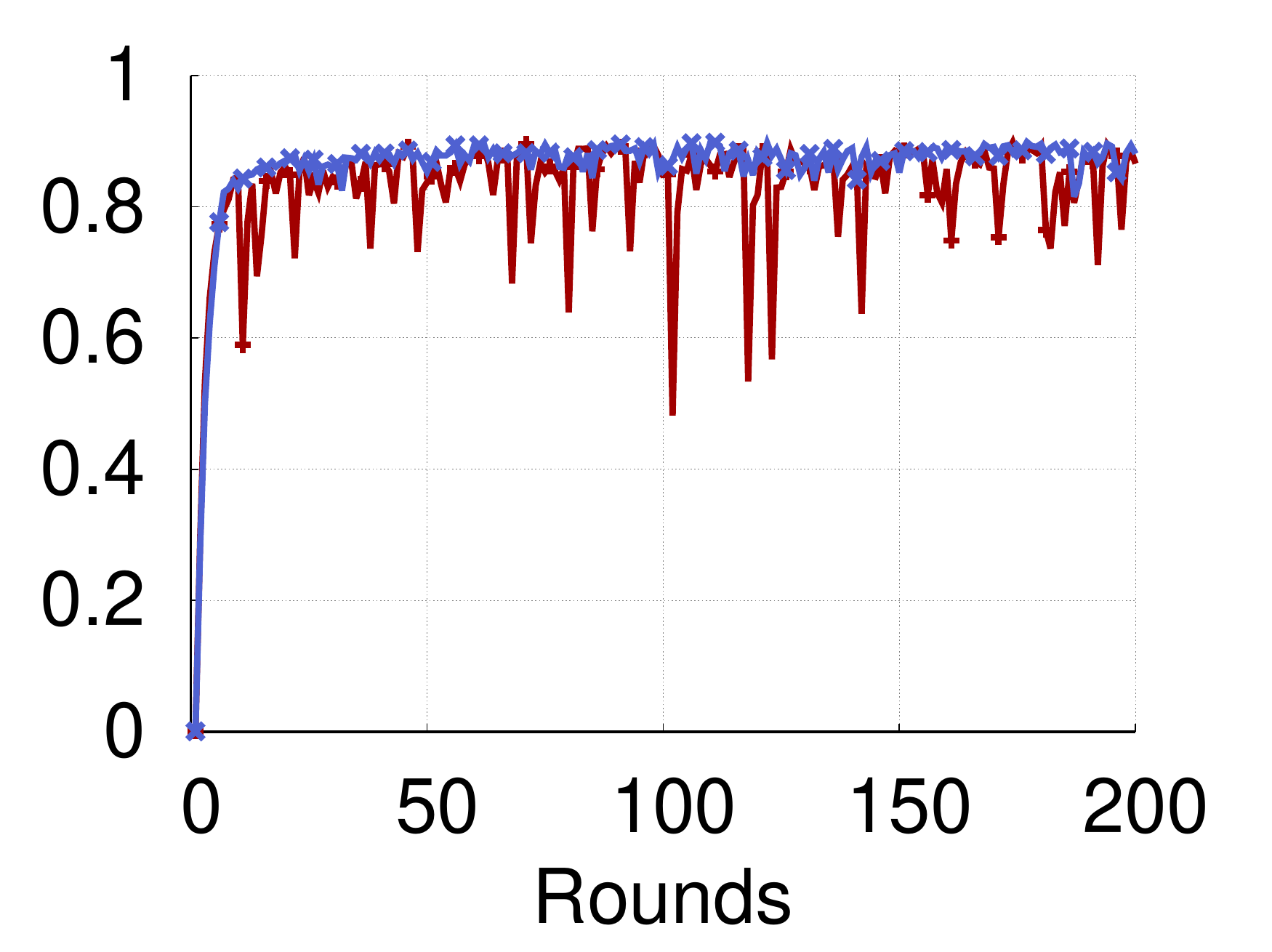}
		\caption{5 classes per device}
	\end{subfigure}
	\hspace{-0.2in}
	\begin{subfigure}[b]{0.261\linewidth}
		\includegraphics[width=\linewidth]{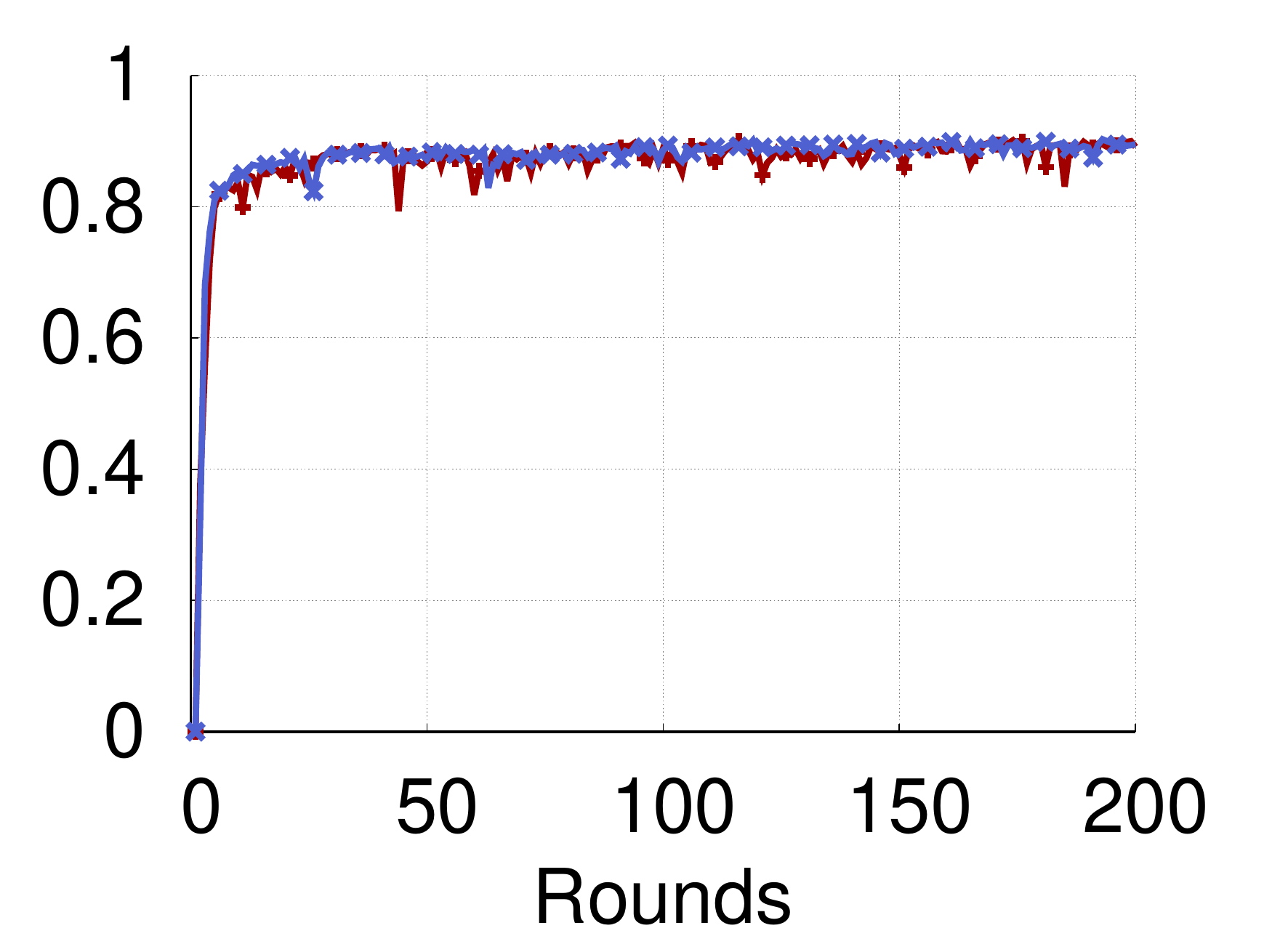}
		\caption{10 classes per device (IID)}
	\end{subfigure}
	\caption{Testing accuracy with different non-IID settings of the MNIST dataset, i.e., randomly assigning images of only a fixed number of different digits to each device. \method{} performs better than \fedprox{} specially in the most extreme non-IID setting.}
	\label{fig:acc_vs_imagesperdevice}
	\vspace{-0.2in}
\end{figure*}
\begin{figure*}[h!]
	\centering
	\begin{subfigure}[b]{0.261\linewidth}
		\includegraphics[width=\linewidth]{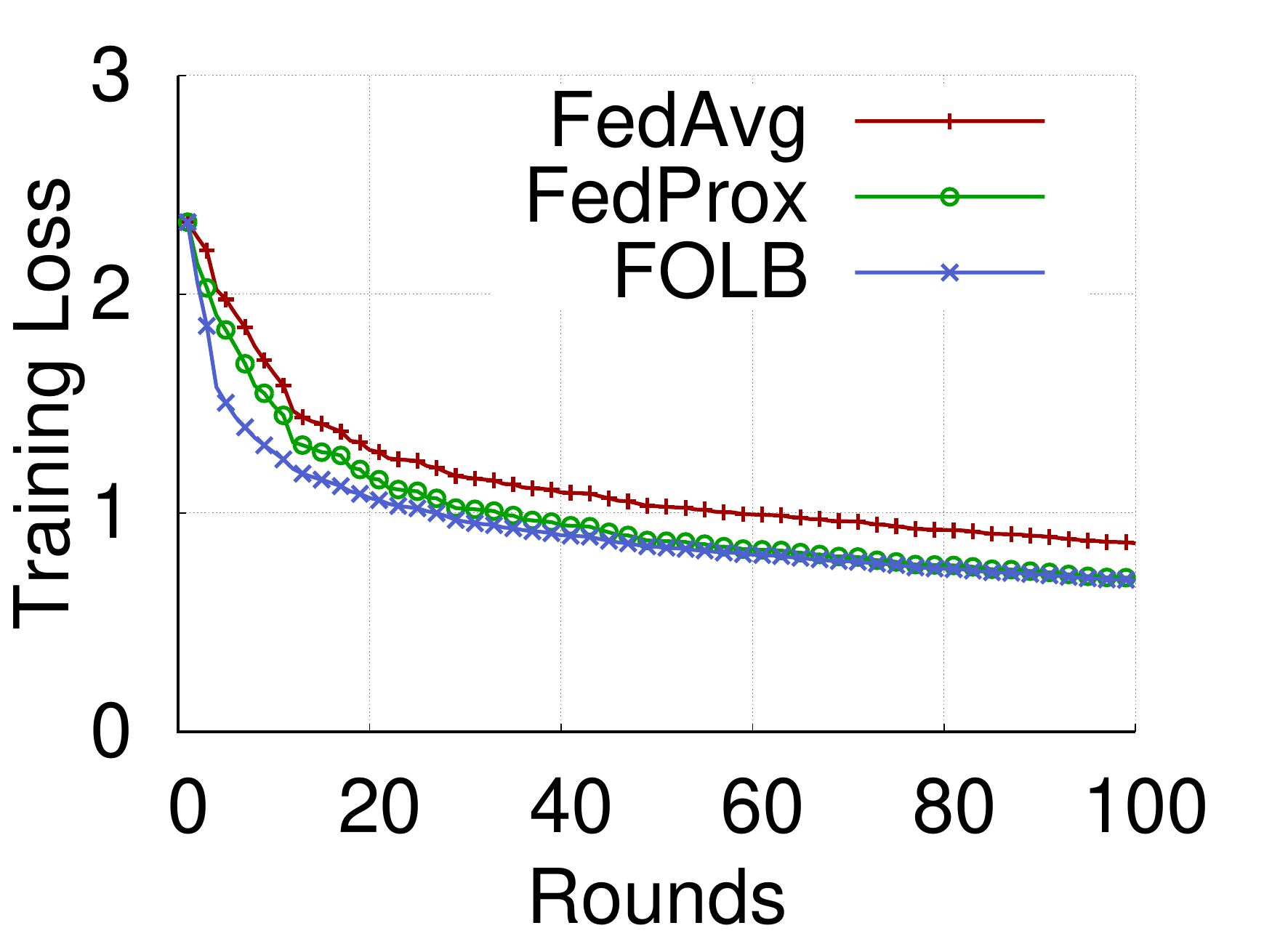}
		\caption{Synthetic\_iid}
	\end{subfigure}
	\hspace{-0.2in}
	\begin{subfigure}[b]{0.261\linewidth}
		\includegraphics[width=\linewidth]{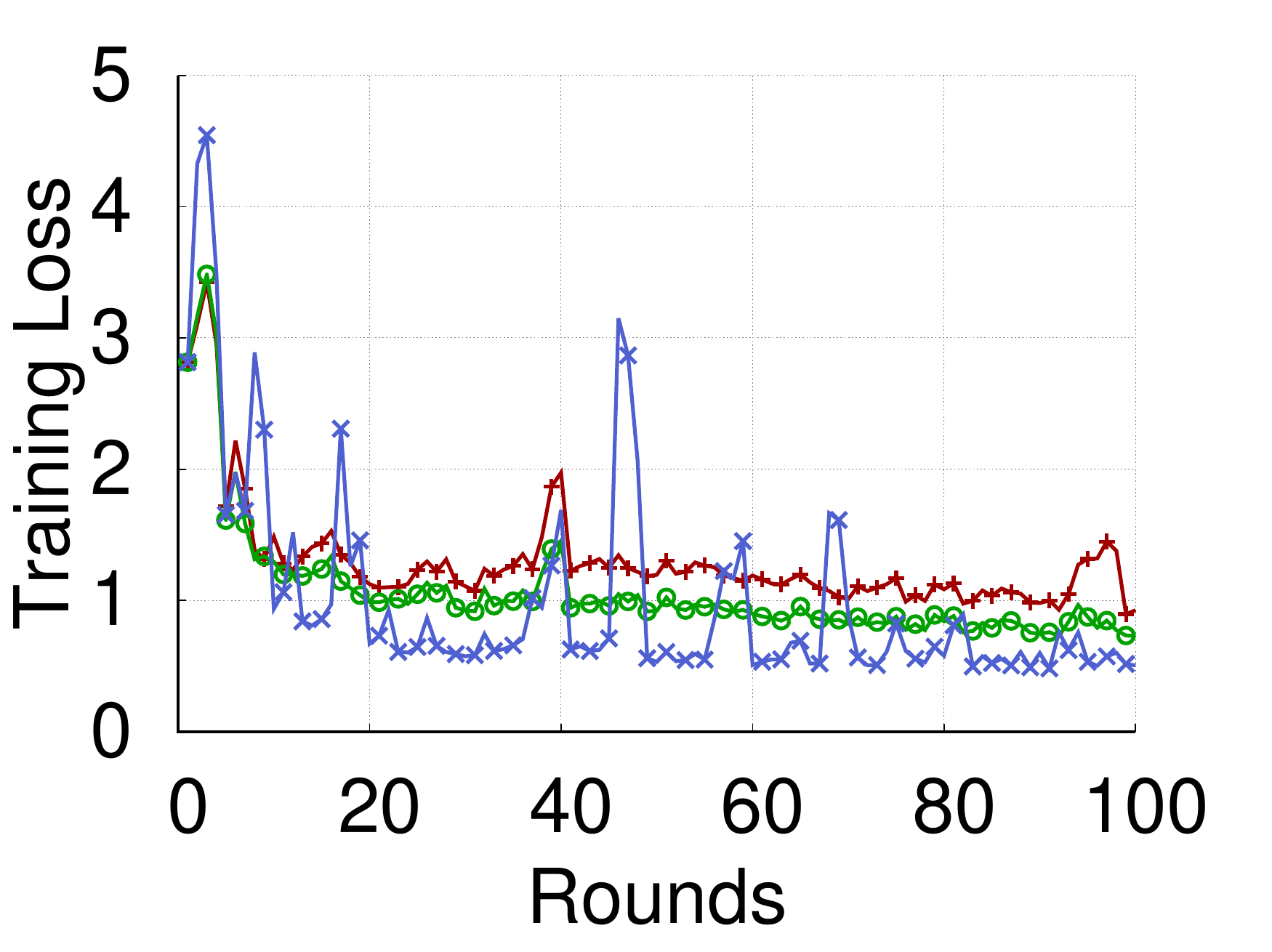}
		\caption{Synthetic\_1\_1}
	\end{subfigure}
	\hspace{-0.2in}
	\begin{subfigure}[b]{0.261\linewidth}
		\includegraphics[width=\linewidth]{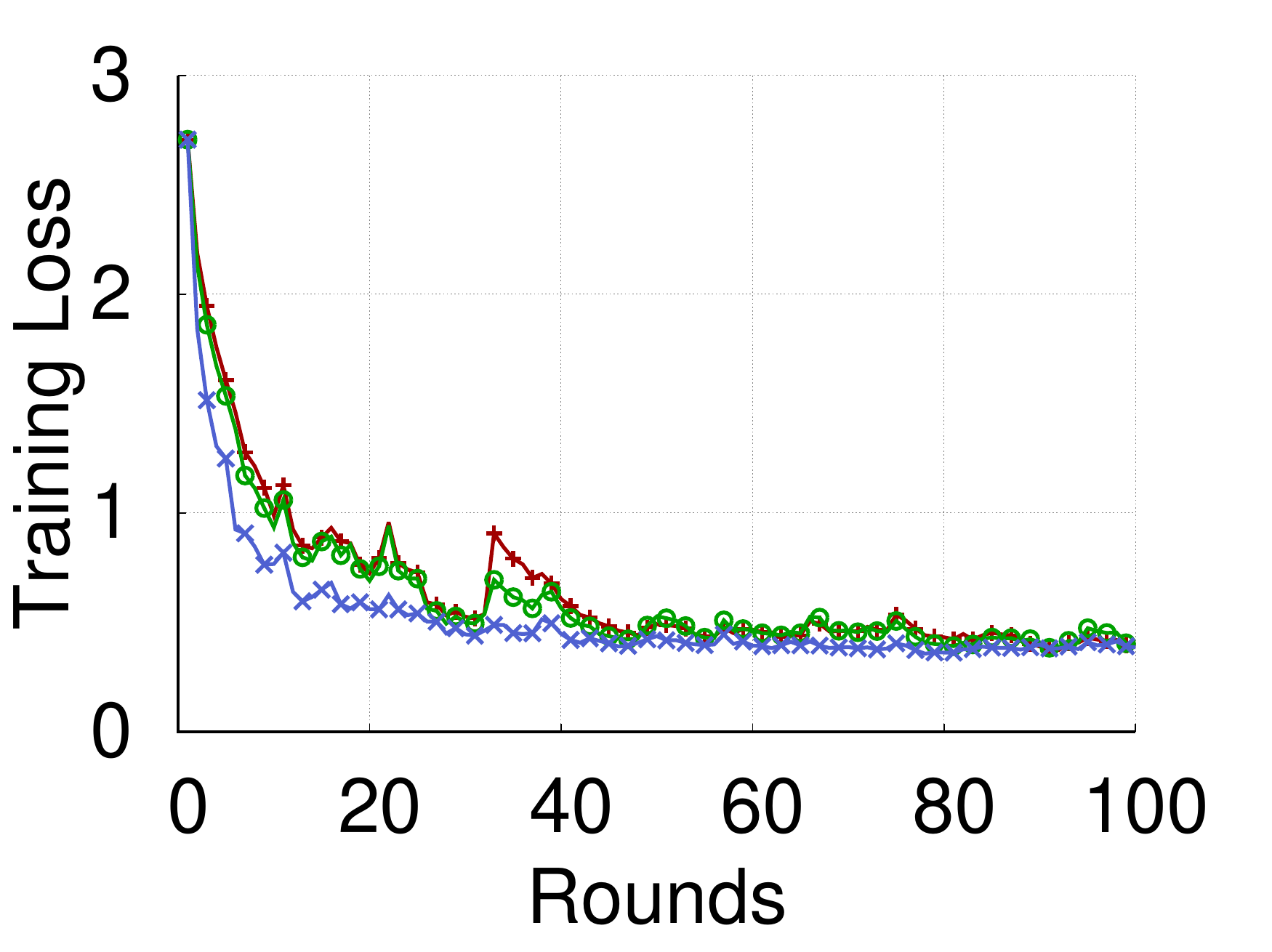}
		\caption{MNIST}
	\end{subfigure}
	\hspace{-0.2in}
	\begin{subfigure}[b]{0.261\linewidth}
		\includegraphics[width=\linewidth]{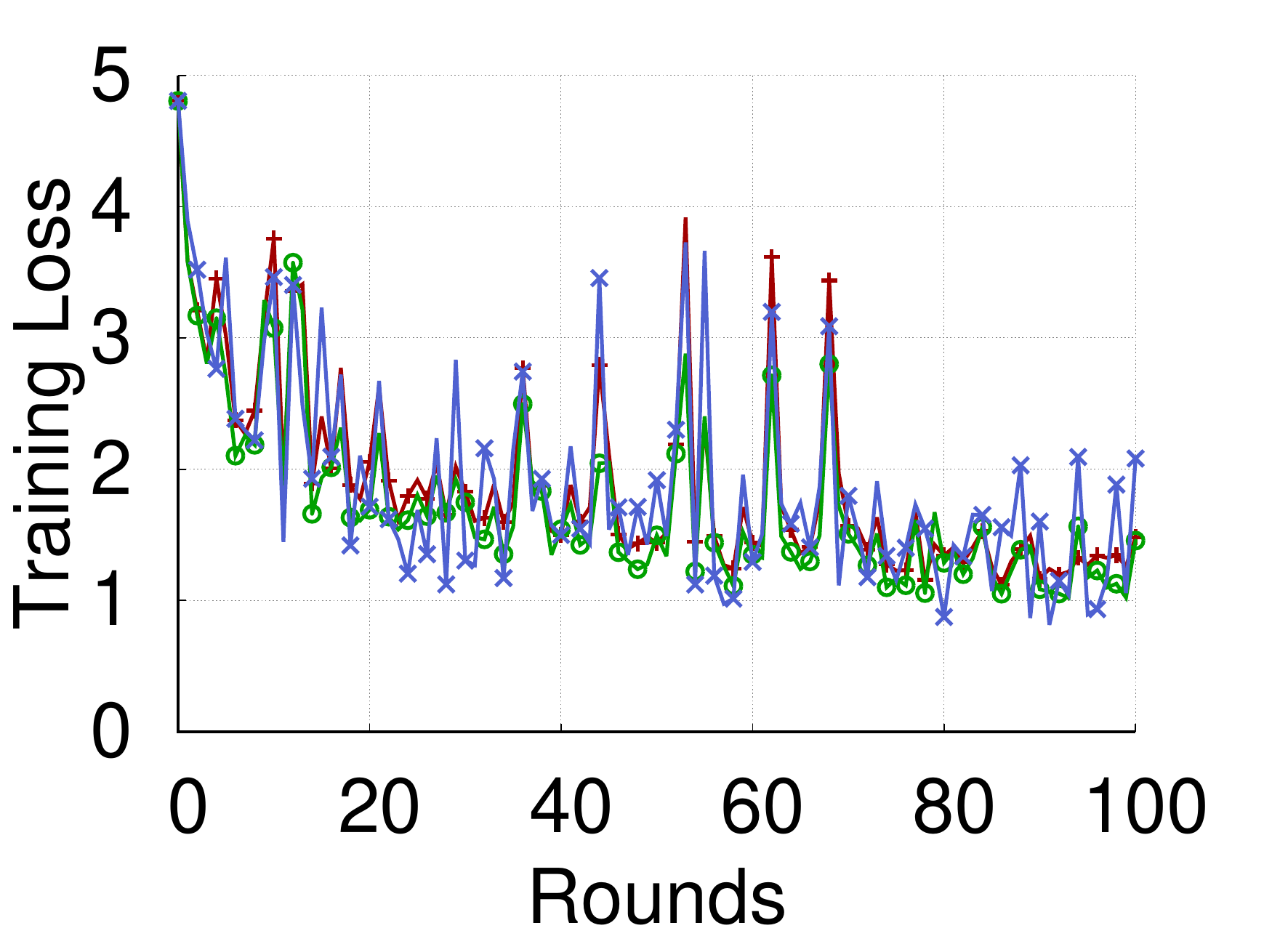}
		\caption{FEMNIST}
	\end{subfigure}
	\caption{Training loss of \method{}, \fedprox{} and \fedave{} on various datasets using linear model (multinomial logistic regression). \method{} can reach lower loss value than the others.}
	\label{fig:compare_loss_all_methods}
	\vspace{-0.2in}
\end{figure*}

\begin{figure*}[h!]
	\centering
	\begin{subfigure}[b]{0.261\linewidth}
		\includegraphics[width=\linewidth]{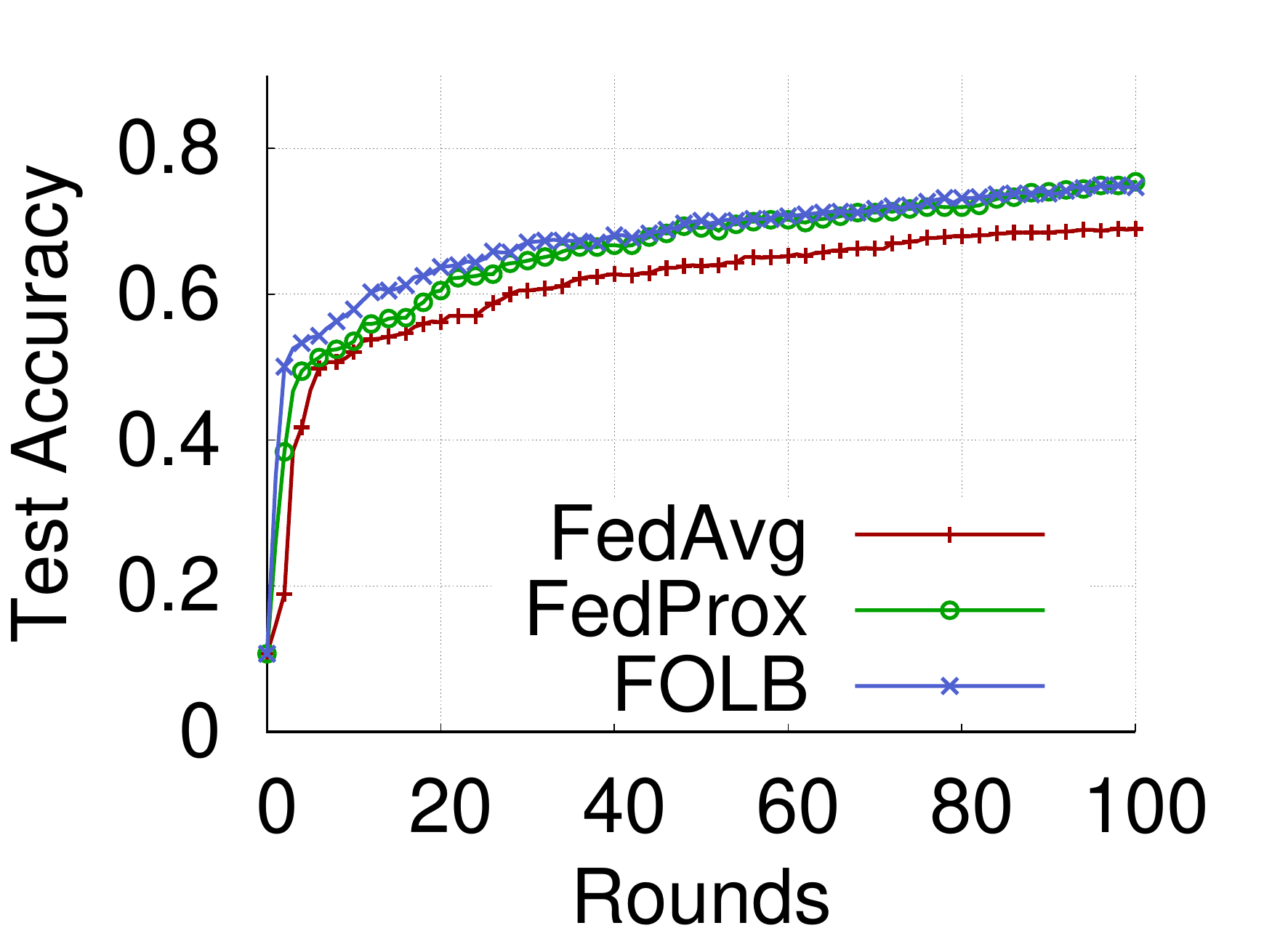}
		\caption{Synthetic\_iid}
	\end{subfigure}
	\hspace{-0.2in}
	\begin{subfigure}[b]{0.261\linewidth}
		\includegraphics[width=\linewidth]{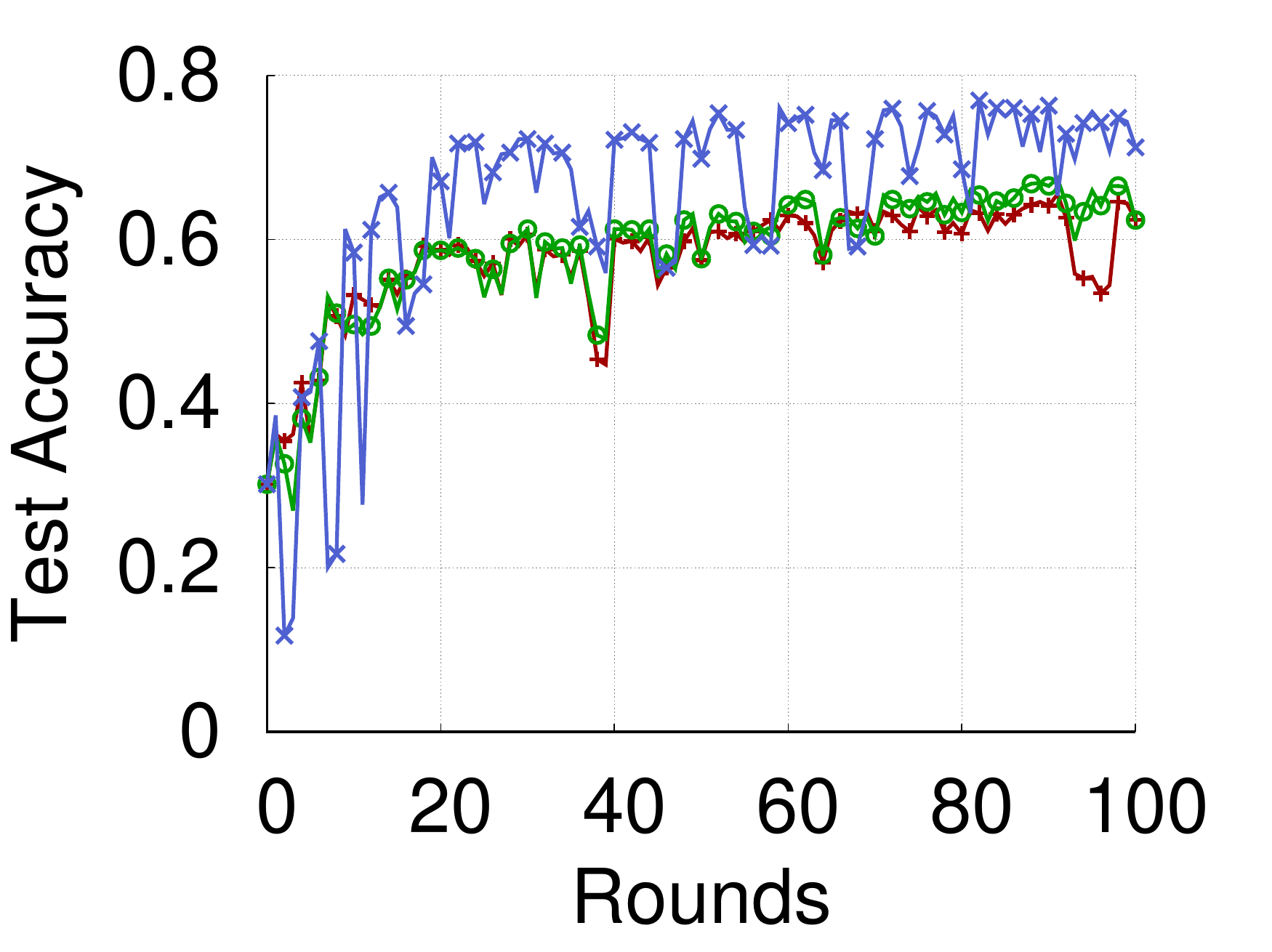}
		\caption{Synthetic\_1\_1}
	\end{subfigure}
	\hspace{-0.2in}
	\begin{subfigure}[b]{0.261\linewidth}
		\includegraphics[width=\linewidth]{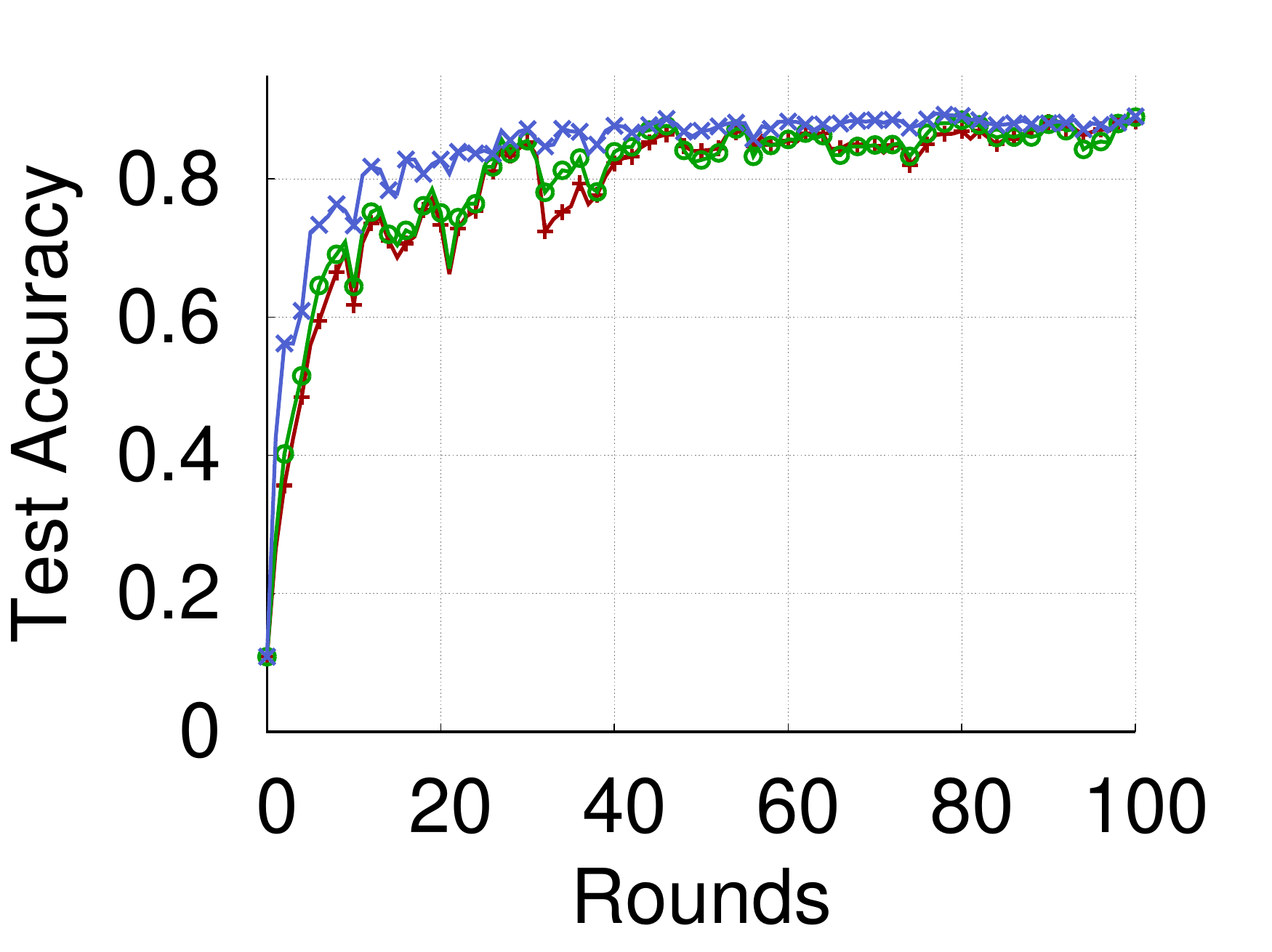}
		\caption{MNIST}
	\end{subfigure}
	\hspace{-0.2in}
	\begin{subfigure}[b]{0.261\linewidth}
		\includegraphics[width=\linewidth]{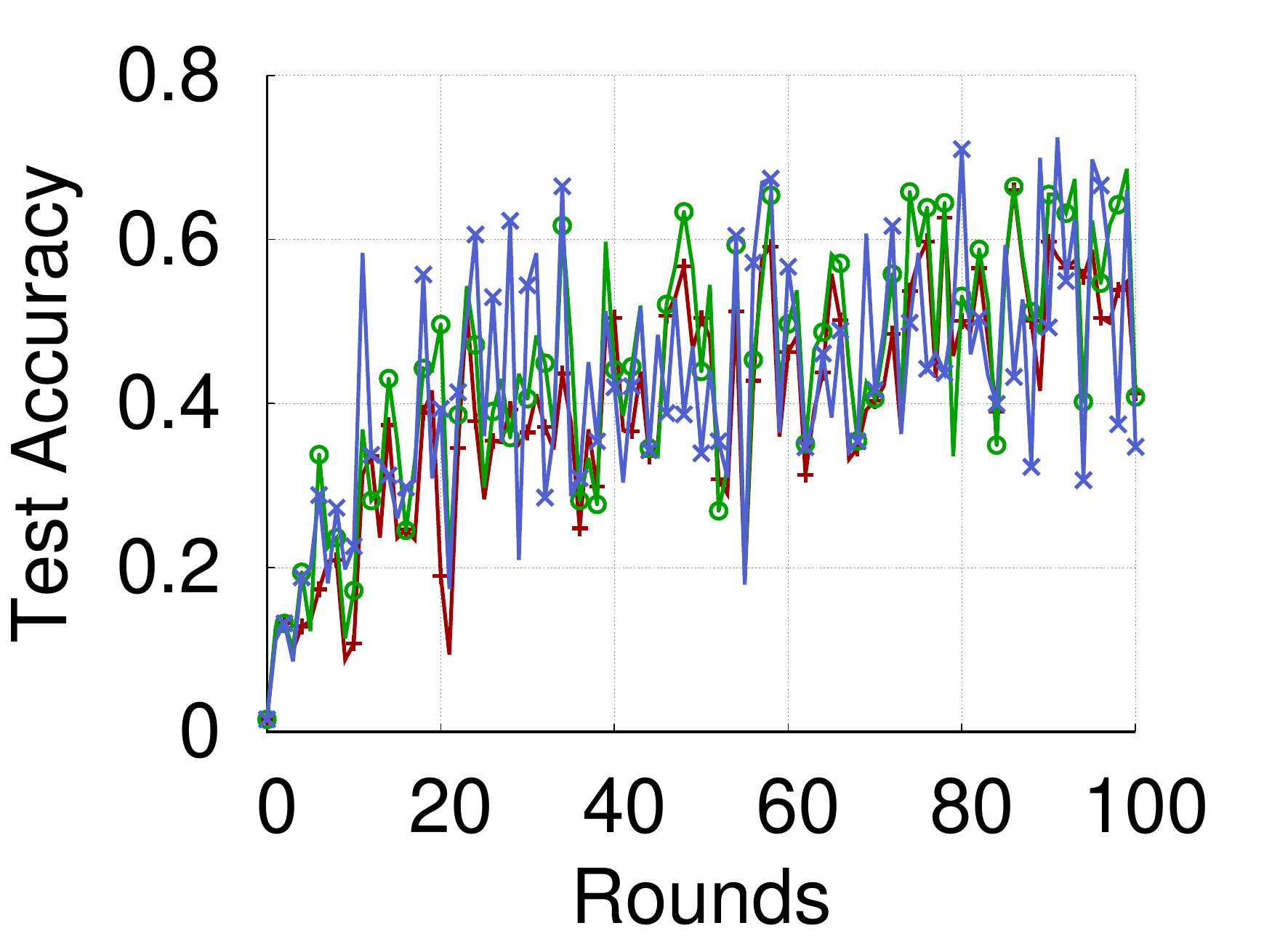}
		\caption{FEMNIST}
	\end{subfigure}
	\caption{Testing accuracy of \method{}, \fedprox{} and \fedave{} on various datasets using linear model (multinomial logistic regression). \method{} can reach higher level of accuracy than the others.}
	\label{fig:compare_acc_all_methods}
	\vspace{-0.2in}
\end{figure*}

%\begin{figure*}[h!]
%	\centering
%	\includegraphics[width=\linewidth]{figs/acc_vs_imagesperdevice}
%	\caption{Testing accuracy with different data partition of MNIST across devices. We simulate non-iid setting by allocating images of only fixed set class at each device.}
%	\label{fig:acc_vs_imagesperdevice}
%	\vspace{-0.2in}
%\end{figure*}

\begin{figure*}[h!]
	\centering
	\begin{minipage}{0.49\linewidth}
		\centering
		\begin{subfigure}[b]{0.51\linewidth}
			\centering
			\includegraphics[width=\linewidth]{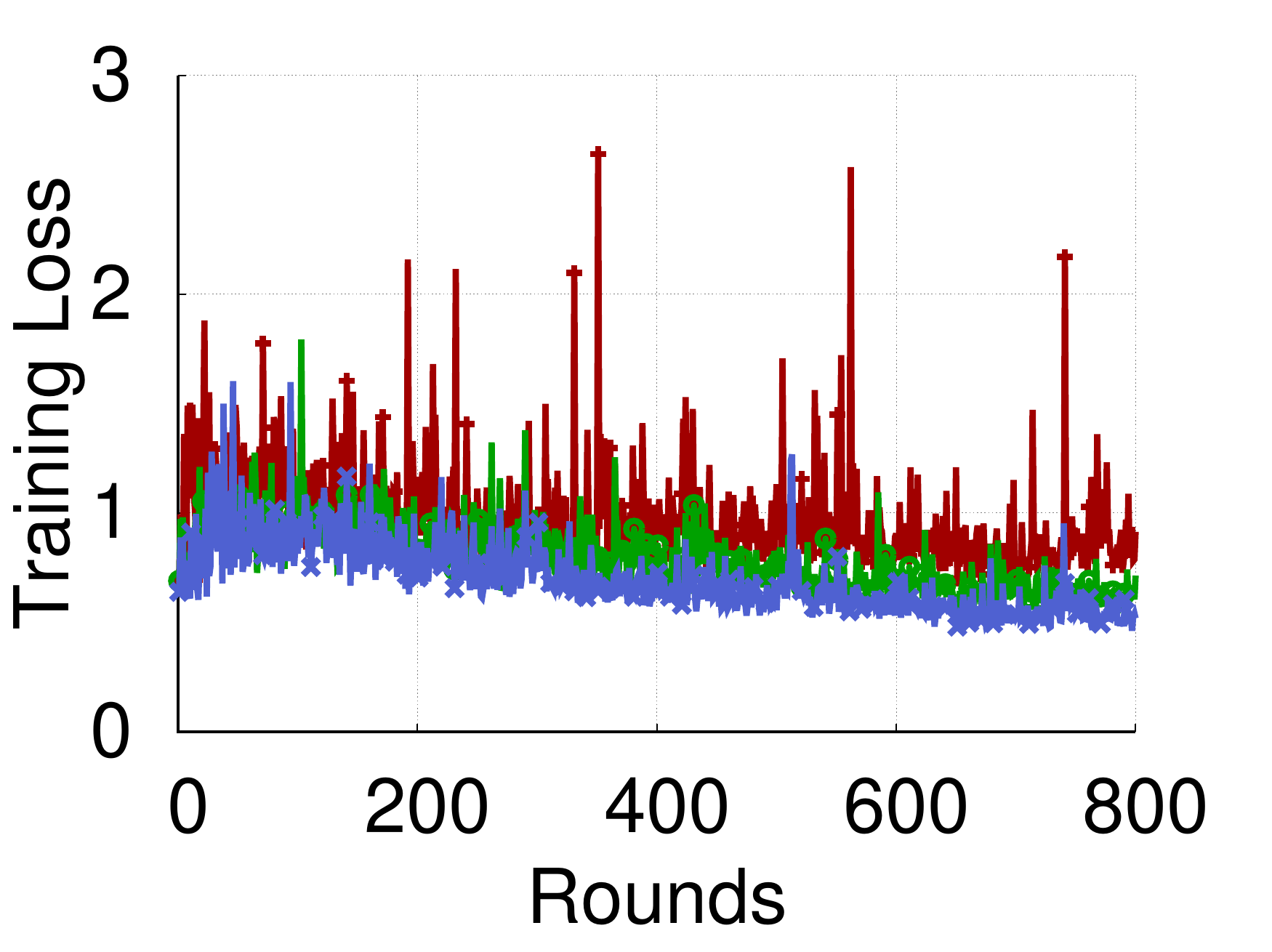}
			\caption{Sent140}
		\end{subfigure}
		\hspace{-0.2in}
		\begin{subfigure}[b]{0.51\linewidth}
			\centering
			\includegraphics[width=\linewidth]{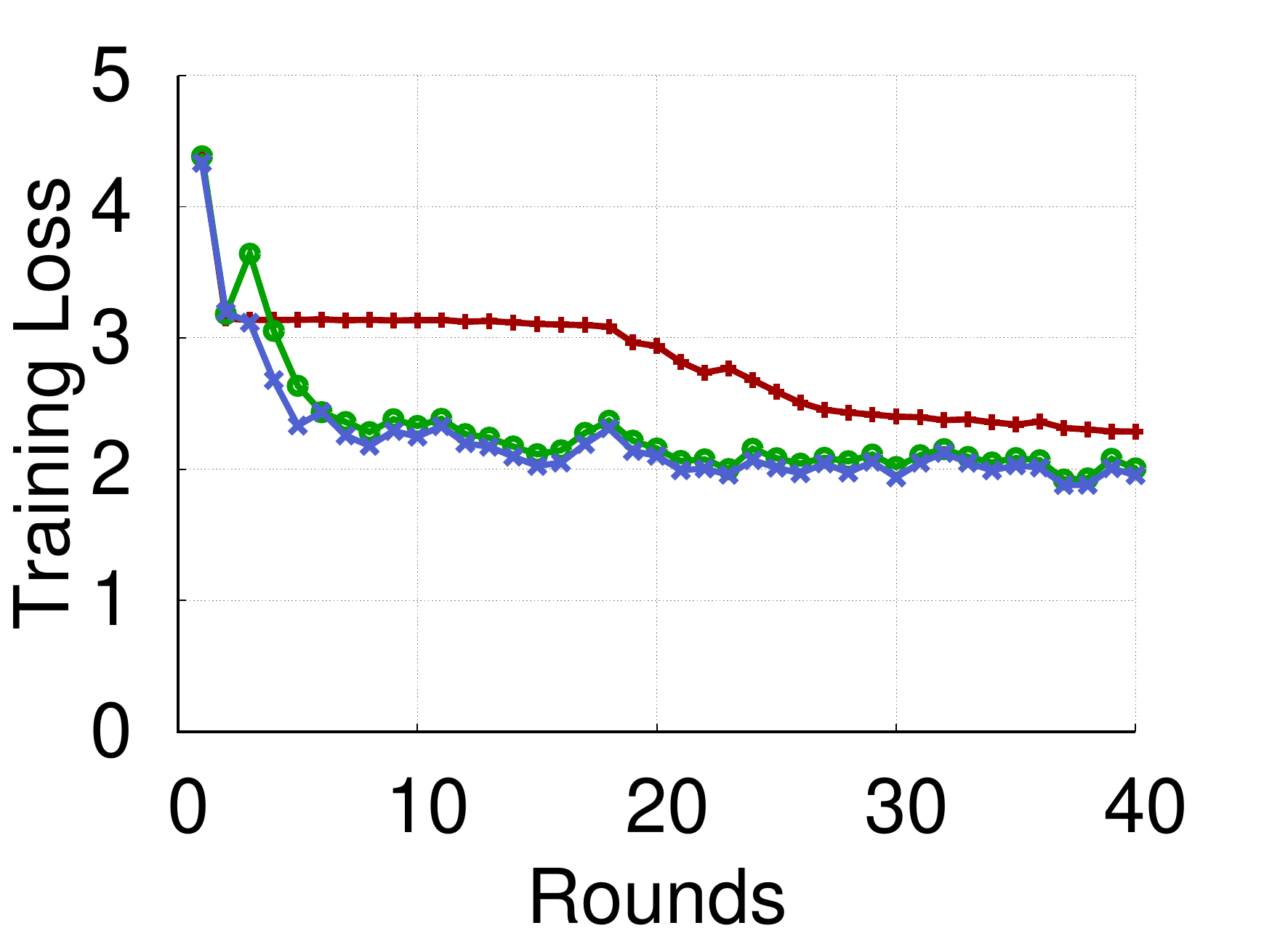}
			\caption{Shakespeare}
		\end{subfigure}
		\caption{Training loss of \method{}, \fedprox{} and \fedave{} on various datasets using non-linear model (LSTM). \method{} can reach lower loss value than the others.}
		\label{fig:compare_loss_all_methods_nonlinear}
	\end{minipage}
	\hspace{0.02in}
	\begin{minipage}{0.49\linewidth}
		\begin{subfigure}[b]{0.51\linewidth}
			\centering
			\includegraphics[width=\linewidth]{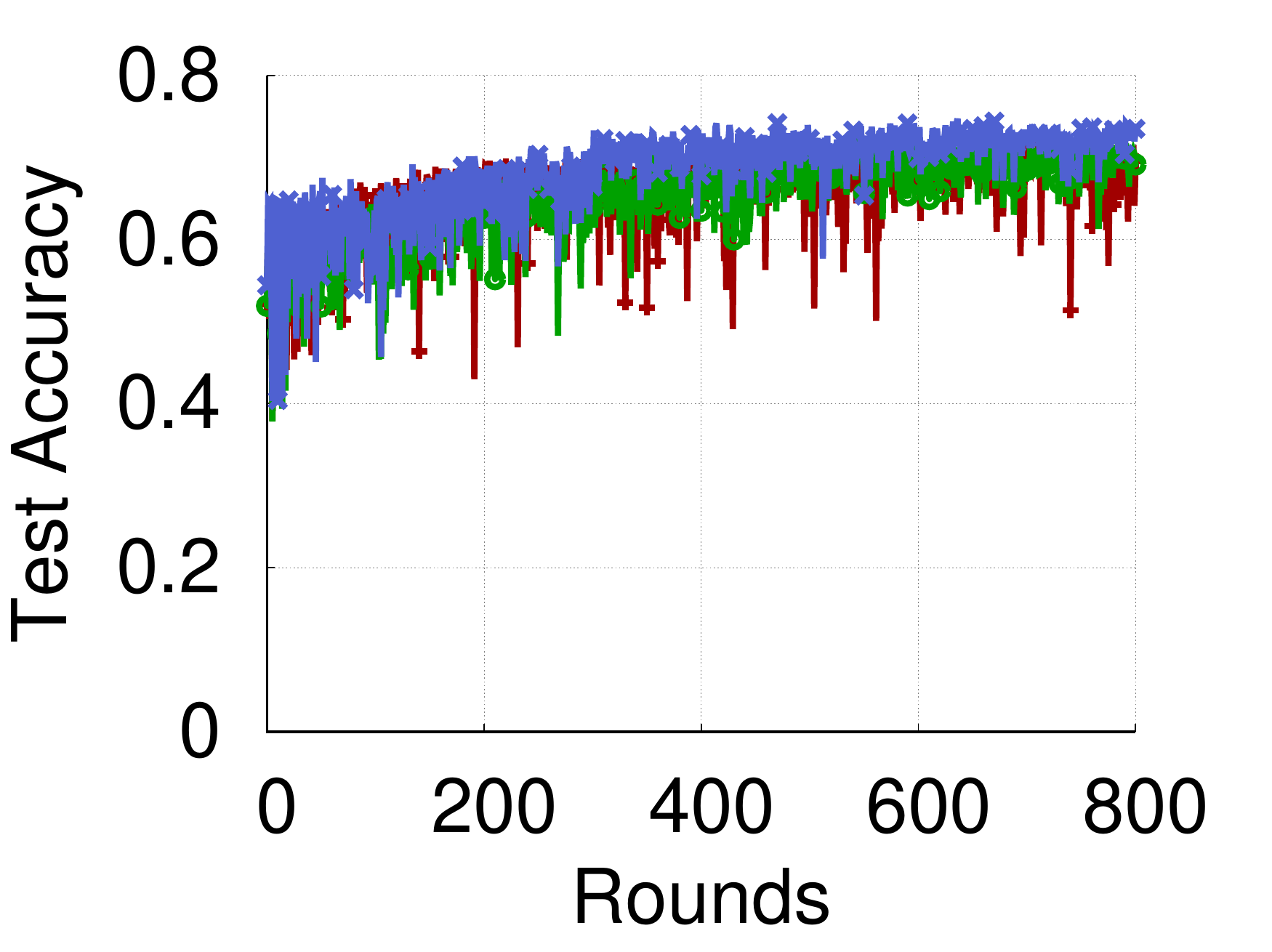}
			\caption{Sent140}
		\end{subfigure}
		\hspace{-0.2in}
		\begin{subfigure}[b]{0.51\linewidth}
			\centering
			\includegraphics[width=\linewidth]{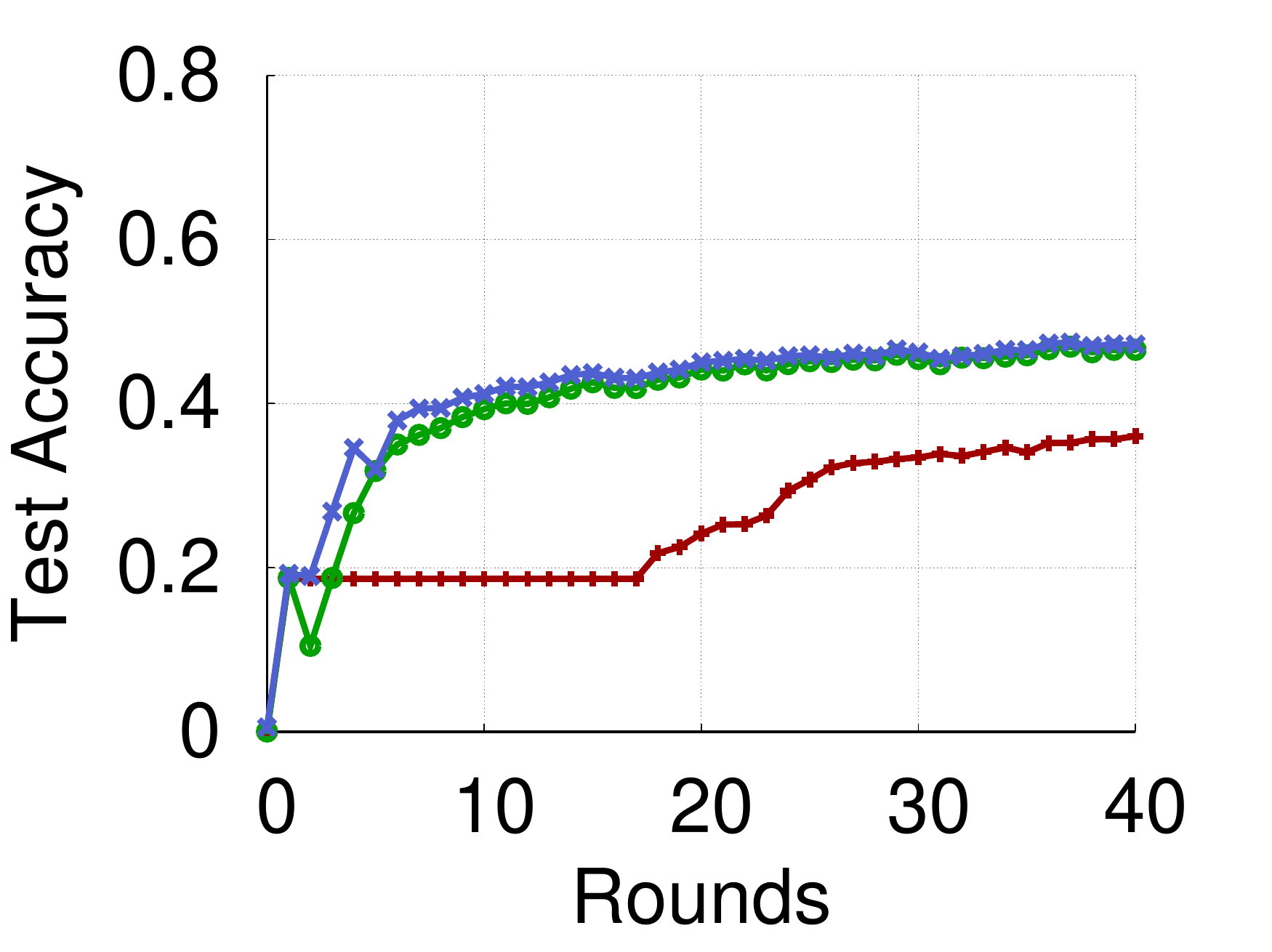}
			\caption{Shakespeare}
		\end{subfigure}
		\caption{Testing accuracy of \method{}, \fedprox{} and \fedave{} on various datasets using non-linear model (LSTM). \method{} can reach higher level of accuracy than the others.}
		\label{fig:compare_acc_all_methods_nonlinear}
	\end{minipage}
	\vspace{-0.2in}
\end{figure*}
\begin{figure*}[t]
	\centering
	\begin{subfigure}[b]{0.95\linewidth}
		\includegraphics[width=\linewidth]{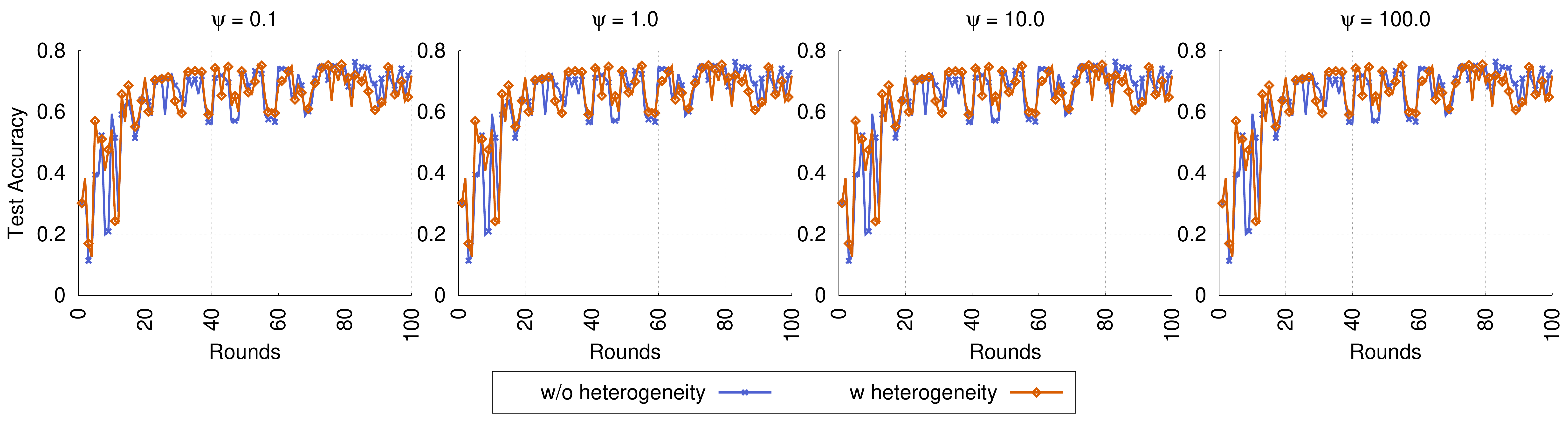}
		\caption{Synthetic\_1\_1}
	\end{subfigure}
	\begin{subfigure}[b]{0.95\linewidth}
		\includegraphics[width=\linewidth]{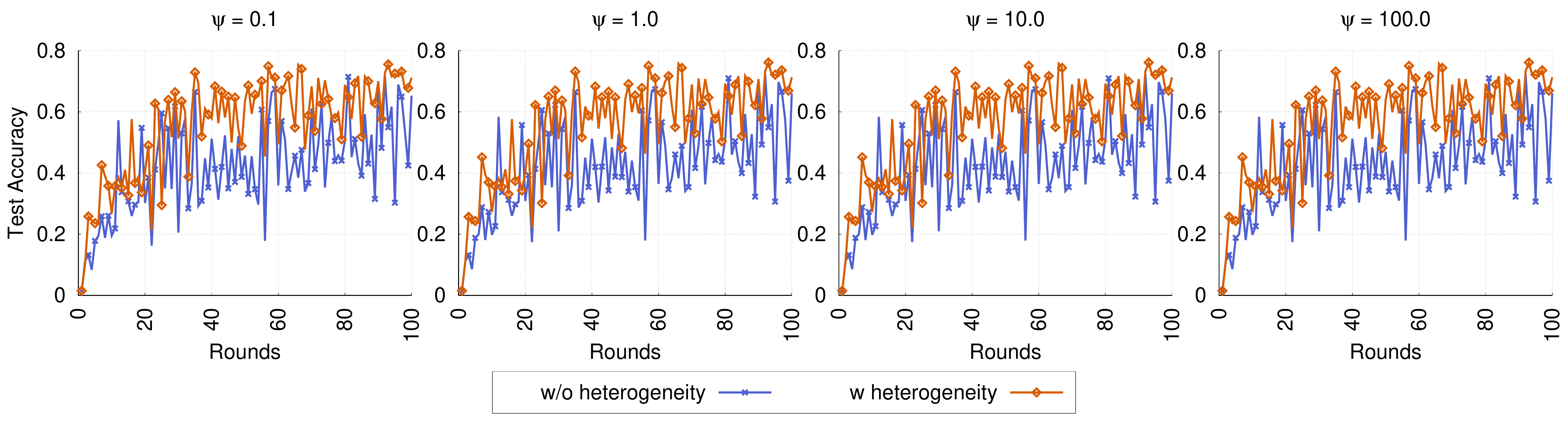}
		\caption{FEMNIST}
	\end{subfigure}
	\caption{Accuracy of \method{} with and without heterogeneity consideration. Heterogeneity-aware \method{} avoids major drops of accuracy between iterations and is more robust than vanilla \method{}.}
	\label{fig:heterogeneity}
	\vspace{-0.2in}
\end{figure*}

\subsection{Experimental settings}
We first describe our setup of datasets, compared algorithms, testing environment and how statistical and system heterogeneity is simulated. We adopt closely the setup in a very recent work \cite{tian2019} on \fedprox{} and provide details of their setup and the changes we made here for completeness.

\textbf{Dataset.} We use the a standard set of datasets used in multiple other works on federated learning \cite{tian2019,shamir14}. Particularly, we use 10-class \emph{MNIST} \cite{lecun1998gradient}, 62 class Federated Extended MNIST (\emph{FEMNIST}) \cite{cohen2017emnist}, and synthetic datasets \cite{shamir14,tian2019} to study with a multinomial logistic regression model, which extends the binary logistic regression model to multi-class scenarios and uses a different linear predictor function for each class to predict the probability that an observation belongs to that class. The synthetic datasets are generated with Gaussian distributions which are parameterized with a set of control parameters to vary the level of heterogeneity (see \cite{shamir14,tian2019} for more details). \emph{Synthetic\_iid} and \emph{Synthetic\_1\_1} denote two datasets with no heterogeneity (i.i.d. distribution of data) and high heterogeneity, respectively. For non-convex setting, similarly to \cite{mcmahan17,tian2019}, we consider a text sentiment analysis task on tweets using \emph{Sent140} \cite{go2009twitter} dataset and next-character prediction task on the dataset of \emph{The Complete Works of William Shakespeare} \cite{mcmahan17}. For MNIST, FEMNIST, sent140, and Shakespeare, we consider 1000, 200, 143, 772 devices, respectively. Particularly, for MNIST and FEMNIST datasets, the data is distributed on each device following a power law under the constraint that each device gets images from only two digits. For Sent140, each twitter account corresponds to one device, while in Shakespeare, each speaking role corresponds to one device. 

\textbf{Compared algorithms.}  We compare \method{} with current state-of-the-art algorithms in the federated learning setting, including the recent \fedprox{} \cite{tian2019} and the original \fedave{} \cite{mcmahan17}. For both \method{}, \fedprox{} and \fedave{}, we use $k = 10$ devices in each round of optimization and investigate the effects of $K$ on performance in a later set of experiments. For \fedprox{}, we set $\mu = 1, 1, 1, 0.001, \text{and } 0.01$, for 5 datasets respectively, as suggested in the original paper \cite{tian2019}. For our algorithm \method{}, we apply a similar line search on $\mu \in \{ 10^{-4}, 10^{-3}, 10^{-2}, 10^{-1}, 1 \}$ and $\psi \in \{ 10^{-1}, 1, 10, 10^2 \}$ when \method{} with heterogeneity consideration is tested. Here, we consider the versions of \method{} that only samples one set of devices in each round of optimization for communication efficiency, i.e., we use the aggregation scheme in (\ref{eq:aggregate_2}) and (\ref{eq:aggregate_hetero}). Thus, the communication cost of \method, \fedave{} and \fedprox{} are the same.

\textbf{Computation and communication heterogeneity simulation.} For all algorithms, we simulate the computation and communication heterogeneity by allowing each device to pick a random number between $1$ and $20$ to be the number of gradient descent steps that the device is able to perform when selected. We initialize the same seed in all the compared algorithms to make sure that these numbers of gradient descents are consistent on all the algorithms. For \fedprox{} and \fedave{}, the received parameters from local devices in every round are simply averaged to get the new set of global model parameters.

\textbf{Environment.} We performed all experiment on a 8x2080Ti GPU cluster using TensorFlow~\cite{abadi2016tf} framework. Our codebase is based on the publicly available implementation of FedProx \cite{tian2019} approach \footnote{\url{https://github.com/litian96/FedProx}}. For each dataset, we use stochastic gradient descent (SGD) as a local solver.

\subsection{Experimental results}

\subsubsection{Quantifying the effectiveness of the proposed aggregation rule}
We first compare our new aggregation rule with the simple averaging in \fedprox{} (similarly in \fedave{}). We vary $\mu$ with values from the set $\{ 10^{-4}, 10^{-3}, 10^{-2}, 10^{-1}, 1 \}$ in both \fedprox{} and \method{}, and fix $\psi = 0$ in \method{}. The training loss and test accuracy on the first real dataset MNIST are shown in Fig.~\ref{fig:aggregation_effective}.

From Fig.~\ref{fig:aggregation_effective}, we observe the better performance of our proposed aggregation rule compared to that of simple averaging in \fedprox{} (and similarly in \fedave{}). Specifically, with \method{}, the loss value is always smaller than that of \fedprox{} and its accuracy is higher than that of \fedprox{} at the same time. This is especially significant in early iterations, showing faster convergence rate of \method. Our results prove the better effectiveness of our proposed aggregation scheme that principally aims at maximizing a lower-bound of loss decrease in every iteration (\ref{eq:loss_thm1}).

Moreover, the better performance of our aggregation rule is more compelling with smaller values of $\mu$. This observation again verifies the critical role of our lower-bound in (\ref{eq:loss_thm1}) and our goal of maximizing it. Since maximizing the lower-bound leads to our approach of maximizing $E \Big[ \sum_{k \in S_t} \langle \nabla f(\mathbf{w}^t),\nabla F_k(\mathbf{w}^t)\rangle \Big]$, which is weighted by $\frac{1}{\mu}$ in (\ref{eq:loss_thm1}), with smaller $\mu$, the results of maximizing $E \Big[ \sum_{k \in S_t} \langle \nabla f(\mathbf{w}^t),\nabla F_k(\mathbf{w}^t)\rangle \Big]$ have bigger impact in maximizing the lower-bound. This observation of having better actual loss values draws a strong correlation between our lower-bound in (\ref{eq:loss_thm1}) and the actual loss decrease and maximizing the lower-bound is sensible.

%\begin{figure*}[h!]
%	\centering
%	\begin{minipage}{0.48\linewidth}
%		\centering
%		\includegraphics[width=\linewidth]{figs/acc_vs_arch.pdf}
%		\hspace{-0.2in}
%		\caption{Effectiveness of \method{}  for Convolutional neural networks. We also compare with another MLP with identical depth over MNIST dataset and $\mu=0.01$.}
%		\label{fig:acc_vs_arch}
%	\end{minipage}
%	\hspace{0.1in}
%	\begin{minipage}{0.48\linewidth}
%		\centering
%		\includegraphics[width=\linewidth]{figs/acc_vs_k.pdf}
%		\hspace{-0.2in}
%		\caption{With increase in number of clients in each round, \method{} converges faster and stabilizes quicker than \fedprox{}. We use MNIST dataset with a 3-layer CNN and $\mu = 0.01$. }
%		\label{fig:acc_vs_k}
%	\end{minipage}
%	\vspace{-0.2in}
%\end{figure*}

\noindent \textbf{Experiments with different neural network models.} In earlier experiments, we used a multinomial logistic regression model. Now we compare the performance of \method{} and \fedprox{} when using a Convolutional Neural Network (CNN) or Multi-Layer Perceptron (MLP) with 3 layers each. The results are illustrated in Fig.~\ref{fig:acc_vs_arch}. We find that \method{} converges faster and is much more stable compared to \fedprox{}.

\noindent \textbf{Experiments with different number of devices in each round.} We present the results with a varying number of devices participating in each round in Fig.~\ref{fig:acc_vs_k}. As expected, more devices make convergence stable and fast. However, we find this effect significantly better with \method{} compared to \fedprox{}, thanks to our aggregation scheme. With a small number of devices in each round, \method{} is quite similar to \fedprox{} since our aggregation scheme becomes closer to simple averaging.

\noindent \textbf{Experiments with different non-IID settings.} We simulate various non-IID scenarios on the MNIST dataset by only assigning random images from a fixed number of different digits to each device, i.e., 1, 2, 5, 10. For example, in the most extreme case, each device only has random images from only one digit. The results are demonstrated in Fig.~\ref{fig:acc_vs_imagesperdevice} and show a recurring observation that \method{} outperforms \fedprox{}, especially in the extreme cases of non-IID (common in reality).

\subsubsection{Comparisons on various datasets and models} We compare \method{} with \fedprox{} and \fedave{} algorithms. Figs.~\ref{fig:compare_loss_all_methods} and \ref{fig:compare_acc_all_methods} present the training loss and test accuracy of all the algorithms on linear model (multinomial logistic regression) and Figs.~\ref{fig:compare_loss_all_methods_nonlinear} and \ref{fig:compare_acc_all_methods_nonlinear} report results for non-linear model (LSTM). It is evident that \method{} consistently outperforms \fedprox{} and \fedave{} in terms of both reducing loss and improving accuracy. For example, on the Synthetic\_1\_1 dataset, \method{} is able to reach a low loss value and high accuracy level in only within 20 iterations while the other two methods never reach that level within 100 iterations and seem to converge at much higher loss and lower accuracy. On the other datasets, \method{} reduces loss value (and also increasing accuracy) faster than both \fedprox{} and \fedave{}, and can even reach lower loss and higher accuracy level than the other two algorithms.
\begin{table}[h]
	%	\small
	\centering	
	\caption{Number of rounds of each method to reach a certain accuracy level on each dataset (Note that on Shakespeare, \fedave{} failed to reach the given accuracy within 40 rounds).}
	\label{table:num_rounds}
	\begin{tabular}{l| r|r|r|r}
		\toprule
		\textbf{Methods} & Accuracy & \method{} & \fedprox & \fedave{} \\
		\midrule
		Synthetic\_iid & $70\%$ & \textbf{50} & 57 & 113 \\
		Synthetic\_1\_1 & $70\%$ & \textbf{19} & 154 & 177 \\
		MNIST & $80\%$ & \textbf{11} & 25 & 25 \\
		FEMNIST & $65\%$ & \textbf{34} & 58 & 86 \\ 
		Sent140 & $65\%$ & \textbf{31} & 132 & 82 \\
		Shakespeare & $45\%$ & \textbf{20} & 25 & \_\\
		\bottomrule
	\end{tabular}
	\vspace{-0.2in}
\end{table}

In Table~\ref{table:num_rounds}, we report the number of optimization rounds that each algorithm needs to perform in order to reach a certain accuracy level (this is chosen based on the maximum accuracy that all three algorithms can reach on each dataset). We see that, usually \method{} only requires half number of rounds taken by \fedprox{} and\fedave{} to reach the same level of accuracy. For example, on Synthetic\_1\_1 dataset, \method{} only needs 19 rounds while \fedprox{} and \fedave{} require 154 and 177 rounds respectively. One exception is on Synthetic\_iid where data is independent and identically distributed across different devices, however, \method{} still need fewer rounds than \fedprox{} and \fedave{}. Note that due to computation heterogeneity, even on Synthetic\_iid, \fedave{} performs poorly compared to \fedprox{} and \method{} which directly address heterogeneity. These results again verify the faster convergence rate of \method{} compared to \fedprox{} and \fedave{}.

\subsubsection{\method{} with and without communication and computation heterogeneity consideration}
In this last set of experiments, we compare \method{} with different aggregation rules, i.e., (\ref{eq:aggregate_2}) and (\ref{eq:aggregate_hetero}) which are corresponding to before and after taking into account the heterogeneity of communication and computation respectively. Fig.~\ref{fig:heterogeneity} shows the test accuracy of these two variants on Synthetic\_1\_1 and EMNIST, where the performance of \method{} varies the most (Fig.~\ref{fig:compare_acc_all_methods}) and with different values of $\psi$ which controls how much heterogeneity contributes in computing aggregation weight of each local update in (\ref{eq:aggregate_hetero}). The results show that by taking into account the inherent heterogeneity, \method{} is more stable than the other variant. In particular, with heterogeneity, \method{} is able to avoid most major drops in accuracy and stays at high accuracy level toward later iterations without any significant fluctuations. On the other hand, the vanilla \method{} can reach high accuracy but fluctuates widely even in later iterations. In addition, from Fig.~\ref{fig:heterogeneity}, $\psi$ can take value in a wide range, i.e., $[0.1, 10]$ and still helps stabilize \method{} well.

\section{Conclusion}
In this work, we have introduced \method{} - a fast-convergent federated learning algorithm, and shown that \method{} theoretically achieves a near-optimal possible lower-bound for the overall loss decrease at every round of communication/optimization. \method{} encloses a novel adaptive aggregation scheme that takes into account both statistical and system heterogeneity inherent in the modern networking environments of massively distributed mobile devices. More importantly, we have shown that across different tasks and datasets, \method{} significantly reduces the number of rounds to reach a certain level of loss value and accuracy. 

For future work, a promising direction is to study a device selection methodology that couples decisions across multiple time periods to bring greater performance gains in the long term. This involves deriving new lower-bound that reflects the performance after a number of optimization rounds and taking into account the communication and computation heterogeneity in all those rounds.

%\section*{Acknowledgment}
%The authors would like to thank...

\ifCLASSOPTIONcaptionsoff
  \newpage
\fi

\bibliographystyle{IEEEtran}
\bibliography{fedlearn}

\appendix
\label{appendix}

\subsection{Proof of Theorem~\ref{theo:loss}}
\label{proof:loss_theorem}

\begin{proof}
	From the $L$-Lipschitz continuity of $f$, we have
	\vspace{-0.03in}
	\begin{align}
		f(\mathbf{w}^{t+1}) \leq f(\mathbf{w}^{t}) + \langle \nabla f(\mathbf{w}^t), \mathbf{w}^{t+1} - \mathbf{w}^t \rangle + \frac{L}{2} \norm{\mathbf{w}^{t+1} - \mathbf{w}^t}^2
		\label{eq:loss}
	\end{align}
	\vspace{-0.23in}
	
	We will separately bound the last two terms on the right-hand side of the above inequality:
	
	\emph{$\bullet$ Bounding $\norm{\mathbf{w}^{t+1} - \mathbf{w}^t}$}: Let $\mathbf{\hat w}^{t+1}_k = \arg\min_\mathbf{w} h_k(\mathbf{w}, \mathbf{w}^t)$. Due to the $\mu'$-strong convexity of $h_k(\mathbf{w}, \mathbf{w}^t)$ and the $\gamma$-inexact local solver assumption for $\mathbf{w}^{t+1}_k$, we have:
	\vspace{-0.03in}
	\begin{align}
		\norm{\mathbf{\hat w}^{t+1}_k - \mathbf{w}^{t+1}_k} & \leq \frac{1}{\mu'} \norm{\nabla h(\mathbf{\hat w}^{t+1}_k, \mathbf{w}^t) - \nabla h(\mathbf{w}^{t+1}_k, \mathbf{w}^t)} \nonumber \\ & \leq \frac{\gamma}{\mu'} \norm{\nabla F_k(\mathbf{w}^t)},
	\end{align}
	and, similarly,
	\vspace{-0.03in}
	\begin{align}
	\norm{\mathbf{\hat w}_k^{t+1} - \mathbf{w}^t} \leq \frac{1}{\mu'} \norm{\nabla F_k(\mathbf{w}^t)}.
	\end{align}
	Hence, by the triangle inequality and $B$-dissimilarity of $\nabla F_k$, we obtain:
	\vspace{-0.03in}
	\begin{align}
	\norm{\mathbf{w}_k^{t+1} - \mathbf{w}^t} \leq \frac{1+\gamma}{\mu'} \norm{\nabla F_k(\mathbf{w}^t)} \leq \frac{B(1+\gamma)}{\mu'} \norm{\nabla f(\mathbf{w}^t)}.
	\label{eq:diss}
	\end{align}
	\vspace{-0.27in}
	
	\noindent Now, noting $\mathbf{w}^{t+1} = \frac{1}{K}\sum_{k \in S_t} \mathbf{w}^{t+1}_k$, we can write
	\vspace{-0.03in}
	\begin{align}
	\norm{\mathbf{w}^{t+1} - \mathbf{w}^t}^2 & \leq \Big(\frac{1}{K} \sum_{k \in S_t} \norm{\mathbf{w}^{t+1}_k - \mathbf{w}^t} \Big)^2 \nonumber \\
	& \leq \frac{B^2(1+\gamma)^2}{\mu'^2} \norm{\nabla f(\mathbf{w}^t)}^2,
	\label{eq:b1}
	\end{align}
	where the first inequality follows from the Cauchy-Schwarz's inequality, and the second follows from applying (\ref{eq:diss}) to each $k$ in the sum.
	
	\emph{$\bullet$ Bounding  $\langle \nabla f(\mathbf{w}^t), \mathbf{w}^{t+1} - \mathbf{w}^t \rangle$}: By definition of the aggregation step for $\mathbf{w}^{t+1}$, we can write
	\vspace{-0.03in}
	\begin{equation}
	\langle \nabla f(\mathbf{w}^t), \mathbf{w}^{t+1} - \mathbf{w}^t \rangle = \frac{1}{K} \sum_{k \in S_t} \langle \nabla f(\mathbf{w}^t), \mathbf{w}^{t+1}_k - \mathbf{w}^t \rangle.
	\label{eq:agg_inp}
	\end{equation}
	For each term in the sum, we can express
	\vspace{-0.03in}
	\begin{align*}
	\mathbf{w}^{t+1}_k - \mathbf{w}^t = &-\frac{1}{\mu}\nabla F_k(\mathbf{w}^t) + \frac{1}{\mu} (\nabla F_k(\mathbf{w}^t) - \nabla F_k (\mathbf{w}^{t+1}_k)) \nonumber \\
	& + \frac{1}{\mu}(\nabla  F_k (\mathbf{w}^{t+1}_k) + \mu(\mathbf{w}^{t+1}_k - \mathbf{w}^t)).
	\end{align*}
	\vspace{-0.25in}
	
	\noindent Thus,
	\vspace{-0.03in}
	\begin{align}
	\langle \nabla f(\mathbf{w}^t), & \mathbf{w}^{t+1}_k - \mathbf{w}^t \rangle = -\frac{1}{\mu} \langle \nabla f(\mathbf{w}^t),\nabla F_k(\mathbf{w}^t)\rangle \nonumber \\
	& + \frac{1}{\mu} \langle \nabla f(\mathbf{w}^t), \; \nabla F_k(\mathbf{w}^t) - \nabla F_k (\mathbf{w}^{t+1}_k)\rangle \nonumber \\ 
	& + \frac{1}{\mu}\langle \nabla f(\mathbf{w}^t), \; \nabla  F_k (\mathbf{w}^{t+1}_k) + \mu(\mathbf{w}^{t+1}_k - \mathbf{w}^t) \rangle \nonumber \\
	& \qquad \qquad \quad \leq -\frac{1}{\mu}\langle \nabla f(\mathbf{w}^t),\nabla F_k(\mathbf{w}^t)\rangle \nonumber \\
	& + \frac{1}{\mu} \norm{\nabla f(\mathbf{w}^t)} \norm{\nabla F_k(\mathbf{w}^t) - \nabla F_k (\mathbf{w}^{t+1}_k)} \nonumber \\
	& + \frac{1}{\mu} \norm{\nabla f(\mathbf{w}^t)} \norm{\nabla  F_k (\mathbf{w}^{t+1}_k) + \mu(\mathbf{w}^{t+1}_k - \mathbf{w}^t)}, \nonumber
	\end{align}
	where the inequality follows again from Cauchy-Schwarz. Noting that $\norm{\nabla F_k(\mathbf{w}^t) - \nabla F_k (\mathbf{w}^{t+1}_k)} \leq L \norm{\mathbf{w}^{t+1}_k - \mathbf{w}^t}$ by Assumption 1, and that $\nabla  F_k (\mathbf{w}^{t+1}_k) + \mu(\mathbf{w}^{t+1}_k - \mathbf{w}^t) = \nabla h(\mathbf{w}^{t+1}_k, \mathbf{w}^t)$ by definition, we have
	\vspace{-0.05in}
	\begin{align}
	\langle \nabla f(\mathbf{w}^t), \mathbf{w}^{t+1}_k - \mathbf{w}^t \rangle \leq -\frac{1}{\mu}\langle \nabla f(\mathbf{w}^t),\nabla F_k(\mathbf{w}^t)\rangle \nonumber \\ + \frac{LB(1+\gamma)}{\mu \mu'} \norm{\nabla f(\mathbf{w}^t)}^2 + \frac{B \gamma}{\mu} \norm{\nabla f(\mathbf{w}^t)}^2,
	\label{eq:b2_1}
	\end{align}
	where we have applied \eqref{eq:diss} to the middle term, and Assumptions 4\&2 to the last term on the right hand side. Combining this with \eqref{eq:agg_inp}, we have
	\vspace{-0.03in}
	\begin{align}
	\langle \nabla f(\mathbf{w}^t), & \mathbf{w}^{t+1} - \mathbf{w}^t \rangle = \frac{1}{K} \sum_{k \in S_t} \langle \nabla f(\mathbf{w}^t), \mathbf{w}^{t+1}_k - \mathbf{w}^t \rangle \nonumber \\
	& \leq -\frac{1}{K\mu}\sum_{k \in S_t}\langle \nabla f(\mathbf{w}^t),\nabla F_k(\mathbf{w}^t)\rangle \nonumber \\
	& \qquad + \frac{B}{\mu} \Big (\frac{L(\gamma+1)}{\mu'}+\gamma\Big) \norm{\nabla f(\mathbf{w}^t)}^2.
	\label{eq:b2}
	\end{align}
	
	Substituting (\ref{eq:b1}) and (\ref{eq:b2}) into (\ref{eq:loss}) and taking the expectation, we obtain
	\vspace{-0.03in}
	%	\begin{align}
	%		f(\mathbf{w}^{t+1}) \leq f(\mathbf{w}^{t}) - \frac{1}{K\mu} \sum_{k \in S_t}\langle \nabla f(\mathbf{w}^t),\nabla F_k(\mathbf{w}^t)\rangle \nonumber \\
	%		+ B \Bigg (\frac{L(\gamma+1)}{\mu \mu'}+\frac{\gamma}{\mu} + \frac{BL(1+\gamma)^2}{2 \mu'^2} \Bigg) \norm{\nabla f(\mathbf{w}^t)}^2,
	%	\end{align}
	%	and
	\begin{align}
		\mathbb{E}&[f(\mathbf{w}^{t+1})] \leq f(\mathbf{w}^{t}) -\frac{1}{K\mu} \mathbb{E} \Bigg[\sum_{k \in S_t}\langle \nabla f(\mathbf{w}^t),\nabla F_k(\mathbf{w}^t)\rangle \Bigg] \nonumber \\
		& + B \Bigg (\frac{L(\gamma+1)}{\mu \mu'}+\frac{\gamma}{\mu} + \frac{BL(1+\gamma)^2}{2 \mu'^2} \Bigg) \norm{\nabla f(\mathbf{w}^t)}^2, \nonumber
	\end{align}
	where the first and last terms on the right hand side are not written in expectation as they do not depend on the selection of devices in round $t$.
\end{proof}
\vspace{-0.05in}

\subsection{Proof of Proposition~\ref{theo:loss_improved}}

\begin{proof}
	The key difference from Theorem~\ref{theo:loss}'s proof is in the decomposition of $\langle \nabla f(\mathbf{w}^t), \mathbf{w}^{t+1} - \mathbf{w}^t \rangle$ in \eqref{eq:b2}. In this case, we write
	\vspace{-0.03in}
	\begin{align}
		\langle \nabla f(\mathbf{w}^t), &\mathbf{w}^{t+1} - \mathbf{w}^t \rangle = \frac{1}{K} \Big [ \sum_{k \in S^+_t} \langle \nabla f(\mathbf{w}^t), \mathbf{w}^{t+1}_k - \mathbf{w}^t \rangle \nonumber \\ & + \sum_{k \in S^-_t} \langle \nabla f(\mathbf{w}^t), \mathbf{w}^t - \mathbf{w}^{t+1}_k \rangle \Big ],
		\label{eq:spm}
	\end{align}
	where $S^+_t = \{k \in S_t : \langle \nabla f(\mathbf{w}^t),\nabla F_k(\mathbf{w}^t)\rangle \ge 0\}$ and $S^-_t = \{k \in S_t : \langle \nabla f(\mathbf{w}^t),\nabla F_k(\mathbf{w}^t)\rangle < 0\}$. For $k \in S^+_t$, the derivation follows (\ref{eq:b2_1}). On the other hand, for $k \in S^-_t$, 
	\vspace{-0.03in}
	\begin{align}
	& \langle \nabla f(\mathbf{w}^t), \mathbf{w}^t - \mathbf{w}^{t+1}_k \rangle \leq \frac{1}{\mu} \langle \nabla f(\mathbf{w}^t),\nabla F_k(\mathbf{w}^t) \rangle \nonumber \\ & \qquad\qquad- \frac{1}{\mu} \langle \nabla f(\mathbf{w}^t), \; \nabla F_k(\mathbf{w}^t) - \nabla F_k (\mathbf{w}^{t+1}_k)\rangle \nonumber \\ 
	& \qquad \qquad - \frac{1}{\mu}\langle \nabla f(\mathbf{w}^t), \; \nabla  F_k (\mathbf{w}^{t+1}_k) + \mu(\mathbf{w}^{t+1}_k - \mathbf{w}^t) \rangle \nonumber \\
	& \qquad \qquad\qquad \qquad \quad \leq -\frac{1}{\mu} |\langle \nabla f(\mathbf{w}^t), \nabla F_k(\mathbf{w}^t)\rangle|  \nonumber \\ & \qquad \qquad + \frac{1}{\mu} \norm{\nabla f(\mathbf{w}^t)} \norm{\nabla F_k(\mathbf{w}^t) - \nabla F_k (\mathbf{w}^{t+1}_k)} \nonumber \\
	& \qquad \qquad + \frac{1}{\mu} \norm{\nabla f(\mathbf{w}^t)} \norm{\nabla  F_k (\mathbf{w}^{t+1}_k) + \mu(\mathbf{w}^{t+1}_k - \mathbf{w}^t)} \nonumber\\
	& \qquad \qquad\qquad \qquad \quad \leq -\frac{1}{\mu} |\langle \nabla f(\mathbf{w}^t),\nabla F_k(\mathbf{w}^t)\rangle| \nonumber \\
	& \qquad \qquad + \frac{LB(1+\gamma)}{\mu \mu'} \norm{\nabla f(\mathbf{w}^t)}^2 + \frac{B \gamma}{\mu} \norm{\nabla f(\mathbf{w}^t)}^2. \nonumber
	\end{align}
	Substituting these expressions in \eqref{eq:spm} gives the result.
\end{proof}
\vspace{-0.05in}

\subsection{Proof of Lemma~\ref{lem:square}}
We sequentially prove the two statements in the following:
\begin{proof}[Proof of Eq.~\ref{lem:s1}]
	We expand $\sum_{k \in S^{t}_1} \langle \nabla F_k(\mathbf{w}^t), \nabla_1 f(\mathbf{w}^t) \rangle^2$ as follows:
	\vspace{-0.03in}
	\begin{align*}
	&\sum_{k \in S^{t}_1} \langle \nabla F_k(\mathbf{w}^t), \nabla_1 f(\mathbf{w}^t) \rangle^2\nonumber \\  & = \frac{1}{K^2} \sum_{k \in S^{t}_1} \Big( \sum_{k' \in S^{t}_1} \langle \nabla F_k(\mathbf{w}^t), \nabla F_{k'}(\mathbf{w}^t) \rangle \Big)^2 \\
	& = \frac{1}{K^2} \sum_{k, k', k'' \in S^{t}_1} \langle \nabla F_k(\mathbf{w}^t), \nabla F_{k'}(\mathbf{w}^t) \rangle \langle \nabla F_k(\mathbf{w}^t), \nabla F_{k''}(\mathbf{w}^t) \rangle
	\end{align*}
	Since $|S^t_1| = K$, the summation in the last equality has $K^3$ terms. Across all possible multisets $S^t_1$, there are $N^3$ possible combinations of $k, k', k''$. Since device selection in Algorithm \ref{alg:fexprox_prob_lb3} occurs uniformly at random, each combination $k, k', k''$ has the same probability of appearing in the summation. Therefore, we can write the expectation as a summation over all combinations of three devices from $[N] = \{1,...,N\}$, and simplify the result as follows:
	\begin{align*}
	& \mathbb{E} \Big[ \sum_{k \in S^{t}_1} \langle \nabla F_k(\mathbf{w}^t), \nabla_1 f(\mathbf{w}^t) \rangle^2 \Big] \nonumber \\ & = \frac{K^3}{K^2 N^3} \sum_{k, k', k''} \langle \nabla F_k(\mathbf{w}^t), \nabla F_{k'}(\mathbf{w}^t) \rangle \langle \nabla F_k(\mathbf{w}^t), \nabla F_{k''}(\mathbf{w}^t) \rangle \\
	& = \frac{K}{N^3} \sum_{k \in [N]} \Big( \sum_{k' \in [N]} \langle \nabla F_k(\mathbf{w}^t), \nabla F_{k'}(\mathbf{w}^t) \rangle \Big)^2 \nonumber \\
	& = \frac{K}{N} \sum_{k \in [N]} \Big( \langle \nabla F_k(\mathbf{w}^t), \; \frac{1}{N} \sum_{k' \in [N]} \nabla F_{k'}(\mathbf{w}^t) \rangle \Big)^2 \\
	& = \frac{K}{N} \sum_{k \in [N]} \Big( \langle \nabla F_k(\mathbf{w}^t), \nabla f(\mathbf{w}^t) \rangle \Big)^2,
	\end{align*}
	where the last step follows from the definition of $\nabla f(\mathbf{w}^t) = \frac{1}{N} \sum_{k \in [N]} \nabla F_{k}(\mathbf{w}^t)$.
\end{proof}

\begin{proof}[Proof of Eq.~\ref{lem:s2}] By definition of $\nabla_2 f(\mathbf{w}^t)$, we have
	\vspace{-0.03in}
	\begin{align*}
	&\sum_{k' \in S^{t}_2} \langle \nabla F_{k'}(\mathbf{w}^t), \nabla_2 f(\mathbf{w}^t) \rangle \nonumber \\& \qquad = \frac{1}{K} \sum_{k', k'' \in S^{t}_2} \langle \nabla F_{k'}(\mathbf{w}^t), \nabla F_{k''}(\mathbf{w}^t) \rangle.
	\end{align*}
	Then, similar to the proof of Eq.~\ref{lem:s1}, we can write the expectation as a summation over all possible combinations of device pairs, and simplify:
	\vspace{-0.03in}
	%In the last equality, we have $K^2$ terms in the summation. Consider all possible subsets $S^t_2$ of $K$ devices, there are $N^2$ possible combinations of $k', k''$ and each of those possibilities have the same chance of appearing in the summation. Thus,
	\begin{align*}
	&\mathbb{E} \Big[ \sum_{k' \in S^{t}_2} \langle \nabla F_{k'}(\mathbf{w}^t), \nabla_2 f(\mathbf{w}^t) \rangle \Big]\nonumber \\&  = \frac{K^2}{KN^2} \sum_{k', k'' \in [N]} \langle \nabla F_{k'}(\mathbf{w}^t), \nabla F_{k''}(\mathbf{w}^t) \rangle \\
	& = \frac{K}{N} \sum_{k' \in [N]} \langle \nabla F_{k'}(\mathbf{w}^t), \; \frac{1}{N} \sum_{k'' \in [N]} \nabla F_{k''}(\mathbf{w}^t) \rangle \nonumber \\
	& = \frac{K}{N} \sum_{k' \in [N]} \langle \nabla F_{k'}(\mathbf{w}^t), f(\mathbf{w}^t) \rangle \leq \frac{K}{N} \sum_{k' \in [N]} | \langle \nabla F_{k'}(\mathbf{w}^t), f(\mathbf{w}^t) \rangle |.
	\end{align*}
	That complete the proof.
\end{proof}
\vspace{-0.03in}

\subsection{Proof of Theorem~\ref{theo:loss_lb}}
\label{proof:theo_loss_lb}
\begin{proof}
	As in Theorem \ref{theo:loss}, we begin with the $L$-Lipschitz inequality for $f(\mathbf{w}^{t+1})$ given in \eqref{eq:loss}, and bound the last two terms on the right-hand side:
	%	\begin{align}
	%		f(\mathbf{w}^{t+1}) \leq f(\mathbf{w}^{t}) + \langle \nabla f(\mathbf{w}^t), \mathbf{w}^{t+1} - \mathbf{w}^t \rangle \nonumber \\ + \frac{L}{2} \norm{\mathbf{w}^{t+1} - \mathbf{w}^t}^2.
	%	\end{align}
	
	\emph{$\bullet$ Bounding $\norm{\mathbf{w}^{t+1} - \mathbf{w}^t}$}: In \eqref{eq:aggregate}, define
	\vspace{-0.03in}
	\begin{equation}
	\label{eq:phatkt}
	{\hat P}^t_k = \frac{\langle \nabla F_k(\mathbf{w}^t), \nabla_1 f(\mathbf{w}^t) \rangle }{\sum_{k' \in S^{t}_2} \langle \nabla F_{k'}(\mathbf{w}^t), \nabla_2 f(\mathbf{w}^t) \rangle },
	\end{equation}
	i.e., an approximation of the \lb-near-optimal selection probability in \eqref{eq:plb}. Following the procedure for this bound in Theorem \ref{theo:loss}, for the update rule \eqref{eq:aggregate} of \method{}, we can write
	\vspace{-0.03in}
	\begin{align}
	\norm{\mathbf{w}^{t+1} - \mathbf{w}^t}^2 & \leq \Big ( \sum_{k \in S^{t}_1} {\hat P}^t_k \norm{\mathbf{w}^{t+1}_k - \mathbf{w}^t} \Big )^2 \nonumber \\ & \leq \Big ( \sum_{k \in S^{t}_1} {\hat P}^t_k \Big )^2 \frac{B^2(1+\gamma)^2}{\mu'^2} \norm{\nabla f(\mathbf{w}^t)}^2.
	\label{eq:lb_loss_1}
	\end{align}
	\vspace{-0.25in}
	%	where ${\hat P}^t_k = \frac{\langle \nabla F_k(\mathbf{w}^t), \nabla_1 f(\mathbf{w}^t) \rangle }{\sum_{k' \in S^{t}_2} \langle \nabla F_{k'}(\mathbf{w}^t), \nabla_2 f(\mathbf{w}^t) \rangle }$.
	
	\emph{$\bullet$ Bounding  $\langle \nabla f(\mathbf{w}^t), \mathbf{w}^{t+1} - \mathbf{w}^t \rangle$}: Similar to the procedure for this bound in Theorem \ref{theo:loss}, we can write
	\vspace{-0.03in}
	%	\begin{align}
	%		\langle \nabla f(\mathbf{w}^t), \mathbf{w}^{t+1}_k - \mathbf{w}^t \rangle \leq -\frac{1}{\mu}\langle \nabla f(\mathbf{w}^t),\nabla F_k(\mathbf{w}^t)\rangle \nonumber \\
	%		+ \frac{LB(1+\gamma)}{\mu \mu'} \norm{\nabla f(\mathbf{w}^t)}^2 + \frac{B \gamma}{\mu} \norm{\nabla f(\mathbf{w}^t)}^2,
	%	\end{align}
	%	then,
	\begin{align}
	\langle \nabla f(\mathbf{w}^t), & \mathbf{w}^{t+1} - \mathbf{w}^t \rangle = \sum_{k \in S^{t}_1} {\hat P}^t_k \langle \nabla f(\mathbf{w}^t), \mathbf{w}^{t+1}_k - \mathbf{w}^t \rangle \nonumber \\
	& \leq - \frac{1}{\mu} \sum_{k \in S^{t}_1} {\hat P}^t_k \langle \nabla F_k(\mathbf{w}^t), \nabla_1 f(\mathbf{w}^t) \rangle \nonumber \\ & + \sum_{k \in S^{t}_1} {\hat P}^t_k \frac{B}{\mu} \Big (\frac{L(\gamma+1)}{\mu'}+\gamma\Big) \norm{\nabla f(\mathbf{w}^t)}^2,
	\label{eq:lb_loss_2}
	\end{align}
	where the equality follows from the \method{} aggregation, and the inequality follows from \eqref{eq:b2_1}.
	
	Now, substituting \eqref{eq:lb_loss_1} and \eqref{eq:lb_loss_2} into \eqref{eq:loss}, we have
	\vspace{-0.03in}
	\begin{align}
	f(\mathbf{w}^{t+1}) \leq f(\mathbf{w}^{t}) - \frac{1}{\mu} \sum_{k \in S^{t}_1} {\hat P}^t_k \langle \nabla F_k(\mathbf{w}^t), \nabla_1 f(\mathbf{w}^t) \rangle
	\nonumber \\ + \sum_{k \in S^{t}_1} {\hat P}^t_k \frac{B}{\mu} \Big (\frac{L(\gamma+1)}{\mu'}+\gamma\Big) \norm{\nabla f(\mathbf{w}^t)}^2 \nonumber \\
	+ \Big ( \sum_{k \in S^{t}_1} {\hat P}^t_k \Big )^2 \frac{B^2(1+\gamma)^2}{\mu'^2} \norm{\nabla f(\mathbf{w}^t)}^2. \nonumber
	\end{align}
	Note that, with random selection of $S^t_1$ and $S^t_2$, we can define two random variables  $\sum_{k \in S^{t}_1}  \langle \nabla F_k(\mathbf{w}^t), \nabla_1 f(\mathbf{w}^t) \rangle$ and $\sum_{k' \in S^{t}_2}  \langle \nabla F_{k'}(\mathbf{w}^t), \nabla_2 f(\mathbf{w}^t) \rangle$ which follow the same distribution and $E \big[ \sum_{k \in S^{t}_1}  \langle \nabla F_k(\mathbf{w}^t), \nabla_1 f(\mathbf{w}^t) \rangle \big] = E \big[ \sum_{k' \in S^{t}_2} \langle \nabla F_{k'}(\mathbf{w}^t), \nabla_2 f(\mathbf{w}^t) \rangle \big]$. Taking expectation with respect to the uniformly random selection of devices in the two sets $S^{t}_1$ and $S^{t}_2$, and using Taylor's expansion give us
	\vspace{-0.03in}
	\begin{align*}
	E[f(\mathbf{w}^{t+1})] \leq f(\mathbf{w}^{t}) -  \frac{1}{\mu} E \Big[ \sum_{k \in S^{t}_1} {\hat P}^t_k \langle \nabla F_k(\mathbf{w}^t), \nabla_1 f(\mathbf{w}^t) \rangle \Big] \nonumber \\
	+ B \Big (\frac{L(\gamma+1)}{\mu \mu'}+\frac{\gamma}{\mu} + \frac{BL(1+\gamma)^2}{2 \mu'^2} \Big) \norm{\nabla f(\mathbf{w}^t)}^2.
	\end{align*}
	Since $S^{t}_1$ and $S^{t}_2$ are independent sets of random devices, the above inequality is equivalent to 
	\begin{align}
	&\mathbb{E}[f(\mathbf{w}^{t+1})] \leq f(\mathbf{w}^{t}) \nonumber \\ & - \frac{1}{\mu} \frac{\mathbb{E} \left[ \sum_{k \in S^{t}_1} \langle \nabla F_k(\mathbf{w}^t), \nabla_1 f(\mathbf{w}^t) \rangle \langle \nabla F_k(\mathbf{w}^t), \nabla_1 f(\mathbf{w}^t) \rangle \right]}{\mathbb{E} \left[\sum_{k' \in S^{t}_2} \langle \nabla F_{k'}(\mathbf{w}^t), \nabla_2 f(\mathbf{w}^t) \rangle \right] } \nonumber \\
	& + B \Big (\frac{L(\gamma+1)}{\mu \mu'}+\frac{\gamma}{\mu} + \frac{BL(1+\gamma)^2}{2 \mu'^2} \Big) \norm{\nabla f(\mathbf{w}^t)}^2.
	\label{eq:lb_loss_3}
	\end{align}
	In the term with expectations, we can apply Eq.~\ref{lem:s1} and ~\ref{lem:s2} from Lemma~\ref{lem:square} to the numerator and denominator, respectively, giving
	\vspace{-0.03in}
	\begin{align}
	& \mathbb{E}[f(\mathbf{w}^{t+1})] \leq f(\mathbf{w}^{t}) - \frac{1}{\mu} \frac{ \sum_{k \in [N]} \langle \nabla f(\mathbf{w}^t),\nabla F_k(\mathbf{w}^t)\rangle^2 }{ \sum_{k' \in [N]} |\langle \nabla f(\mathbf{w}^t),\nabla F_{k'}(\mathbf{w}^t)\rangle| } \nonumber \\
	& + B \Big (\frac{L(\gamma+1)}{\mu \mu'}+\frac{\gamma}{\mu} + \frac{BL(1+\gamma)^2}{2 \mu'^2} \Big) \norm{\nabla f(\mathbf{w}^t)}^2,
	\label{eq:lb_loss_4}
	\end{align}
	%Note that 
	%\begin{align}
	%	&E \Bigg[ \sum_{k \in S^{t}_1} \langle \nabla F_k(\mathbf{w}^t), \nabla_1 f(\mathbf{w}^t) \rangle^2 \Bigg] \nonumber \\
	%	&\quad\quad = \frac{K}{N} \sum_{k = 1}^{N} \langle \nabla f(\mathbf{w}^t),\nabla F_k(\mathbf{w}^t)\rangle^2 , \nonumber
	%\end{align}
	%and,
	%\begin{align}
	%	& E \Bigg[ \sum_{k' \in S^{t}_2} \langle \nabla F_{k'}(\mathbf{w}^t), \nabla_2 f(\mathbf{w}^t) \rangle \Bigg] \nonumber \\
	%	& \quad\quad \leq \frac{K}{N} \sum_{k' = 1}^{N} |\langle \nabla f(\mathbf{w}^t),\nabla F_{k'}(\mathbf{w}^t)\rangle|, \nonumber
	%\end{align}
	which is equivalent to \eqref{eq:lb-plb}.
\end{proof}
\vspace{-0.05in}

\subsection{Proof of Theorem~\ref{theo:loss_hetero}}
\begin{proof}
	From the $L$-Lipschitz continuity of $f$, we have
	\vspace{-0.03in}
	\begin{align}
	f(\mathbf{w}^{t+1}) \leq f(\mathbf{w}^{t}) + \langle \nabla f(\mathbf{w}^t), \mathbf{w}^{t+1} - \mathbf{w}^t \rangle + \frac{L}{2} \norm{\mathbf{w}^{t+1} - \mathbf{w}^t}^2.
	\label{eq:loss_h}
	\end{align}
	\vspace{-0.25in}
	
	\noindent We will bound the last two terms in the right-hand side of the above inequality as follows:
	
	\emph{$\bullet$ Bound $\norm{\mathbf{w}^{t+1} - \mathbf{w}^t}$}: Similar to the proof of Theorem~\ref{theo:loss}, we derive the following bound:
	\vspace{-0.03in}
	\begin{align}
	\norm{\mathbf{w}^{t+1} - \mathbf{w}^t}^2 \leq \Big(\frac{1}{K} \sum_{k \in S_t} \norm{\mathbf{w}^{t+1}_k - \mathbf{w}^t} \Big)^2 \nonumber \\
	\leq \frac{B^2}{K^2\mu'^2} \Big( \sum_{k \in S^{t}} (1 + \gamma_k) \Big)^2 \norm{\nabla f(\mathbf{w}^t)}^2.
	\label{eq:b1_h}
	\end{align}
	
	\emph{$\bullet$ Bound $\langle \nabla f(\mathbf{w}^t), \mathbf{w}^{t+1} - \mathbf{w}^t \rangle$}: Following the similar steps in the proof of Theorem~\ref{theo:loss}, we obtain the following:
	\begin{align}
	& \langle \nabla f(\mathbf{w}^t), \mathbf{w}^{t+1}_k - \mathbf{w}^t \rangle \leq -\frac{1}{\mu} \langle \nabla f(\mathbf{w}^t),\nabla F_k(\mathbf{w}^t)\rangle \nonumber \\ & \qquad \qquad + \frac{1}{\mu} \langle \nabla f(\mathbf{w}^t),(\nabla F_k(\mathbf{w}^t) - \nabla F_k (\mathbf{w}^{t+1}_k))\rangle \nonumber \\ 
	& \qquad\qquad + \frac{1}{\mu}\langle \nabla f(\mathbf{w}^t), (\nabla  F_k (\mathbf{w}^{t+1}_k) + \mu(\mathbf{w}^{t+1}_k - \mathbf{w}^t)) \rangle \nonumber \\
	& \leq - \frac{1}{\mu}\langle \nabla f(\mathbf{w}^t),\nabla F_k(\mathbf{w}^t)\rangle \nonumber \\ & \qquad\qquad  + \frac{1}{\mu} \norm{\nabla f(\mathbf{w}^t)} \norm{\nabla F_k(\mathbf{w}^t) - \nabla F_k (\mathbf{w}^{t+1}_k)} \nonumber \\
	& \qquad\qquad + \frac{1}{\mu} \norm{\nabla f(\mathbf{w}^t)} \norm{(\nabla  F_k (\mathbf{w}^{t+1}_k) + \mu(\mathbf{w}^{t+1}_k - \mathbf{w}^t))} \nonumber\\
	& \leq -\frac{1}{\mu}\langle \nabla f(\mathbf{w}^t),\nabla F_k(\mathbf{w}^t)\rangle + \frac{LB(1+\gamma_k)}{\mu \mu'} \norm{\nabla f(\mathbf{w}^t)}^2 \nonumber \\ & \qquad\qquad + \frac{B \gamma_k}{\mu} \norm{\nabla f(\mathbf{w}^t)}^2, \nonumber \\
	& \leq -\frac{1}{\mu}\langle \nabla f(\mathbf{w}^t),\nabla F_k(\mathbf{w}^t)\rangle + \frac{LB}{\mu \mu'} \norm{\nabla f(\mathbf{w}^t)}^2 \nonumber \\ & \qquad\qquad + \frac{B}{\mu} \left(\frac{L}{\mu'} + 1\right)\gamma_k \norm{\nabla f(\mathbf{w}^t)}^2,
	\label{eq:b2_1_h}
	\end{align}
	and, consequently,
	\begin{align}
	& \langle \nabla f(\mathbf{w}^t), \mathbf{w}^{t+1} - \mathbf{w}^t \rangle = \frac{1}{K} \sum_{k \in S_t} \langle \nabla f(\mathbf{w}^t), \mathbf{w}^{t+1}_k - \mathbf{w}^t \rangle \nonumber \\ & \leq -\frac{1}{K\mu}\sum_{k \in S_t}\langle \nabla f(\mathbf{w}^t),\nabla F_k(\mathbf{w}^t)\rangle \nonumber \\
	& + \frac{B}{\mu} \sum_{k \in S^{t}} \left(\frac{L}{\mu'} + 1\right)\gamma_k \norm{\nabla f(\mathbf{w}^t)}^2  + \frac{LB}{\mu\mu'} \norm{\nabla f(\mathbf{w}^t)}^2.
	\label{eq:b2_h}
	\end{align}
	Combine (\ref{eq:loss_h}), (\ref{eq:b1_h}) and (\ref{eq:b2_h}), we obtain:
	\begin{align}
	f(\mathbf{w}^{t+1}) \leq & f(\mathbf{w}^{t}) -\frac{1}{K\mu}\sum_{k \in S_t}\langle \nabla f(\mathbf{w}^t),\nabla F_k(\mathbf{w}^t)\rangle \nonumber \\ & + B \sum_{k \in S^{t}_1} \left(\frac{L}{\mu \mu'} + \frac{1}{\mu} + \frac{ L B}{K \mu'^2}\right)\gamma_k \norm{\nabla f(\mathbf{w}^t)}^2 \nonumber \\ 
	& + \frac{L B^2}{2 K^2 \mu'^2} \left(\sum_{k \in S^{t}} \gamma_k\right)^2 \norm{\nabla f(\mathbf{w}^t)}^2 \nonumber \\ & + \left( \frac{L B^2}{2 \mu'^2} + \frac{LB}{\mu\mu'}\right) \norm{\nabla f(\mathbf{w}^t)}^2.
	\end{align}
	Thus,
	\begin{align}
	& E[f(\mathbf{w}^{t+1})] \leq f(\mathbf{w}^{t}) -\frac{1}{K\mu} E \Bigg[ \sum_{k \in S_t} \Bigg( \langle \nabla f(\mathbf{w}^t),\nabla F_k(\mathbf{w}^t)\rangle \nonumber \\ & \qquad - B \left(\frac{L}{\mu \mu'} + \frac{1}{\mu} + \frac{ L B}{K \mu'^2}\right)\gamma_k \norm{\nabla f(\mathbf{w}^t)}^2 \nonumber \\ 
	& \qquad - \frac{L B^2}{2 K^2 \mu'^2} \sum_{k' \in S^{t}} \gamma_{k'} \gamma_k \norm{\nabla f(\mathbf{w}^t)}^2 \Bigg) \Bigg] \nonumber \\ & \qquad \quad + \left( \frac{L B^2}{2 \mu'^2} + \frac{LB}{\mu\mu'}\right) \norm{\nabla f(\mathbf{w}^t)}^2 \nonumber \\
	& \leq -\frac{1}{K\mu} E \Bigg[ \sum_{k \in S_t} \Bigg( \langle \nabla f(\mathbf{w}^t),\nabla F_k(\mathbf{w}^t)\rangle \nonumber \\ & \qquad - B \left(\frac{L}{\mu \mu'} + \frac{1}{\mu} + \frac{3 L B}{2 K \mu'^2}\right)\gamma_k \norm{\nabla f(\mathbf{w}^t)}^2 \Bigg) \Bigg] \nonumber \\
	& \qquad\quad + \left( \frac{L B^2}{2 \mu'^2} + \frac{LB}{\mu\mu'}\right) \norm{\nabla f(\mathbf{w}^t)}^2.
	\end{align}
	This completes the proof.
\end{proof}

\begin{IEEEbiography}[{\includegraphics[width=1in,height=1.25in,clip,keepaspectratio]{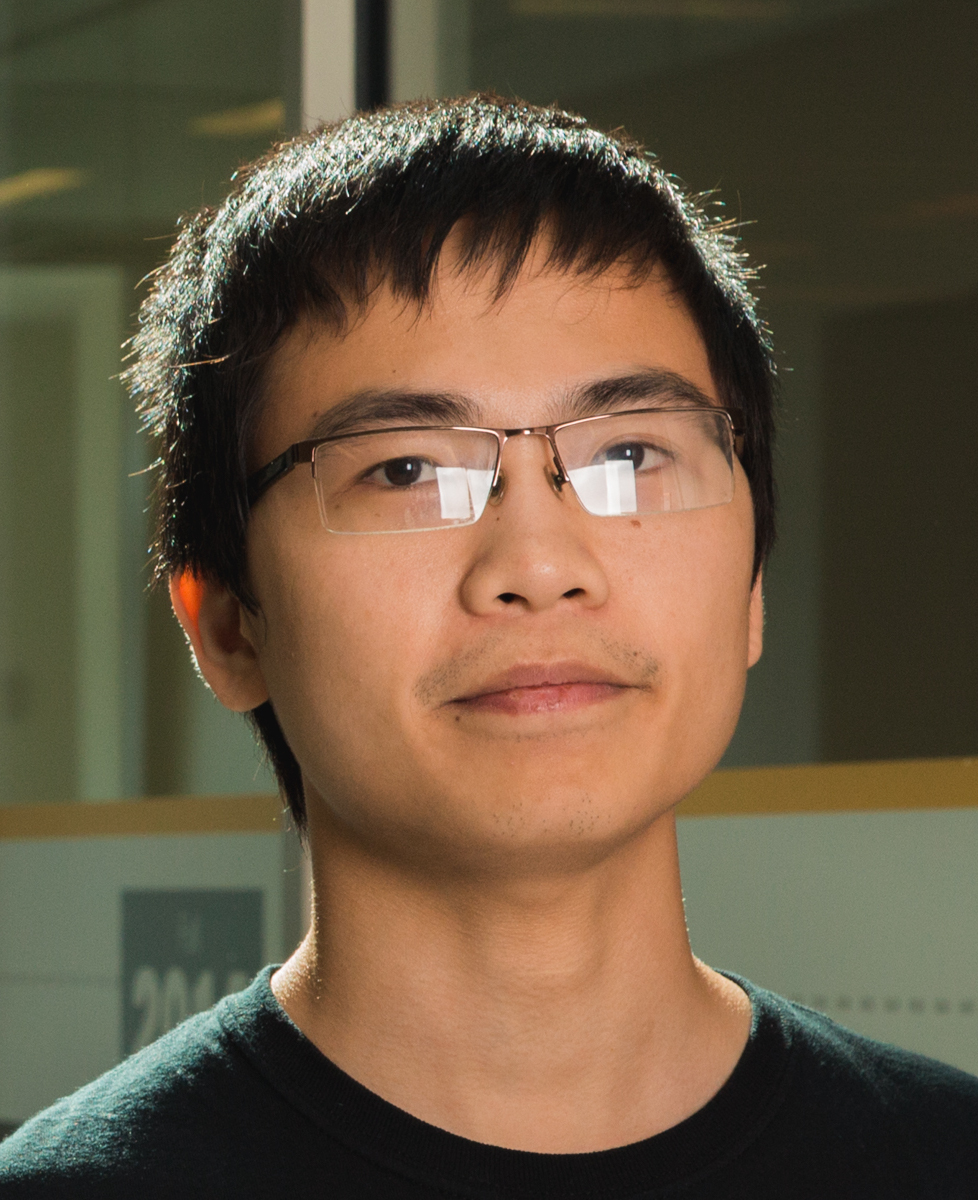}}]{Hung T. Nguyen} is a Postdoctoral Research Associate at Princeton University working with Professors Mung Chiang and Vincent Poor. His research focuses on distributed optimization, machine learning and approximation algorithms for graph problems. He received his PhD in computer science in 2018 from Virginia Commonwealth University and spent a year at Carnegie Mellon University as a postdoctoral fellow before joining Princeton University. He received multiple best paper awards from conferences and recognition from universities where he worked.
\end{IEEEbiography}

\begin{IEEEbiography}[{\includegraphics[width=1in,height=1.25in,clip,keepaspectratio]{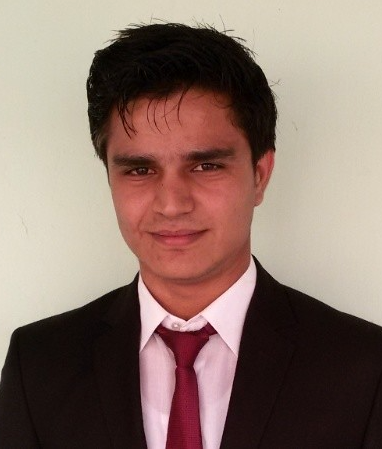}}]{Vikash Sehwag}
is a PhD candidate in electrical engineering at Princeton University. His primary research focuses on embedding the \textit{security by design} principle in machine learning where he aims to improve both performance and robustness of machine learning, simultaneously. His research interests include unsupervised machine learning, open-world machine learning, federated learning, and design of compact and efficient neural networks. He was the recipient of Qualcomm Innovation Fellowship 2019. 
\end{IEEEbiography}

\begin{IEEEbiography}[{\includegraphics[width=1in,height=1.25in,clip,keepaspectratio]{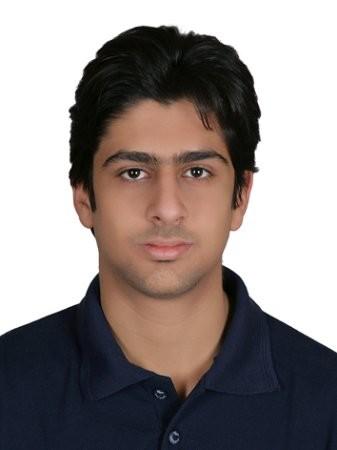}}]{Seyyedali Hosseinalipour}
received  B.S.  degree  from  Amirkabir University of Technology (Tehran Polytechnic) in 2015 and Ph.D. degree from NC State University in 2020, both in electrical engineering. He received ECE doctoral scholar of the year award at NC State. He is currently a postdoctoral researcher at Purdue University. His research interests mainly include analysis of modern wireless networks and communication systems.
\end{IEEEbiography}

\begin{IEEEbiography}[{\includegraphics[width=1in,height=1.25in,clip,keepaspectratio]{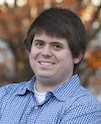}}]{Christopher G. Brinton} is an Assistant Professor of ECE at Purdue University. His research interest is at the intersection of data science and network optimization, particularly in distributed machine learning. Since joining Purdue in 2019, he has won the Purdue Seed for Success Award, the Purdue ECE Outstanding Faculty Mentor Award, and the 2020 Ruth and Joel Spira Outstanding Teacher Award. He received his Masters and PhD in Electrical Engineering from Princeton University in 2013 and 2016, respectively, where his research won the Bede Liu Best Dissertation Award. He is a co-founder of Zoomi Inc., and a co-author of the book \textit{The Power of Networks: 6 Principles that Connect our Lives}.
\end{IEEEbiography}

\begin{IEEEbiography}[{\includegraphics[width=1in,height=1.25in,clip,keepaspectratio]{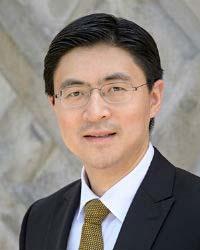}}]{Mung Chiang} was the Arthur LeGrand Doty Professor of electrical engineering at Princeton University, where he also served as the Director of the Keller Center for Innovations in Engineering Education and the inaugural Chairman of the Princeton Entrepreneurship Council. He founded the Princeton EDGE Lab in 2009, which bridges the theory-practice gap in edge networking research by spanning from proofs to prototypes. He also co-founded a few startup companies in mobile data, IoT, and AI, and co-founded the global nonprofit Open Fog Consortium. He is currently the John A. Edwardson Dean of the College of Engineering and the Roscoe H. George Professor of Electrical and Computer Engineering at Purdue University. His research on networking received the 2013 Alan T. Waterman Award, the highest honor to US young scientists and engineers. His textbook \emph{Networked Life}, popular science book \emph{The Power of Networks}, and online courses reached over 400,000 students since 2012. In 2019, he was named to the steering committee of the newly expanded Industrial Internet Consortium (IIC).
\end{IEEEbiography}

\begin{IEEEbiography}[{\includegraphics[width=1in,height=1.25in,clip,keepaspectratio]{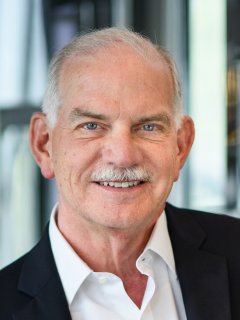}}]{H. Vincent Poor}
(S’72, M’77, SM’82, F’87) received the Ph.D. degree in EECS from Princeton University in 1977. From 1977 until 1990, he was on the faculty of the University of Illinois at Urbana-Champaign. Since 1990 he has been on the faculty at Princeton, where he is currently the Michael Henry Strater University Professor of Electrical Engineering. During 2006 to 2016, he served as Dean of Princeton’s School of Engineering and Applied Science. He has also held visiting appointments at several other universities, including most recently at Berkeley and Cambridge. His research interests are in the areas of information theory, machine learning and network science, and their applications in wireless networks, energy systems and related fields. Among his publications in these areas is the recent book \emph{Multiple Access Techniques for 5G Wireless Networks and Beyond.} (Springer, 2019).

Dr. Poor is a member of the National Academy of Engineering and the National Academy of Sciences, and is a foreign member of the Chinese Academy of Sciences, the Royal Society, and other national and international academies. Recent recognition of his work includes the 2017 IEEE Alexander Graham Bell Medal and a D.Eng. \emph{honoris causa} from the University of Waterloo awarded in 2019
\end{IEEEbiography}

\end{document}